\documentclass[10pt,letterpaper]{article}
\usepackage[utf8]{inputenc}
\usepackage[T1]{fontenc}
\usepackage{amsmath}
\usepackage{amsfonts}
\usepackage{amssymb}
\usepackage{amsthm}
\usepackage{graphicx}
\usepackage{csquotes}
\usepackage{microtype}
\usepackage{enumerate}
\usepackage{subcaption}
\usepackage[svgnames,table]{xcolor}
\usepackage[labelfont={bf}]{caption}
\usepackage[hidelinks]{hyperref}
\usepackage{cleveref}
\usepackage{calc}
\usepackage{bm}
\usepackage{rotating}
\usepackage[citestyle=authoryear,style=authoryear,natbib=true,sorting=nyt]{biblatex}
\usepackage{wasysym}
\usepackage{afterpage}
\usepackage{listings}
\usepackage{newfloat}
\usepackage{booktabs}
\DeclareFloatingEnvironment[
	fileext=lop,
	listname={List of Algorithms},
	name=Algorithm,
	placement=t]{algorithm}

\renewcommand{\epsilon}{\varepsilon}
\renewcommand{\phi}{\varphi}

\newtheorem{lemma}{Lemma}

\newtheorem{thm}{Theorem}
\theoremstyle{definition}
\newtheorem{defn}{Definition}

\newtheorem{example}{Example}

\renewcommand{\vec}[1]{\mathbf{#1}}
\newcommand{\mat}[1]{\mathbf{#1}}
\newcommand{\marginnote}[1]{\leavevmode\marginpar{\footnotesize\raggedright\slshape{#1}}}

\definecolor{tabBlue}{HTML}{4E79A7}
\definecolor{tabOrange}{HTML}{F28E2B}
\definecolor{tabRed}{HTML}{E15759}
\definecolor{tabLightBlue}{HTML}{76B7B2}
\definecolor{tabGreen}{HTML}{59A14F}
\definecolor{tabYellow}{HTML}{EDC948}
\definecolor{tabPurple}{HTML}{B07AA1}
\definecolor{tabPink}{HTML}{FF9DA7}
\definecolor{tabBrown}{HTML}{9C755F}
\definecolor{tabGray}{HTML}{BAB0AC}

\author{Andreas Stöckel\\Centre for Theoretical Neuroscience\\University of Waterloo}
\title{Discrete Function Bases and\\Convolutional Neural Networks}
\date{January 26, 2021}

\begin{document}
	\maketitle

	\begin{abstract}
	We discuss the notion of \enquote{discrete function bases} with a particular focus on the discrete basis derived from the Legendre Delay Network (LDN).
	We characterize the performance of these bases in a delay computation task, and as fixed temporal convolutions in neural networks. Networks using fixed temporal convolutions are conceptually simple and yield state-of-the-art results in tasks such as psMNIST.
	\end{abstract}
	\vfill
	\renewcommand{\abstractname}{Main Results}
	\begin{abstract}
	\vspace*{-1.25em}\noindent\begin{enumerate}[(1)]
		\item We present a numerically stable algorithm for constructing a matrix of DLOPs $\mat L$ in $\mathcal{O}(qN)$.
		\item The Legendre Delay Network (LDN) can be used to form a discrete function basis with a basis transformation matrix $\mat H \in \mathbb{R}^{q \times N}$.
		\item If $q < 300$, convolving with the LDN basis \emph{online} has a lower run-time complexity than convolving with arbitrary FIR filters.
		\item Sliding window transformations exist for some bases (Haar, cosine, Fourier) and require $\mathcal{O}(q)$ operations per sample and $\mathcal{O}(N)$ memory.
		\item LTI systems similar to the LDN can be constructed for many discrete function bases; the LDN system is superior in terms of having a finite impulse response.
		\item We compare discrete function bases by linearly decoding delays from signals represented with respect to these bases. Results are depicted in \cref{fig:evaluate_bases_contours}. Overall, decoding errors are similar. The LDN basis has the highest and the Fourier and cosine bases have the smallest errors.
		\item The Fourier and cosine bases feature a uniform decoding error for all delays. These bases should be used if the signal can be represented well in the Fourier domain.
		\item Neural network experiments suggest that fixed temporal convolutions can outperform learned convolutions.
		The basis choice is not critical; we roughly observe the same performance trends as in the delay task.
		\item The LDN is the right choice for small $q$, if the $\mathcal{O}(q)$ Euler update is feasible, and if the low $\mathcal{O}(q)$ memory requirement is of importance.
	\end{enumerate}
	\end{abstract}

  \begin{figure}[b]
  \setlength\fboxsep{0.5em}
  \noindent\colorbox{GhostWhite}{\parbox{\textwidth - 1em}{\small  Many of the equations in this technical report are accompanied by a Python reference implementation; the name of the corresponding Python function is indicated in the margin. The latest version of this code is on GitHub, see\\[0.125cm] \hspace*{0.5cm}\url{https://github.com/astoeckel/dlop_ldn_function_bases}
  }}
  \end{figure}
  
  \section{Introduction}

  The \enquote{Delay Network} is a recurrent neural network that approximates a time-delay of $\theta$ seconds \citep{voelker2018improving}.
  That is, given an input signal $u(t)$, the output of the network is approximately $u(t - \theta)$.
  \Citet{voelker2019} points out that the impulse response of a variant of the dynamical system underlying this network traces out the Legendre polynomials.
  We hence refer to the Delay Network as the \enquote{Legendre Delay Network} (LDN), and to the linear time-invariant (LTI) system underlying the LDN as the \enquote{LDN system}.

  \Citet{voelker2019lmu} demonstrate that a generalised neural network architecture derived from the LDN, the \enquote{Legendre Memory Unit} (LMU), can outperform other recurrent neural network architectures such as Long Short-Term Memories (LSTMs) in a wide variety of tasks.
  Preliminary work by Chilkuri and Eliasmith (publication in preparation) furthermore suggests that most weights in the LMU can be kept constant without negatively impacting the performance of the network.
  Surprisingly, this includes the recurrent connections in the LMU.
  Constant recurrent weights can be replaced by a set of static feed-forward Finite Impulse Response (FIR) filters arranged in a basis transformation matrix $\mat H$.
  This facilitates parallel training, leading to significant speed-ups.

  The basis transformation matrix $\mat H$ can be interpreted as a discrete function basis. This report is concerned with characterizing such function bases, including the related \enquote{Discrete Legendre Orthogonal Polynomials} (DLOPs) introduced by \citet{neuman1974discrete}.
  Our goal is to gain a better understanding of the LDN system and to explore whether it could make sense to instead use other discrete function bases.

  \paragraph{Structure of this report}
  We first review the notion of a \enquote{discrete function basis} and \enquote{generalised Fourier coefficients}.
  In particular, we discuss the Fourier and cosine series, as well as the Legendre polynomials.
  We review Discrete Legendre Orthogonal Polynomials (DLOPs), a discrete version of the Legendre polynomials proposed by \citet{neuman1974discrete}.
  We compare DLOPs to a discrete version of the LDN basis used by Chilkuri et al., followed by a method to reverse this process, i.e., to derive an LTI system from a (discrete) function basis.
  Furthermore, we discuss applying anti-aliasing filters to discrete function bases.
  We perform a series of experiments in which we characterise these bases in terms of the decoding error when computing delayed versions of signals represented in each basis.
  Lastly, we test each basis as a fixed temporal convolution in multi-layer neural networks and compare their performance to fully learned convolutions.

  \clearpage

  \section{Function Bases}

  As we will discuss in more detail in \Cref{sec:ldn_basis}, the LDN system can be characterized as continuously computing the generalised Fourier coefficients of an input signal $u(t)$ over a window $[t - \theta, t]$ with respect to the orthonormal Legendre function basis.
  The point of this section is to define more thoroughly what we mean by that.

  To this end, in \Cref{sec:function_bases}, we first review some basic concepts from functional analysis, a field of mathematics that generalises linear algebra to infinite-dimensional function spaces.
  In \Cref{sec:leg_poly_revisited}, we review the orthonomal Legendre polynomials as an example of an orthonormal continuous function basis.
  Readers already familiar with the topic are welcome to skip ahead to \Cref{sec:discrete_function_bases}, where we introduce the non-canonical notion of a discrete function basis and the corresponding notation, roughly following \citet{neuman1974discrete}.
  
  \subsection{Review: Function and Hilbert Spaces}
  \label{sec:function_bases}

  The concept of vector spaces in linear algebra is general enough to include infinite-dimensional spaces, or, in other words, spaces that can only be spanned by an infinite number of basis vectors.
  A mathematically useful subset of possible vector spaces that encompasses both finite- and infinite-dimensional spaces are so-called \enquote{Hilbert spaces}. We review this concept and discuss function bases that span the $L^2(a, b)$ Hilbert space, such as the Fourier and cosine bases.

  Most of the material in this subsection closely follows \citet{young1988introduction}.
  We strongly advise the reader to consult this book for a more thorough (and undoubtedly more correct) treatment of the topic.
  A recommended gentle introduction to linear algebra itself is \citet{hefferon2020linear}.
  Since we are not concerned with complex numbers in this report, we generally define all concepts over $\mathbb{R}$ instead of $\mathbb{C}$.

  \begin{defn}[{Inner product space, induced norm, induced metric; cf.~\cite{young1988introduction}, Definitions 1.2, 1.6, Theorem 2.3}]
  	An \emph{inner product space} is a vector space\footnote{A vector space is a set $V$ with an addition and scalar multiplication operation over a field $F$. These operations must fulfil a set of requirements; see \cite[Definition 1.1]{hefferon2020linear}.} $V$ with an associated inner product $\langle \cdot, \cdot \rangle : V \times V \longrightarrow \mathbb{R}$. The inner product must fulfil the following properties
  	\begin{enumerate}
  		\item \emph{Symmetry:} $\langle \vec x, \vec y \rangle = \langle \vec y, \vec x \rangle\,$.
  		\item \emph{Linearity:} $\langle \alpha \vec x + \vec y, \vec z \rangle = \alpha \langle \vec x, \vec y \rangle + \langle \vec y, \vec z \rangle$ for any $\alpha \in \mathbb{F}$.
  		\item \emph{Positive definite:} $\langle \vec x, \vec x \rangle > 0$ if $\vec x \neq 0\,$.
  	\end{enumerate}
  	If $\langle \vec x, \vec y \rangle = 0$, then $\vec x$ and $\vec y$ are called orthogonal. The norm $\|\vec x\| = \sqrt{\langle \vec x, \vec x \rangle}$ is the \emph{induced norm} of an inner product space; the metric $d(\vec x, \vec y) = \|\vec x - \vec y\|$ is its \emph{induced metric}.
  \end{defn}

  \begin{defn}[Function space]
  	A \emph{function space} is an inner product space with $V = \{ f \mid f : X \longrightarrow Y\}$. In other words, each $f \in V$ is a function mapping from a domain $X$ onto a codomain $Y$.
  	In this report we are concerned with $X, Y \subseteq \mathbb{R}$.
  \end{defn}

  \begin{defn}[Function basis]
  	A \emph{function basis} of a function space $V$ is an infinite sequence $(e_n)_{n \in \mathbb{N}}$ of linearly independent functions $e_n \in V$ that span $V$.
  	This is equivalent to demanding (cf.~Theorem~1.12~in \cite{hefferon2020linear}) that each function $f \in V$ must be representable as a \emph{unique} linear combination of $e_n$.
  	There exists a unique sequence $(\xi_n)_{n \in \mathbb{N}}$ over $\mathbb{R}$ such that $f(x) = \sum_{n = 0}^\infty \xi_n e_n(x)$ for each $f \in V$.
  	Conversely, each function $f$ constructed through such a linear combination must be an element of $V$.
  \end{defn}

  \begin{defn}[Orthogonal and orthonormal function bases]
  	A function basis is \emph{orthogonal} if $\langle f_i, f_j \rangle = 0 \Leftrightarrow i \neq j$. A function basis is \emph{orthonormal} if, additionally,  $\langle f_i, f_j \rangle = 1 \Leftrightarrow i = j$.%
  	\footnote{Note that the concept of a (function) basis being orthogonal is confusingly different from that of a matrix $\mat A$ being orthogonal, which is defined as $\mat A^T \mat A = \mat I$ and thus closer to the concept of an orthonormal basis.}
  \end{defn}

  \begin{example}[{Continuous function space $\mathcal{C}[a, b]$}]
  An example of a function space would be the set of continuous scalar functions over an interval $[a, b]$, denoted as
  \begin{align*}
  	\mathcal{C}[a, b] &= \{ f \mid f : [a, b] \longrightarrow \mathbb{R} \text{ and } f \text{ is continuous} \} \,.
  \end{align*}
  This set is a vector space when coupled with addition $(f + g)(x) = f(x) + g(x)$ and scalar multiplication $(\lambda f)(x) = \lambda f(x)$ for $\lambda \in \mathbb{R}$. Furthermore, it can be shown that the following inner product over $\mathcal{C}[a, b]$ fulfils the above properties:
  \begin{align}
  	\langle f, g \rangle &= \int_a^b f(x) g(x) \,\mathrm{d}x \,.
  	\label{eqn:C_inner_product}
  \end{align}
  \end{example}

  One might be inclined to think that the concept of a continuous function space $\mathcal{C}[a, b]$ is sufficient for most purposes. However, when trying to find a basis that spans $\mathcal{C}[a, b]$, one would eventually notice that any candidate basis can be used to generate discontinuous functions. Thus, the candidate basis does not span $\mathcal{C}[a, b]$, but a slightly larger space.
  The next example illustrates this.

  \begin{figure}[t]
    \centering
    \includegraphics{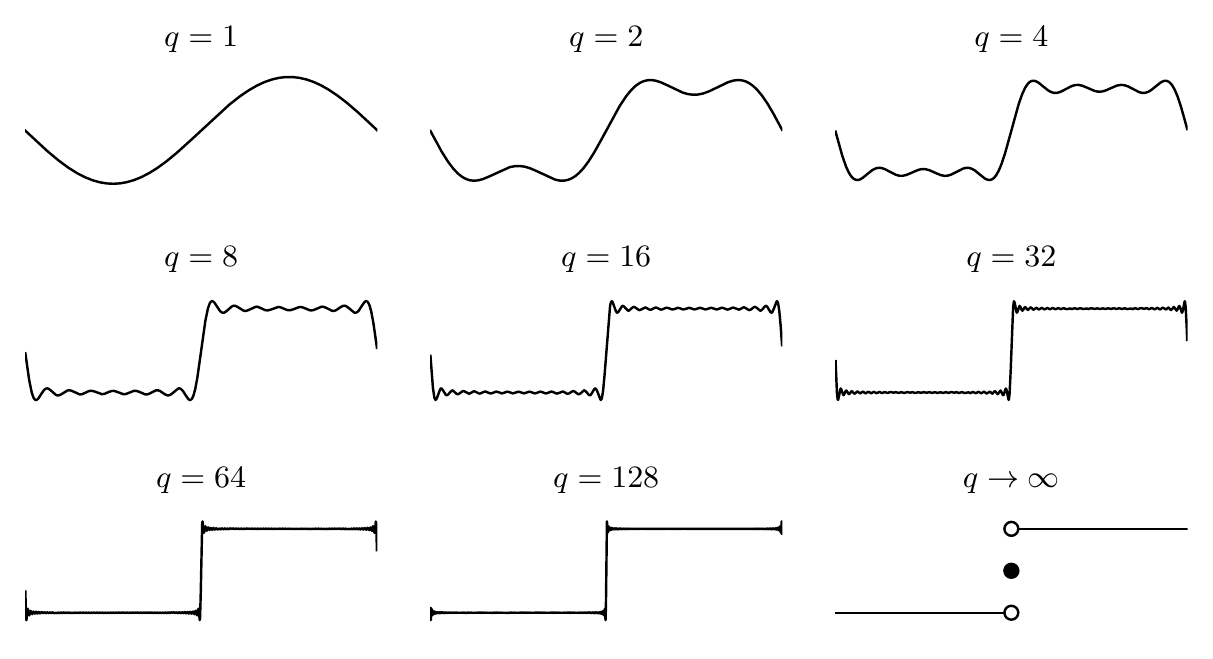}
  	\caption{Approximating the discontinuous sign function (bottom right) using a sum of continuous sine waves according to \cref{eqn:sign_sine}.
  	Cauchy sequences of continuous functions can converge to a discontinuous function.}
  	\label{fig:sign_sine}
  \end{figure}

  \begin{example}[Sign function as a series of continuous functions]
  Consider the following sequence of basis functions $(f'_n)_{n \in \mathbb{N}}$ over $\mathcal{C}[-\pi, \pi]$
  \begin{align}
  	f'_0(x) &= \frac{1}{\sqrt{2\pi}} \,,&
  	f'_{2n + 1}(x) &= \frac{\sin\big(nx\big)}{\sqrt{\pi}} \,, &
  	f'_{2n} &= \frac{\cos(nx)}{\sqrt{\pi}} \,.
  	\label{eqn:fourier_series_pi}
  \end{align}
  This basis is a variant of the \enquote{canonical Fourier series}, an orthonormal function basis.
  Each individual $f'_n$ is obviously in $\mathcal{C}[-\pi, \pi]$, and a linear combination of these basis functions can approximate any function in $\mathcal{C}[-\pi, \pi]$ (this follows from Theorem 5.1 in \cite{young1988introduction}).
  However, the same basis can be used to express discontinuous functions.
  For example, one can show that a weighted series of the sine terms of $f_n$ is equal to the sign function (see also \cref{fig:sign_sine}):
  \begin{align}
  	\mathrm{sign}(x) &= \lim_{q \to \infty}\sum_{n = 1}^q \frac{4 \sin\big((2n - 1)x\big)}{(2n - 1) \pi} = \begin{cases}
  		1 & \text{if } x > 0 \,,\\
  		0 & \text{if } x = 0 \,,\\
  		-1 & \text{if } x < 0 \,.
  	\end{cases}
  	\label{eqn:sign_sine}
  \end{align}
  Hence $\mathcal{C}[-\pi, \pi]$ has no basis that spans the space, which is slightly problematic.
  \end{example}

  The notion of a \enquote{Hilbert space} restricts inner product spaces to those in which \enquote{well-behaved} sequences, so-called Cauchy sequences, converge to an element within that space.
  The sum of sines sequence implicitly defined in \cref{eqn:sign_sine} is an example of such a Cauchy sequence.
  Correspondingly, the function space $\mathcal{C}[a, b]$ cannot be a Hilbert space.

  \begin{defn}[{Hilbert space; cf.~\cite{young1988introduction}, Definition 3.4}]
  	A \emph{Hilbert space} is an inner product space $V$ in which all Cauchy sequences (relative to the metric induced by the inner product) converge to an element in $V$.
  \end{defn}

  \begin{example}[{The Hilbert space $L^2(a, b)$; cf.~\cite{young1988introduction}, Example 3.5, Theorem 5.1}]
    The Fourier series in \cref{eqn:fourier_series_pi} spans $L^2(-\pi, \pi)$. In general $L^2(a, b)$ is a function space $V$ with the inner product defined in \cref{eqn:C_inner_product}. Each $f \in V$ is a function $f : [a, b] \longrightarrow \mathbb{R}$ for which the following Lebesgue integral converges; i.e., the function is square Lebesgue integrable:
    \begin{align*}
    	\int_{a}^b f(x)^2 \,\mathrm{d}t < \infty \,, \quad \text{ where \enquote{$\textstyle\int$} is the Lebesque integral.}
    \end{align*}
    Note that $L^2(a, b)$ is a superset of all square Riemann integrable functions.
  \end{example}

  Since the canonical Fourier series defined in \cref{eqn:fourier_series_pi} spans $L^2(-\pi, \pi)$, any function in $L^2(-\pi, \pi)$ can be represented as a linear combination of Fourier basis functions.
  Of course, the same holds for any orthonormal function basis.

  \begin{defn}[{Generalised Fourier series and coefficients; cf.~\cite{young1988introduction}, Definition 4.3}]
  Consider an orthonormal basis $(e_n)_{n \in \mathbb{N}}$, where each $e_n \in L^2(a, b)$, as well as a function $f \in L^2(a, b)$. Then the series
  \begin{align*}
    f &= \sum_{n = 0}^\infty \langle f, e_n \rangle e_n = \sum_{n = 0}^\infty \xi_n e_n
  \end{align*}
  is the generalised Fourier series of $f$ and $\xi_n$ are the generalised Fourier coefficients, also called the \emph{spectrum} of $f$.
  \end{defn}

  Note that the canonical Fourier series from \cref{eqn:fourier_series_pi} and the associated Fourier coefficients $\xi_n$ are related to, but not to be confused with, the Fourier \emph{transformation}.
  The Fourier transformation represents any integrable function $f$ in terms of a another function $\hat f(\xi)$.

  In the following, we provide equations for the Fourier, cosine, and Legendre basis.
  We already introduced the canonical Fourier series in an example above over the interval $[-\pi, \pi]$.
  From now on, we define all bases over the interval $[0, 1]$ for the sake of consistency.
  Any orthonormal basis function $e_n$ over $[0, 1]$ can be easily converted to an orthonormal basis function $e'_n$ over $[a, b]$:
  \begin{align}
  	e'_n(x) &= \frac{1}{\sqrt{|b - a|}} e_n\left( \frac{x - a}{b - a} \right) \,.
  	\label{eqn:rescale}
  \end{align}

\begin{figure}[p]
  	\begin{subfigure}{\textwidth}
	  	\centering
	  	\includegraphics{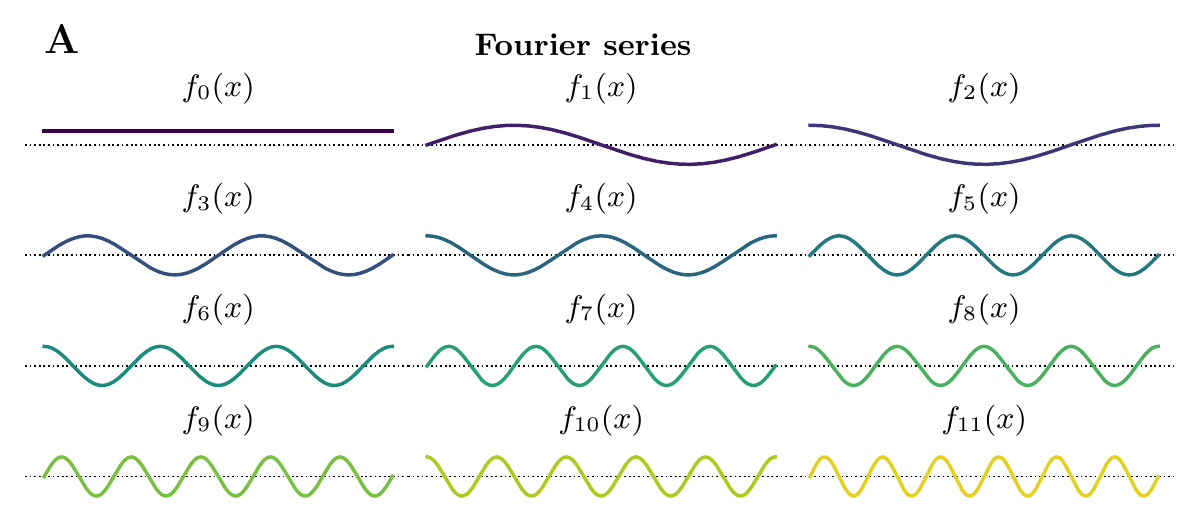}
	  	\phantomcaption
	  	\label{fig:fourier_series}
	\end{subfigure}
  	\begin{subfigure}{\textwidth}
	  	\centering
	  	\includegraphics{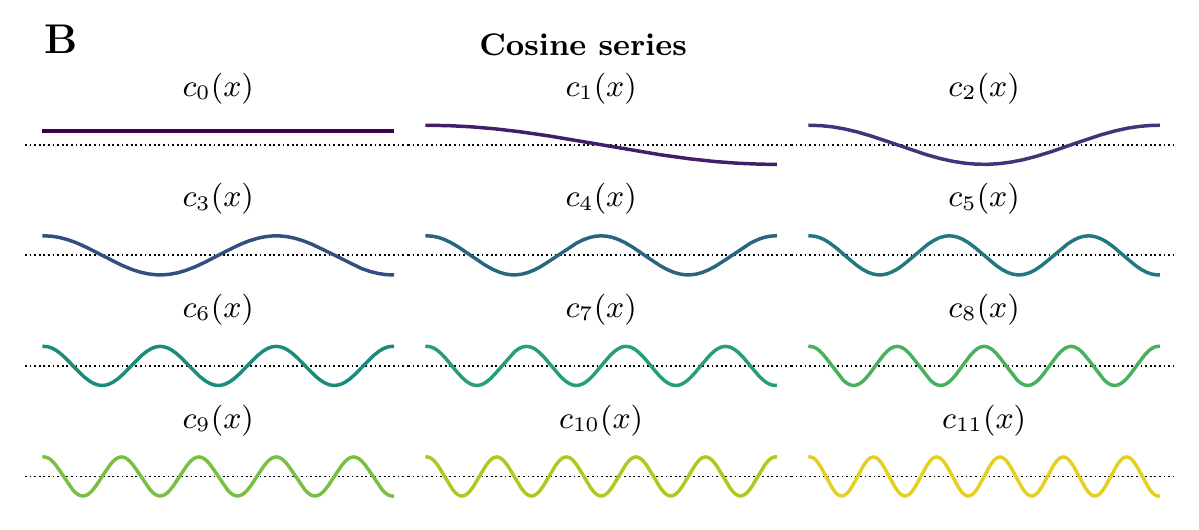}
	  	\phantomcaption
	  	\label{fig:cosine_series}
	\end{subfigure}
  	\begin{subfigure}{\textwidth}
	  	\centering
	  	\includegraphics{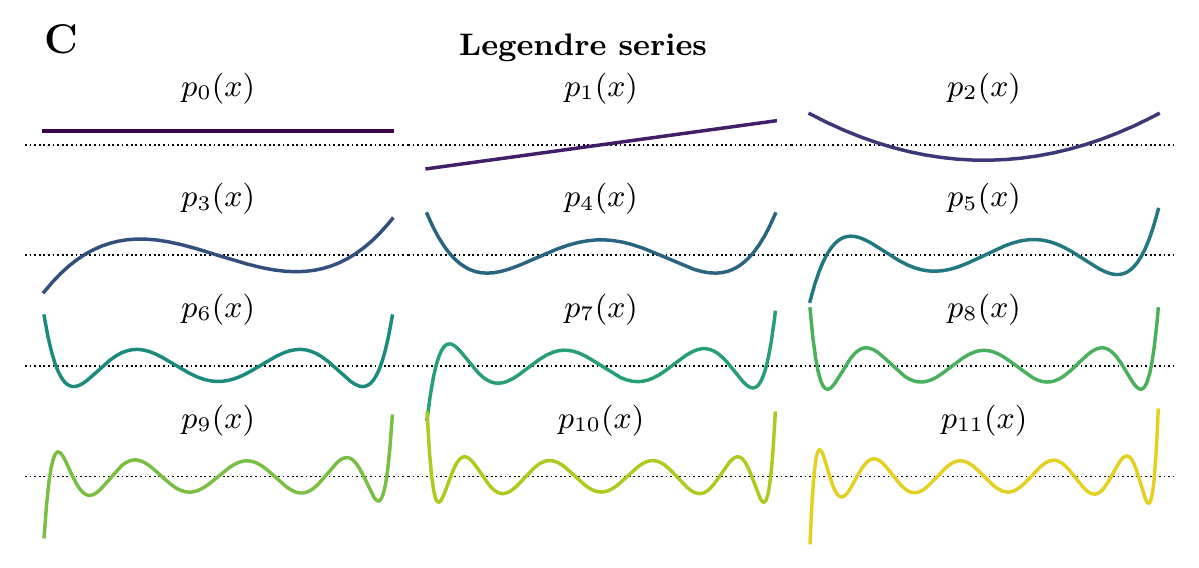}
	  	\phantomcaption
	  	\label{fig:legendre_series}
	\end{subfigure}
	\caption{First functions in the orthonormal bases discussed in this section.
	All functions are plotted to the same scale; axes were omitted as the functions can be rescaled as described in \cref{eqn:rescale}. Dotted line is zero.}
  \end{figure}

  \begin{defn}[Fourier series]
  The Fourier series $(f_n)_{n \in \mathbb{N}}$ over $[0, 1]$ is given as
  \begin{align}
  	f_0(x) &= 1 \,,&
  	f_{2n + 1}(x) &= \sqrt{2}\sin\big(2 \pi nx\big) \,, &
  	f_{2n} &= \sqrt{2}\cos\big(2 \pi nx\big) \,.
  	\label{eqn:fourier_series}
  \end{align}
  This orthonormal basis spans $L^2(0, 1)$ and is depicted in \Cref{fig:fourier_series}.
  \end{defn}

  \begin{defn}[Cosine series]
  An arguably simpler alternative to the Fourier series is the cosine series.
  The cosine series skips the \enquote{sine} terms of the Fourier series and increments the frequency in steps of $\pi$ instead of $2\pi$.
  \begin{align}
  	c_0(x) &= 1 \,, &
  	c_n(x) &= \sqrt{2}\cos(\pi n x) \,.
  	\label{eqn:cosine_basis}
  \end{align}
  The orthonormal cosine series spans $L^2(0, 1)$ and is depicted in \Cref{fig:cosine_series}.
  \end{defn}

  \subsection{Review: Legendre Polynomials}  
  \label{sec:leg_poly_revisited}

  In this report, we are mostly interested in the orthonormal Legendre basis generated by the Legendre polynomials.
  We first review the \emph{orthogonal} Legendre polynomials $p'_n$; the \emph{orthonormal} polynomials $p_n$ are a scaled version of $p'_n$
  
  \begin{defn}[Legendre polynomials]
  \label{def:legendre_polynomial}
  Legendre polynomials are uniquely defined as a sequence of functions $(p'_n)_{n \in \mathbb{N}}$ over $[-1, 1]$ with the following properties
  \begin{enumerate}
    \item \emph{Polynomial:} $p'_n$ is a linear combination of $n$ monomials $p'_n(x) = \sum_{i = 0}^n \alpha_{n, i} x^n \,$.
  	\item \emph{Orthogonal:} $\langle p'_i, p'_j \rangle = 0$ exactly if $i \neq j\,$.
  	\item \emph{Normalisation:} $p'_n(1) = 1\,$.
  \end{enumerate}
  Below we summarize two methods to construct $p'_n$ that fulfil these properties.
  \end{defn}

  \paragraph{Recurrence relation} Starting with the base cases $p'_0(x) = 1$ and $p'_1(x) = x$, $p'_n(x)$ is given as a recurrence relation \citep[Section 5.4, p. 219]{press2007numerical}:
  \begin{align}
  	(n + 1) p'_{n + 1}(x) &= (2n + 1) x p'_n(x) - n p'_{n - 1}(x) \,.
  	\label{eqn:leg_rec}
  \end{align}
  Rewriting this in terms of the polynomial coefficients $\alpha_{n, i}$ (see above) we get
  \begin{align*}
  	(n + 1) \alpha_{n + 1, i} &= \begin{cases}
  		\hspace{6.6em} - \, n \alpha_{n - 1, i} & \text{if } i = 0 \,, \\
  		(2n + 1) \alpha_{n, i - 1} - n \alpha_{n - 1, i} & \text{if } i > 0 \,.
  	\end{cases}
  \end{align*}

  \paragraph{Closed form equation}
  Alternatively, the Legendre polynomial $p'_n$ is given in closed form as
  \begin{align*}
  	p'_n(x) &= \sum_{i = 0}^n \binom{n}{i} \binom{n + i}{i} \left( \frac{x - 1}{2}\right)^i \,.
  \end{align*}
  The Legendre Delay Network approximates the shifted Legendre polynomials $\tilde p_n$ over $[0, 1]$ given as $\tilde p_n(x) = p_n(2 x - 1)$.
  This substitution yields
  \begin{align}
  	\tilde p_n(x) &= (-1)^n \sum_{i = 0}^n (-1)^i \binom{n}{i} \binom{n + i}{i} x^i \,.
  	\label{eqn:shifted_legendre}
  \end{align}
  This equation can be easily decomposed into the monomial coefficients $\alpha_{n, i}$.

  \begin{defn}[Orthonormal Legendre series]
    We can derive an orthnormal function basis $(p_n)_{n \in \mathbb{N}}$ over $[0, 1]$ simply by dividing each shifted polynomial by the norm $\|\tilde p_n\|$. The basis is depicted in \Cref{fig:legendre_series}. It spans $L^2(0, 1)$, just like the Fourier and cosine basis.\footnote{Sketch of a proof: for a bounded function space, an orthogonal polynomial function basis can be used to construct any analytic function over that interval. This includes sine and cosine, which can be made to span $L^2(0, 1)$ by forming a canonical Fourier series.}
    \begin{align}
    	p_n(x) &= \frac{\tilde p_n(x)}{\| \tilde p_n \|} = \sqrt{2n + 1} \, (-1)^n \sum_{i = 0}^n (-1)^i \binom{n}{i} \binom{n + i}{i} x^i \,.
    	\label{eqn:legendre_basis}
    \end{align}
  \end{defn}

  \subsection{Discrete Function Bases}
  \label{sec:discrete_function_bases}

  From a mathematical perspective, the notion of \enquote{discrete function bases} is at most moderately exciting.
  Once we discretise functions over an interval, we end up with boring, finite-dimensional vectors.
  Unfortunately, in practice, we more often than not have to work with discrete functions.
  Still, there is some potential for defining the concept of \enquote{discrete function bases} in relation to their continuous counterparts more rigorously.

  Below, we define the notion of a \enquote{discrete function basis}, as well as the corresponding \enquote{basis transformation matrix}.
  Although our definitions are non-canonical, our notation roughly follows \citet{neuman1974discrete}.

  \begin{defn}[Discrete Function Basis]
    \label{def:discrete_function_basis}
  	A \emph{discrete function basis} with an associated continuous function basis $(e_n)_{n \in \mathbb{N}}$ is a finite sequence of discrete basis functions $(E_n(k; N))_{n < N}$.
  	$n \in \{0, \ldots, N - 1\}$ is the basis function index, $k \in \{0, \ldots, N - 1\}$ is the sample index, and $N \geq 1 \in \mathbb{N}$ is the number of samples.
  	The codomain of $E_n(k; N)$ is $\mathbb{R}$.
  	In the limit of $N \to \infty$ it must hold
  	\begin{align}
  		\lim_{N \to \infty} E_n \left( k; N \right) = \frac{1}{\sqrt{N}} e_n\left(\frac{k}{N - 1}\right) \,.
  		\label{eqn:discrete_basis}
  	\end{align}
  \end{defn}

  Intuitively, when sampling densely, an $E_n$ fulfilling the above definition is indistinguishable from a scaled continuous basis function $e_n$.
  The scaling factor $1 / \sqrt{N}$ ensures that inner products are preserved. It holds:
  \begin{align*}
  	\langle e_i, e_j \rangle = \lim_{N \to \infty} \sum_{k = 0}^{N - 1} E_i(k; N) E_j(k; N) \,.
  \end{align*}

  \begin{defn}[Basis Transformation Matrix]
  	Let $E_n(k; N)$ be a discrete function basis. Given an order $q \leq N$, the basis transformation matrix $\mat E \in \mathbb{R}^{q \times N}$ is defined as
  	\begin{align*}
  		\big(\mat{E}\big)_{ij} &= \frac{E_{i - 1}(j - 1; N)}{\sqrt{\sum_{k = 0}^{N - 1} E_{i - 1}(k; N)^2}} \,, \quad \text{where } i \in \{1, \ldots, q\} \,, j \in \{ 1, \ldots, N \} \,.
  	\end{align*}
  	The denominator ensures that each basis vector $(\mat E)_n$ (the $n$th row in $\mat E$) has unit length.
  	We call $\mat E$ \emph{orthogonal} if $\mat E^T \mat E = \mat I$, where $\mat I$ is the $q \times q$ identity matrix. In contrast to the canonical meaning of \enquote{orthogonal}, this includes non-square $\mat E$.
  \end{defn}

  \paragraph{Interpreting $\mat E$ as a basis transformation}
  Let $\vec u = (u_1, \ldots, u_{N})$ be a discrete signal consisting of $N$ samples. The matrix-vector product $\mat{E} \vec u = \vec m$ results in $q$ inner products $\vec m = (m_1, \ldots, m_{q})$ between the input signal $\vec u$ and each of the discrete basis functions in $\mat E$.
  For orthogonal $\mat E$, the resulting $\vec m$ can be interpreted as a set of discrete generalised Fourier coefficients.
  That is, the vector $\mat m$ represents the signal $\vec u$ with respect to a normalised version of the discrete function basis $E_n(k; N)$.
  Correspondingly, this operation is a basis transformation in the same sense as the discrete Fourier and cosine transformations (discussed below).

  \paragraph{Interpreting $\mat E$ as a set of FIR filters}
  Another way to think about $\mat E$ is as a set of finite impulse-response (FIR) filters. Let $\vec u_t$ represent the last $N$ samples of an input signal relative to a time $t$. Specifically, $\vec u_t = (u_{t - (N - 1)}, \ldots, u_t)$; i.e., the newest sample is shifted in from the right. Then, $\vec m = \mat E \vec u_t$ can be written as
  \begin{align}
  	m_n &= \big\langle \big(\mat E\big)_n, \vec u \big\rangle = \sum_{k = 0}^{N - 1} \big(\mat E\big)_{n, N - k} u_{t - k} \,.
  	\label{eqn:fir}
  \end{align}
  This is exactly the definition of a FIR filter of order $N - 1$ (cf.~\cite[Section 13.5.1, p.~668]{press2007numerical}); but notice the inverted matrix column order $N - k$.

  Now that we have defined discrete function bases and the corresponding normalised basis transformation matrix, we should discuss how to construct discrete function bases.
  Unfortunately, there is no universal method, and the next two examples are only two of many possible methods.

  \begin{figure}
    \centering
  	\includegraphics{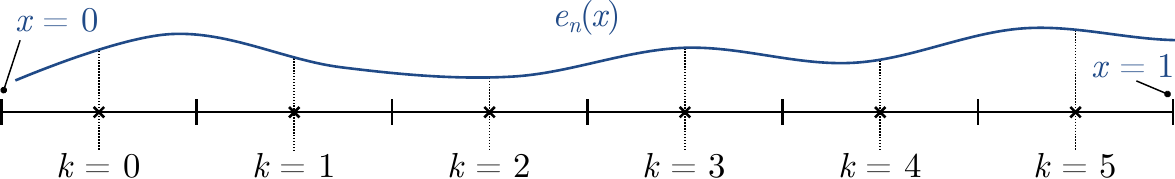}
  	\caption{Illustration of the sampling process in \cref{eqn:naive_sampling}. The function $e_n(x)$ is sampled at the centre of $N = 6$ intervals.}
  	\label{fig:sampling}
  \end{figure}
  \begin{example}[Naive sampling]
  \Cref{eqn:discrete_basis} directly suggests a way to generate discrete function bases. This \enquote{naively sampled discrete function basis} is
  \begin{align}
  	E_n ( k; N ) = \frac{1}{\sqrt{N}} e_n\left(\frac{k + \frac{1}2}{N}\right) \,.
  	\label{eqn:naive_sampling}
  \end{align}
  The offset of one half ensures that samples are taken at the centre of the $N$  discrete intervals (cf.~\cref{fig:sampling}).
  \end{example}

  \begin{example}[Mean sampling]
  \label{xpl:mean_sampling}
  Another way to construct a discrete function basis is to compute the mean over each of the $N$ intervals. That is
  \begin{align}
   	E_n ( k; N ) = \sqrt{N} \int_{x_0}^{x_1} e_n(x) \,\mathrm{d}x \,, &
   	\text{ where } x_0 = k / N \text{ and } x_1 = (k + 1) / N \,.
   	\label{eqn:mean_sampling}
  \end{align}
  \end{example}
  
  Unfortunately, one caveat with both sampling methods is that they do not necessarily preserve orthogonality of the function basis that is being sampled.
  This is illustrated in the three examples below.
  While naive sampling perfectly preserves orthogonality of the Fourier and cosine series, neither method results in an orthogonal basis transformation matrix for the Legendre polynomials.

  Maintaining orthogonality can be important.
  Mathematically, having orthogonal matrices can simplify some equations, as we will see later.
  From an information-theoretical perspective, orthogonal bases minimize correlations between individual state dimensions and thus (when considering a probability distribution of input signals) minimize pairwise mutual information between the generalised Fourier coefficients, maximizing their negative entropy \citep[cf.][Sections 2.1-2.3 for definitions and the relationship between mutual information and negentropy]{comon1994independent}.
  It should be noted that this can be undesirable if the resulting representation is subject to noise.

\begin{figure}[p]
  	\centering
  	\includegraphics{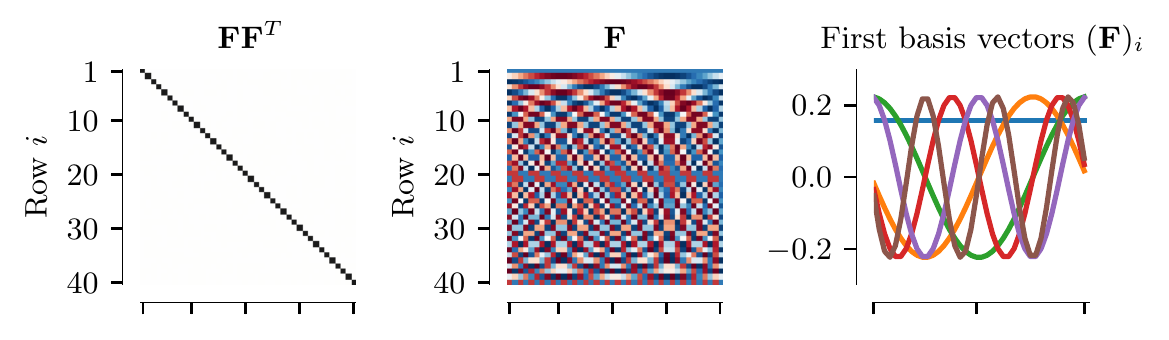}\\[-0.125cm]
  	\includegraphics{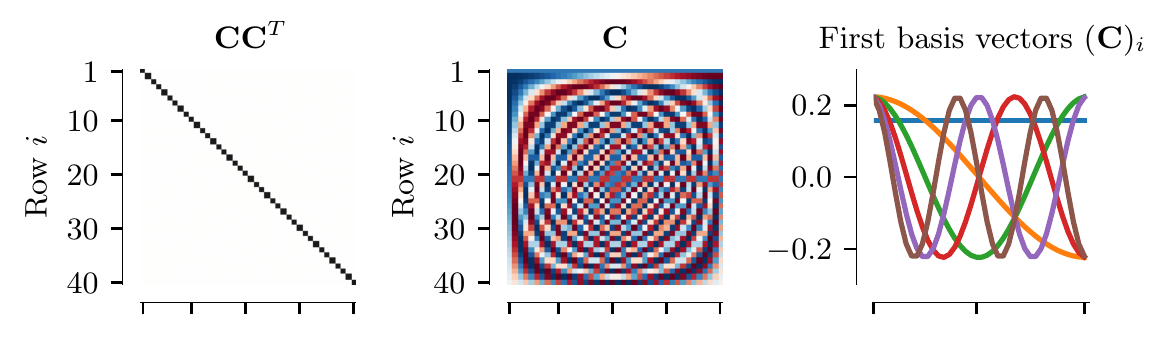}\\[-0.125cm]
  	\includegraphics{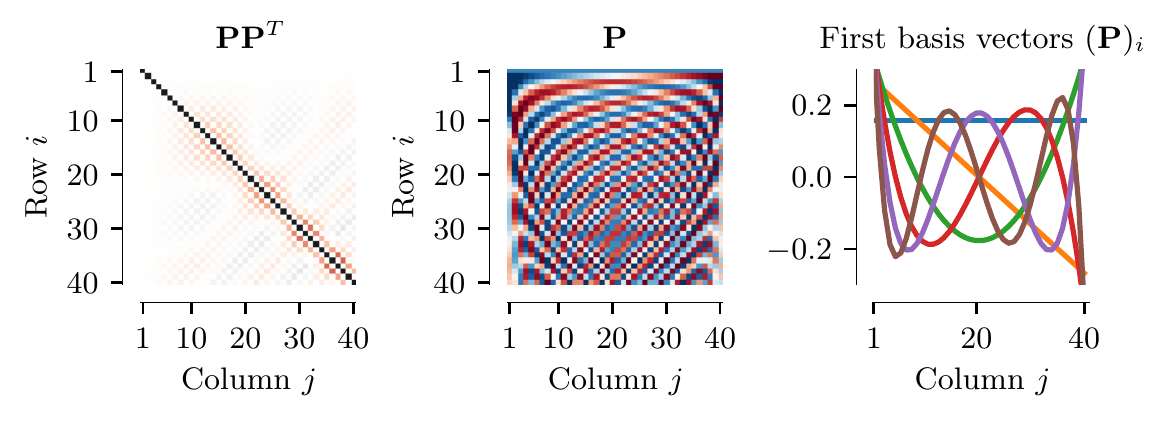}\\[-0.125cm]
  	\includegraphics{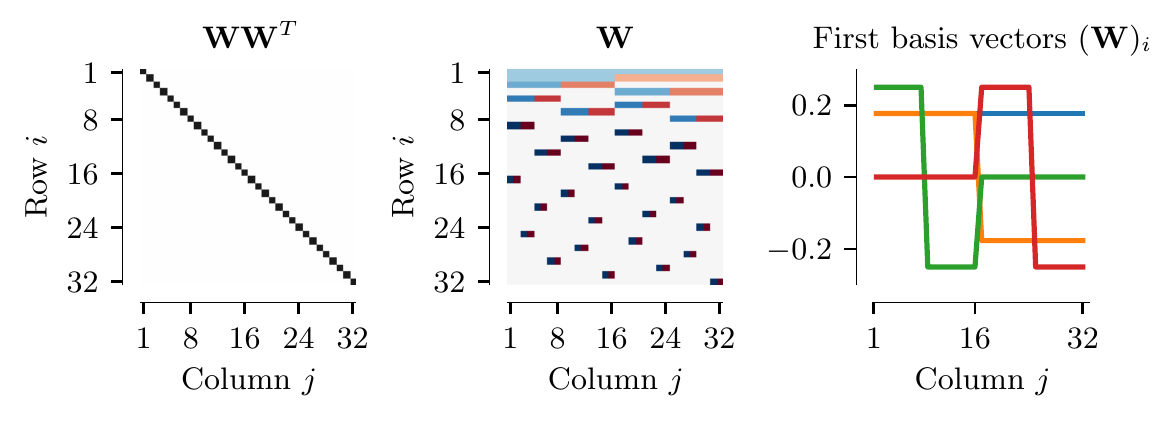}
  	\caption{The discrete Fourier, cosine, Legendre and Haar basis transformation matrices (top to bottom). \emph{Left:} Visualisation of the outer product of each matrix. Each pixel is a matrix cell $i, j$. Zero is white, one is black, negative values are red. The matrices $\mat F$, $\mat C$, $\mat W$ are orthogonal; $\mat P$ is not orthogonal. \emph{Centre:} Basis matrices themselves, where white corresponds to zero, red to negative and blue to positive numbers (colour maps rescaled to cover 95\% of the represented values without saturating). \emph{Right:} Visualisation of the first basis vectors.}
  	\label{fig:dft_dct_basis}
  \end{figure}

  \begin{example}[Discrete Fourier Basis]
  As mentioned above, applying naive sampling from \cref{eqn:naive_sampling} to the Fourier series in \cref{eqn:fourier_series} yields an orthogonal basis transformation matrix $\mat F$.
  This $\mat F$ can be interpreted as the linear operator implementing the discrete Fourier transformation (DFT).
  That is, multiplying a real signal $\vec u$ with $\mat F$ computes the DFT of $\vec u$.
  The individual discrete basis functions are given as
  \begin{align}
  	\begin{aligned}
  	F_0(k; N) &= \frac{1}{\sqrt{N}} \,,\\
  	F_{2n + 1}(k; N) &= \frac{\sqrt{2}}{\sqrt{N}} \sin\left(
  		2 \pi n \frac{k + \frac{1}2}{N}\right) \,, \\
  	F_{2n}(k; N) &= \frac{\sqrt{2}}{\sqrt{N}} \cos\left(
  		2 \pi n \frac{k + \frac{1}2}{N}\right) \,.
    \end{aligned}
    \label{eqn:dft}
  \end{align}
  \marginnote{~~\\[-12.5em]This equation is implemented in the function \texttt{mk\_fourier\_basis}\,.}%
  Normalisation of the matrix $\mat F$ as defined in \cref{eqn:discrete_basis} is not required if one special case is taken into account.
  The normalisation factor needs to be updated in the case $n + 1 = q = N$ for even $N$
  \begin{align*}
  	F_{N - 1}(k; N) &= \frac{1}{\sqrt{N}} \sin\left(
  	  		2 \pi n \frac{k + \frac{1}2}{N}\right) = \frac{(-1)^k}{2 \pi} \quad \text{if $N = q$ even.}
  \end{align*}
  Taking this special case into account, the discrete Fourier function basis is orthonormal, i.e., it holds
  \begin{align*}
  	\sum_{k = 0}^{N - 1} F_{i}(k; N) F_{j}(k; N) = \begin{cases}
  		1 & \text{if } i = j \,, \\
  		0 & \text{if } i \neq j \,.
  	\end{cases}
  \end{align*}
  The corresponding basis transformation matrix $\mat F$ is depicted in \Cref{fig:dft_dct_basis}.
  \end{example}

  \begin{example}[Discrete Cosine Basis]
  Similarly to the discretisation of the Fourier series, applying naive discretisation from \cref{eqn:naive_sampling} to the cosine series results in the discrete Cosine transformation (DCT).
  Again, the resulting equations are significantly simpler than the discrete Fourier transformation:
  \marginnote{~~\\[0.85em]This equation is implemented in \texttt{mk\_cosine\_basis}\,.}
  \begin{align}
  	C_0(k; N) &= \frac{1}{\sqrt{N}} \,, &
  	C_n(k; N) &= \frac{\sqrt{2}}{\sqrt{N}} \cos\left(\pi n \frac{k + \frac{1}2}{N}\right) \,.
  	\label{eqn:dct}
  \end{align}
  This discrete function basis is orthonormal.
  The corresponding basis transformation matrix $\mat C$ is depicted in \Cref{fig:dft_dct_basis} as well.
  \end{example}

  \paragraph{Time complexity}
  The matrices $\mat F$ and $\mat C$ can be computed in time $\mathcal{O}(q N)$; a constant number of operations is required to evaluate each cell.

  Multiplication of a vector $\vec u$ with $\mat F$ or $\mat C$ can be performed in $\mathcal{O}(q N \log(N))$.
  This is due each left and right half of $\mat F$ and $\mat C$ resembling a scaled and mirrored version of the full matrix.
  This suggests a divide and conquer algorithm if $N$ is a power of two---the Fast Fourier Transformation (FFT; \cite{cooley1965algorithm}) and the related Fast Cosine Transformation (FCT; \cite{makhoul1980fast}).
  Both algorithms can be generalised to non-power of two $N$, and the publications cited above discuss how to accomplish this.

  \begin{example}[Naive and Mean Sampled Discrete Legendre Basis]
  We can similarly apply the naive sampling (eq.~\ref{eqn:naive_sampling}) to the shifted Legendre polynomials (eq.~\ref{eqn:shifted_legendre}).
  For the sake of consistency with the bases presented in the next sections, we furthermore \enquote{mirror} the shifted Legendre polynomials, i.e., we compute $\tilde p_n(1 - x)$ instead of $\tilde p(x)$.
  We get
  \begin{align*}
  	P'_n(k; N) &= \tilde p_n\left(1 - \frac{k + \frac{1}2}N\right) \,.
  \end{align*}%
  Unfortunately, as mentioned above, the corresponding basis transformation matrix $\mat P'$ is not orthogonal.

  A slightly \enquote{more orthogonal} (in terms of the off-diagonal elements having a smaller magnitude) discrete function basis can be obtained by applying mean sampling as defined in \cref{eqn:mean_sampling}, resulting in
  \begin{align*}
  	P_n(k; N) &= \sqrt{N} \int_a^b\tilde p_n(x) \,\mathrm{d}x \,, & \text{where } a = 1 - \frac{k + 1}{N} \text{ and } b = 1 - \frac{k}{N}\,.
  \end{align*}%
  Since the $\tilde p_n$ are polynomials, the antiderivatives $\tilde P_n$ are given in closed form:
  \marginnote{~~\\[0.3em]This equation is implemented in \texttt{mk\_leg\_basis}.}
  \begin{align}
  	P_n(k; N) &= \sqrt{N} \left( \tilde P_n \left(1 - \frac{k}{N} \right) - \tilde P_n \left(1 - \frac{k + 1}{N} \right) \right) \,.
  	\label{eqn:leg}
  \end{align}
  The corresponding basis transformation matrix $\mat P$ is depicted in \Cref{fig:dft_dct_basis}.
  \end{example}

  \paragraph{Time complexity}
  Computing the basis transformation matrix $\mat P$ has a time-complexity in $\mathcal{O}(q^2 N)$---evaluating one of the $q \times N$ polynomials in $\mat P$ requires up tp $q$ multiplications using Horner's method.
  The standard run-time costs $\mathcal{O}(q N)$ for a matrix-vector multiplication apply when evaluating $\mat P \vec u$.

  \begin{example}[Discrete Haar Wavelet Basis]
  Continuous and discrete wavelet transformations are a popular alternative to Fourier-like transformations.
  The basic idea of wavelet bases is to have a single \enquote{mother} wavelet from which the individual basis functions are derived.
  In contrast to the Fourier series, wavelet bases are sparse; basis functions tend to be zero for most of the covered interval.

  One popular wavelet basis is the Haar basis $(w_n)_{n \in \mathbb{N}}$. Aside from the first basis function $w_0(x) = 1$, each $w_n(x)$ for $n \geq 2$ is a scaled and shifted version of $w_1(x)$. The complete orthonormal Haar basis over $[0, 1]$ is given as
  \marginnote{~~\\[1.1em]A discrete version of this basis can be obtained using \texttt{mk\_haar\_basis}.}
  \begin{align}
  	w_1(x) &= \begin{cases}
  		1 & \text{if } 0 \leq x < {\textstyle\frac{1}2}\,, \\
  		-1 & \text{if } \textstyle\frac{1}2 \leq x \leq 1\,, \\
  		0 & \text{otherwise} \,,
  	\end{cases} &
  	\begin{aligned}
  		&&\text{and }
		   w_n(x) = \sqrt{\phi} \, w_1 \big( \phi x - n + \phi \big) \,, \\
	  	&&\text{where } \phi = 2^{\lfloor\log_2(n)\rfloor} \,.
  	\end{aligned}
  	\label{eqn:haar_basis}
  \end{align}
  A discrete version $W_n(k; N)$ with basis transformation matrix $\mat W$ can be easily computed in $\mathcal{O}(qN)$; such a (reordered) $\mat W$ is depicted in \Cref{fig:dft_dct_basis}.
  An interesting property of this basis is that the \enquote{Fast Haar transformation} can be computed in $\mathcal{O}(N)$ \citep{kaiser1998fast}.
  This is even faster than the fast Fourier or cosine transformations, which require $\mathcal{O}(N \log(N))$ operations.
 
  \end{example}

  \clearpage
  
  \section{A Discrete Orthogonal Legendre Basis: DLOPs}

  The previous section introduced the notion of a discrete function basis.
  We saw that the cosine and Fourier series could be trivially discretised while preserving orthogonality.
  However, doing the same for the Legendre polynomials did not preserve orthogonality.

  In this section, we construct a discrete, orthogonal Legendre function basis.
  In \Cref{sec:leg_poly_linear} we translate the definition of a Legendre polynomial to discrete function basis, resulting in \enquote{Discrete Legendre Orthogonal Polynomials}, or \enquote{DLOPs} in short.
  DLOPs were originally proposed by \citet{neuman1974discrete}.
  Fortunately, Neuman and Schonbach present a simple equation that can be used to construct DLOPs.
  We review this equation in \Cref{sec:dlop}.
  We close in \Cref{sec:dlop_efficient} with the description of an efficient and numerically stable algorithm to compute DLOPs in $\mathcal{O}(qN)$.

  \subsection{Discrete Legendre Orthogonal Polynomials}
  \label{sec:leg_poly_linear}

  We can apply the definition of a Legendre polynomial (\Cref{def:legendre_polynomial}) to a discrete function basis.
  The unique basis fulfilling this definition is a discrete function basis of the Legendre polynomials in the strict sense of \Cref{def:discrete_function_basis}.\footnote{Sketch of a proof: for $N \to \infty$ the sums over $N$ turn into integrals that match the exact definition of Legendre polynomials.}

  \begin{defn}[Discrete Legendre Orthogonal Polynomials, DLOPs; adapted from \cite{neuman1974discrete}]
  DLOPs are defined as the discrete function basis $L_n(k; N)$ with the following properties
  \begin{enumerate}
    \item \emph{Polynomial:} Each $L_n(k; N)$ is a linear combination of $n$ monomials. It holds $L_n(k; N) = \sum_{j = 0}^{n} \alpha_{n, i} \left( \frac{k}{N - 1} \right)^i$ for $k \in \{0, \ldots, N - 1\}\,$.
  	\item \emph{Orthogonal:} $\sum_{k = 0}^{N - 1} L_i(k; N) L_j(k; N) = 0 \text{ exactly if } i \neq j \,$.
  	\item \emph{Normalisation:} $L_n(0; N) = \frac{1}{\sqrt{N}}\,$.
  \end{enumerate}
  \end{defn}

  \paragraph{Numerically solving for DLOPs}
  \marginnote{The function \texttt{mk\_dlop\_basis\_linsys} implements this algorithm.}
  This definition suggests a simple algorithm that can be used to construct a discrete orthogonal Legendre basis matrix $\mat L \in \mathbb{R}^{q \times N}$. We initialize the first row of $\mat L$ as ones.
  To obtain a row $n$, we solve for the polynomial coefficients $\alpha_{n, i}$ such that the above conditions are fulfilled.
  That is, the new row is orthogonal to all preceding rows and the entry in last column is equal to one.
  Finding coefficients $\alpha_{n, i}$ that fulfil these requirements is simply a matter of solving a system of linear equations.

  While this algorithm works in theory, it is numerically unstable in practice. Monomials $x^k$ with $|x| \ll 1$ and $k \gg 20$ cannot be represented well using double-precision floating point arithmetic.
  This mandates the use of arbitrary-precision rational numbers.
  Fortunately, there is no need to actually implement this algorithm since a closed-form solution exists.

  \subsection{Closed-Form Solution for DLOPs}
  \label{sec:dlop}

  Neuman and Schonbach show that the $n$th DLOP is given in closed form as
  \begin{align}
  	L_n(k; N) &=
  		\frac{1}{\sqrt{N}}\sum_{i = 0}^n (-1)^i \binom{n}{i} \binom{n + i}{i} \frac{k^{(i)}}{(N - 1)^{(i)}} \,,
  	\label{eqn:dlop_basis} \\
 		\text{where } k^{(i)} &= \prod_{j = 0}^{i - 1} (k - j) = \frac{k!}{(k - i)!} \quad \text{is the $i$th \emph{fading factorial} of $k$.}
 	\label{eqn:fading_factorial}
  \end{align} 
  \marginnote{~~\\[-0.4em]\texttt{mk\_dlop\_basis\_direct} uses arbitrary precision integers to evaluate this equation.}%
  Factoring out $(N - 1)^{(n)}$ facilitates the use of arbitrary precision integers
  \begin{align}
  	L_n(k; N) &=
  		\frac{1}{\sqrt{N}(N - 1)^{(n)}} \sum_{i = 0}^n (-1)^i \binom{n}{i} \binom{n + i}{i} k^{(i)} (N - 1 - i)^{(n - i)} \,.
  	\label{eqn:dlop_basis_norm}
  \end{align}
  \begin{figure}%
  	\centering
  	\includegraphics{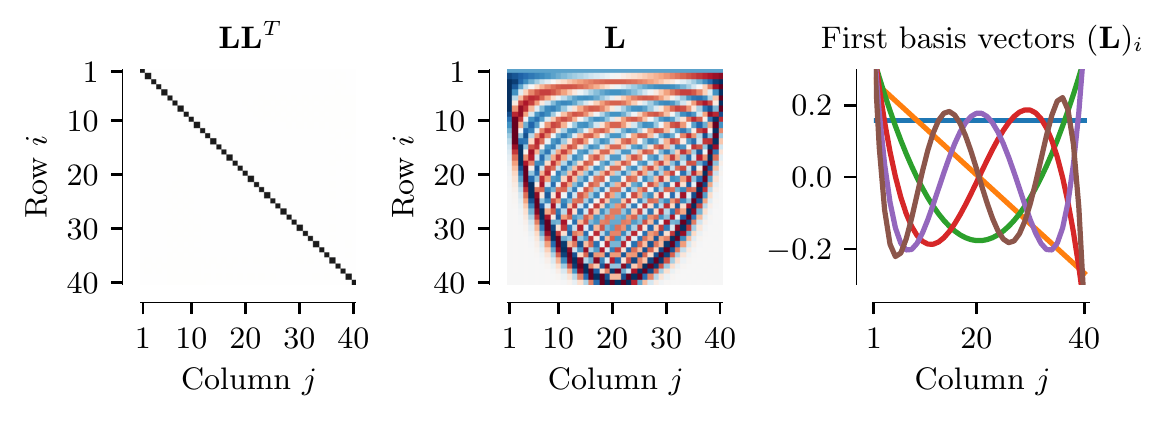} 
  	\caption{Visualisation of the DLOP basis defined in \cref{eqn:dlop_basis} for $q = N = 40$. See \cref{fig:dft_dct_basis} for the complete legend and a description of the colour scheme.}
  	\label{fig:dlop_basis}
  \end{figure}%
  The corresponding normalised matrix $\mat L \in \mathbb{R}^{q \times N}$ for $q = N = 40$ is depicted in \Cref{fig:dlop_basis}.
  Notice that the resulting matrix is perfectly orthogonal.
  Comparing the DLOP matrix $\mat L$ to the naive discrete Legendre basis $\mat P$ (cf.~\cref{fig:dft_dct_basis}), we find that the two bases are strikingly different for higher-order terms. The last rows in $\mat L$ have many near-zero entries with non-zero values centred around $k = N / 2$.

  The time-complexity of evaluating the corresponding basis transformation matrix $\mat L \in \mathbb{R}^{q \times N}$ using \cref{eqn:dlop_basis} in $\mathcal{O}(q^2 N)$.

  \subsection{Efficiently Computing DLOP Coefficients}
  \label{sec:dlop_efficient}

  As noted above, the time-complexity of evaluating \cref{eqn:dlop_basis} is in $\mathcal{O}(q^2 N)$.
  Furthermore, the equation can only evaluated reliably using arbitrary precision integers.
  \Citet{neuman1974discrete} propose an $\mathcal{O}(q N)$ algorithm that relies on a variant of the recurrence relation from \cref{eqn:leg_rec}.
  In this section, we discuss a version of this algorithm that generates an orthonormal matrix $\mat L$ using standard double-precision floating point arithmetic.

  \newpage

  \marginnote{~~\\[0.75em]The function \texttt{mk\_dlop\_basis\_\\recurrence} implements this particular equation, using an algorithm proposed by Neuman and Schonbach.}
  \noindent The discrete Legendre recurrence relation presented in the paper is
  \begin{align*}
  	L_{0}(k; N) &= \frac{1}{\sqrt{N}} \,,\\
  	L_{1}(k; N) &= \frac{1}{\sqrt{N}} \,\frac{(2 k - N + 1)}{N - 1} \,, \\
  	L_{n}(k; N) &= \quad\, L_{n - 1}(k; N) \frac{(2n - 1) (N - 2k - 1)}{n (N - n)} \\
  	                      &\hphantom{{}={}} - L_{n - 2}(k; N) \frac{(n - 1) (N + n - 1)}{n (N - n)} \,.
  \end{align*}
  Naively evaluating this recurrence numerically is not stable for $n > 40$.
  Some of the columns $k$ grow exponentially in magnitude with $n$.

  For a numerically stable algorithm we suggest to ensure that each $L_{n}(k; N)$ is normalised. This normalised discrete function basis $L'_{n}(k; N)$ has the property
  \begin{align*}
  	\sum_{k = 0}^{N - 1} L'_{n}(k; N)^2 = \sum_{k = 0}^{N - 1} \left( \sqrt{\alpha_n(N)} L_{n}(k; N) \right)^2 = 1\,,
  \end{align*}
  i.e., it is orthonormal. According to Neuman and Schonbach (p.~746), the normalisation factor $\alpha_n(N)$ is
  \begin{align*}
  	\alpha_n(N) &= \frac{(2n + 1) (N - 1)^{(n)}}{(N + n)^{(n + 1)}} \,,
  \end{align*}
  where $k^{(i)}$ is the $i$th fading factorial of $k$, as defined in \cref{eqn:fading_factorial}.
  Applying the normalisation, we get the following recurrence relation for $L'_{n}$:
  \marginnote{~~\\[1.75em]The function \texttt{mk\_dlop\_basis} implements this set of equations and addresses the numerical instability mentioned below.}
  \begin{align}
  	\begin{aligned}
  	L'_{0}(k; N) &= \frac{1}{\sqrt{N}} \,,\\
  	L'_{1}(k; N) &= \frac{(2 k - N + 1)}{N - 1} \sqrt{\frac{3 (N - 1)}{N (N + 1)}} \,, \\
  	L'_{n}(k; N) &= \quad\, L'_{n - 1}(k; N) \frac{(2n - 1) (N - 2k - 1)}{n (N - n)} \sqrt{\frac{\alpha_{n}(N)}{\alpha_{n - 1}(N)}} \\
  	                      &\hphantom{{}={}} - L'_{n - 2}(k; N) \frac{(n - 1) (N + n - 1)}{n (N - n)} \sqrt{\frac{\alpha_{n}(N)}{\alpha_{n - 2}(N)}} \,.
    \end{aligned}
    \label{eqn:dlop_basis_rec}
  \end{align}
  Multiplying with the square root of $\alpha_n(N)$ applies the normalisation, dividing by the square roots of $\alpha_{n - 1}(N)$ and $\alpha_{n - 2}(N)$ reverts the normalisation applied to the lower-order discrete basis function.
  These fractions can be simplified to
  \begin{align*}
  	\frac{\alpha_{n}(N)}{\alpha_{n - 1}(N)} &= \frac{(2n + 1)(N - n)}{(2n - 1) (N + n)} \,,
    & \frac{\alpha_{n}(N)}{\alpha_{n - 2}(N)} &= \frac{(2n + 1) (N - n) (N - n + 1)}{(2n - 3) (N + n) (N + n - 1)} \,.
  \end{align*}

  \paragraph{Numerical stability}
  Applications requiring utmost numerical robustness should use \cref{eqn:dlop_basis_norm} with arbitrary precision integers and subsequent normalisation; the recurrence relation in \cref{eqn:dlop_basis_rec} inadvertently propagates numerical errors.

  In particular, special care must be taken when implementing \cref{eqn:dlop_basis_rec}.
  Whenever a cell in column $k$ is close to zero in two consecutive rows $n - 1$, $n$, all consecutive cells in column $k$ of rows $n + i$ must be zero as well.
  Ensuring this is important, since small non-zero values caused by numerical instabilities can rebound exponentially when applying the recurrence relation.

  \begin{figure}
  	\centering
  	\includegraphics{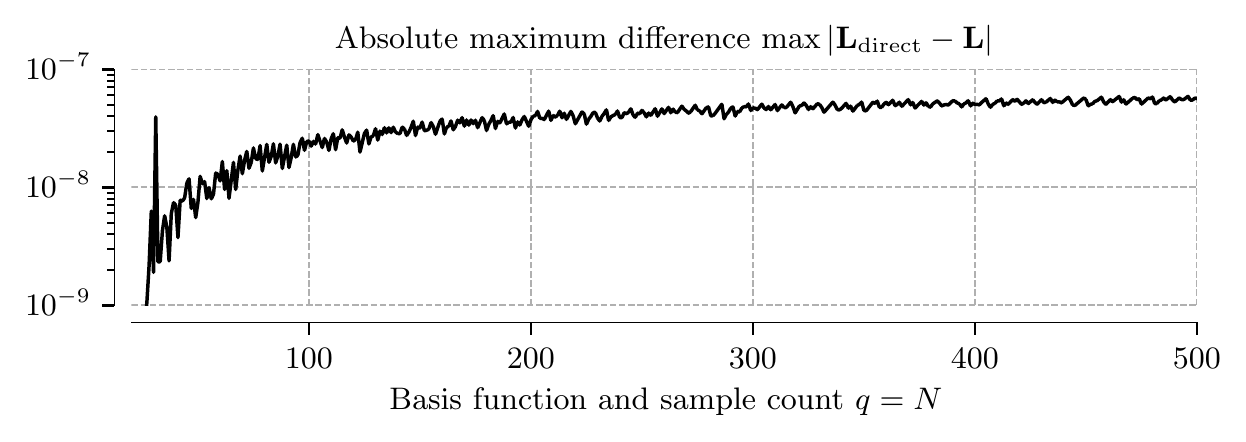}
  	\caption{Comparing the basis transformation matrix $\mat L$ obtained when evaluating the recurrence relation $\mat L$ to the matrix $\mat L_\mathrm{direct}$ obtained when evaluating the closed-form equation \cref{eqn:dlop_basis}. Even for large $q$, errors do not exceed $10^{-7}$.}
  	\label{fig:dlop_numerical_stability}
  \end{figure}
  \Cref{fig:dlop_numerical_stability} shows the maximum absolute difference between the matrix $\mat L_\mathrm{direct}$ obtained when using \cref{eqn:dlop_basis_norm} and $\mat L$ computed using our proposed recurrence relation in \cref{eqn:dlop_basis_rec}.
  Errors do not exceed $10^{-7}$, even for $q = 500$.

  \section{Constructing a Discrete LDN Basis}
  \label{sec:ldn_basis}

  In this section, we focus on the Legendre Delay Network (LDN) and the corresponding basis transformation matrix $\mat H \in \mathbb{R}^{q \times N}$.
  We first review the Linear Time Invariant (LTI) system underlying the LDN and observe that the impulse response of the LDN system resembles a continuous Legendre function basis over the interval $[0, \theta]$.
  The impulse response sharply decays to zero for $t > \theta$, that is, the system has an \emph{almost} finite impulse response.
  Second, as originally proposed by Chilkuri, we construct the matrix $\mat H$ to approximate the impulse response.

  We close by discussing why the LDN system may be particularly useful.
  To summarize, since $\mat H$ was derived from an LTI system with an (almost) finite impulse response, we can either use the basis transformation matrix $\mat H$ (i.e., a set of FIR filters) \emph{or} the LTI system itself to compute generalised Fourier coefficients $\vec m$ of an input signal $\vec u_t$.
  The FIR filter representation is useful when training neural networks; during inference the LDN LTI system can directly be used as a fast \enquote{sliding transformation}.
  In contrast to other sliding transformations the LDN LTI system requires a minimal amount of state memory.
  
  \subsection{Review: The LDN System}
  The LDN system (\cite{voelker2018improving}) can be thought of as continuously compressing a $\theta$ second long time-window of a function $u(t)$ into a $q$-dimensional vector $\vec m(t)$.
  The system is derived from the Padé approximants \citep{baker2012pade} of a Laplace-domain delay $e^{-s\theta}$, along with a set of transformations that make the system numerically stable.
  Let $\mat A \in \mathbb{R}^{q \times q}$, $\mat B \in \mathbb{R}^{q \times 1}$.
  Then, the LDN system is given as
  \marginnote{~~\\~~\\The function \texttt{mk\_ldn\_lti} generates the LDN system matrices $\mat A$, $\mat B$.}%
  \begin{align}
  	\frac{d\theta \vec m(t)}{dt} &= \mat A \vec m(t) + \mat B u(t) \,, \notag\\
 	\big(\mat A\big)_{ij} &= (2i + 1)
 	\begin{cases}
 		-1 & \text{if } i \leq j \,,\\
 		(-1)^{i - j + 1} & \text{if } i > j \,,
 	\end{cases} &
 	\big(\mat B\big)_{ij} &= (2i + 1) (-1)^{i} \,.
 	\label{eqn:ldn}
  \end{align}%
  As discovered by \citet[Section 6.1.3, p.~134]{voelker2019}, the normalised impulse response  $\vec{\tilde m}^q(t)$ of this system over a time window $[0, \theta]$ resembles the first $q$ shifted Legendre polynomials scaled to the interval $[0, \theta]$.
  Judging from numerical evidence, it seems reasonable to assume that for $q \to \infty$ the impulse response and the Legendre polynomials are exactly equal.
  Correspondingly, $(\tilde m^q_n(t))_{n \in \mathbb{N}}$ for $q \to \infty$ forms a function basis over the interval $[0, \theta]$.
  While empirical evidence suggests that this is true, we do \emph{not} have a rigorous proof for this.

  \begin{figure}[p]
  	\centering
  	\includegraphics{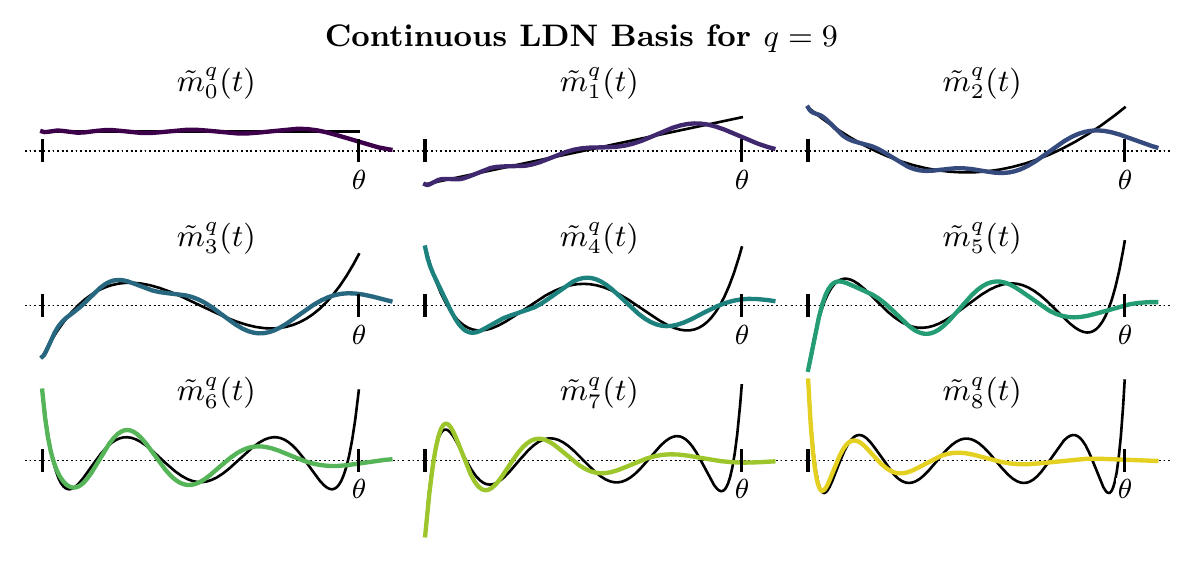}\\[0.125cm]
  	\includegraphics{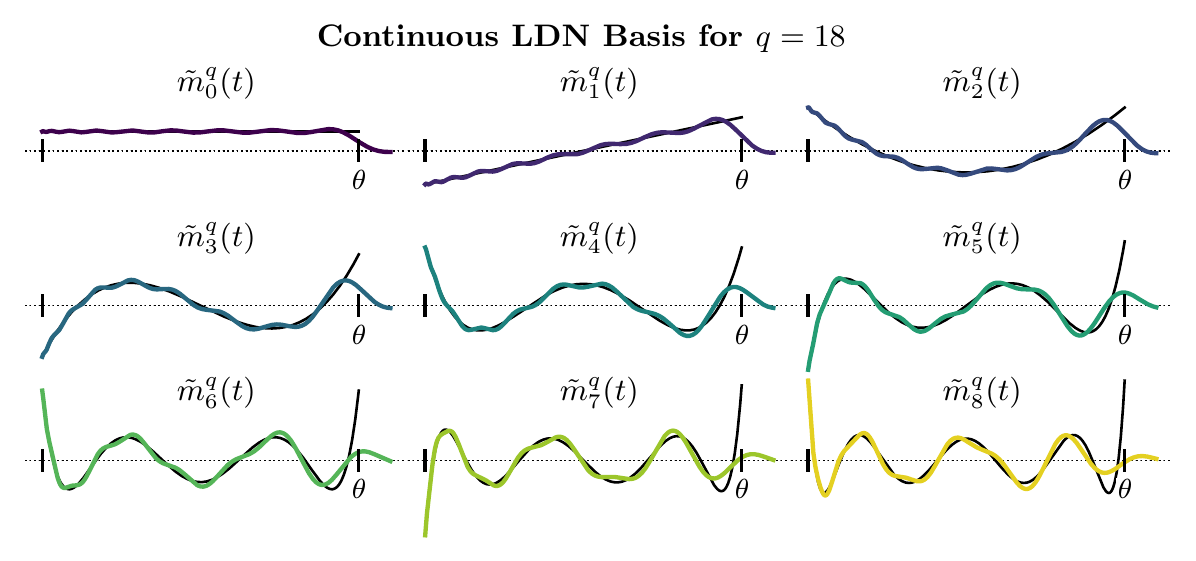}\\[0.125cm]
  	\includegraphics{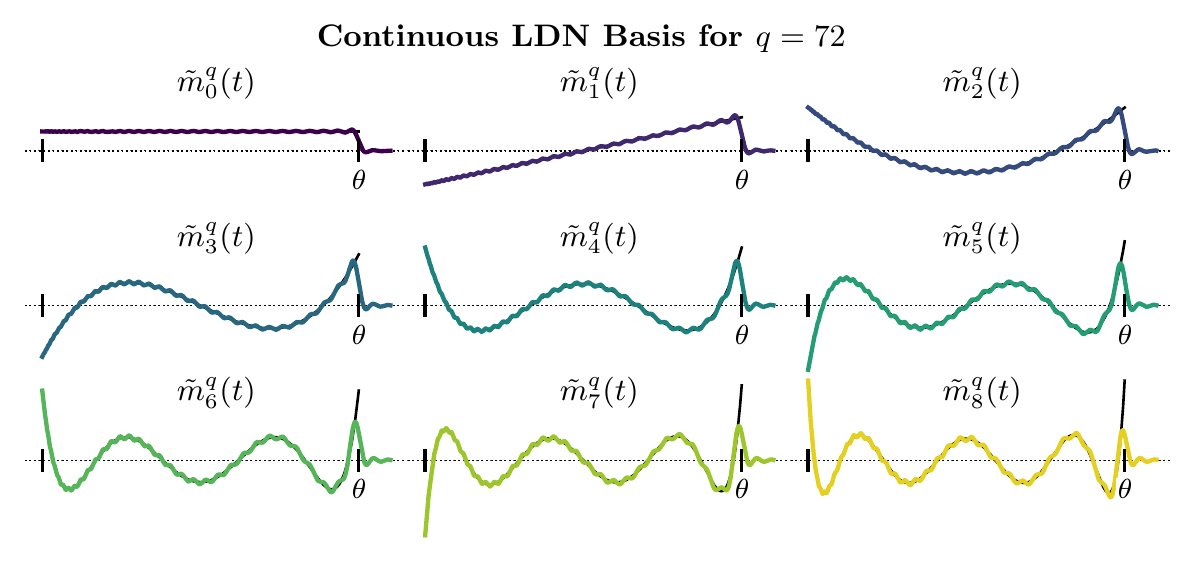}
  	\caption{Normalised impulse response of the first six state dimensions of the LDN as defined in \cref{eqn:ldn_impulse_response} for different state dimensionalities $q$ (coloured lines).
  	Solid black lines correspond to the orthonormal Legendre basis $p_n$. As $q$ increases, the impulse response more closely resembles the Legendre basis.
  	}
  	\label{fig:delay_network_impulse_response}
  \end{figure}%
  Mathematically, the impulse response $\vec{\tilde m}^q(t)$ of the normalised LDN system and the presumed relationship to the Legendre polynomials is given as
  \begin{align}
  	\tilde m^q_n(t) &= \frac{\sqrt{\theta}}{\sqrt{2 n + 1}} e^{\mat A t} \vec B = \frac{1}{\sqrt{\theta}} p_n\left(\frac{t}{\theta}\right) \,, \quad \text{where } t \in [0, \theta], \text{ for } q \to \infty \,.
  	\label{eqn:ldn_impulse_response}
  \end{align}
  $\mat A \in \mathbb{R}^{q \times q}$ and $\mat B \in \mathbb{R}^{q \times 1}$ are as defined in \cref{eqn:ldn}; the orthonormal Legendre polynomial $p_n$ is as defined in \cref{eqn:legendre_basis} with the re-scaling from \cref{eqn:rescale} applied.

  The LDN impulse response and the corresponding Legendre polynomials are depicted in \Cref{fig:delay_network_impulse_response}.
  Two observations are worth being pointed out.
  
  First, notice how the impulse response of the LDN system sharply converges to zero for $t > \theta$.
  In other words, the LDN system has no memory of anything happening more than $\theta$ seconds ago.
  This is one of the key properties of the LDN system, and we revisit this in the next section, when we discuss how to construct LTI systems from discrete function bases (i.e., the inverse of what we are doing in this section).
  
  Second, for finite $q$, the implicit basis created by the LDN system is \emph{not exactly} the Legendre basis, but an \emph{approximation}. We refer to the finite sequence of $q$ functions generated by the LDN as the LDN \enquote{basis}, although, technically, a finite sequence of functions cannot form a continuous function basis.

  \subsection{Constructing the LDN Discrete Function Basis}
  \label{sec:construct_ldn_basis}

  To compute $\mat H$, we could just apply the \enquote{mean sampling} discussed in \Cref{xpl:mean_sampling} to the impulse responses $\tilde m_n^q(t)$.
  In fact, this is exactly what we will end up doing.
  However, this was not how we derived $\mat H$ in the first place, and we find it considerably more instructive to discuss our original derivation. We then prove equivalence of the resulting expression to mean sampling.
  Impatient readers interested in the main result may wish to skip ahead to the end of this subsection.

  \paragraph{Compressing signals using the LDN}
  \begin{figure}[t]
  	\centering
  	\includegraphics{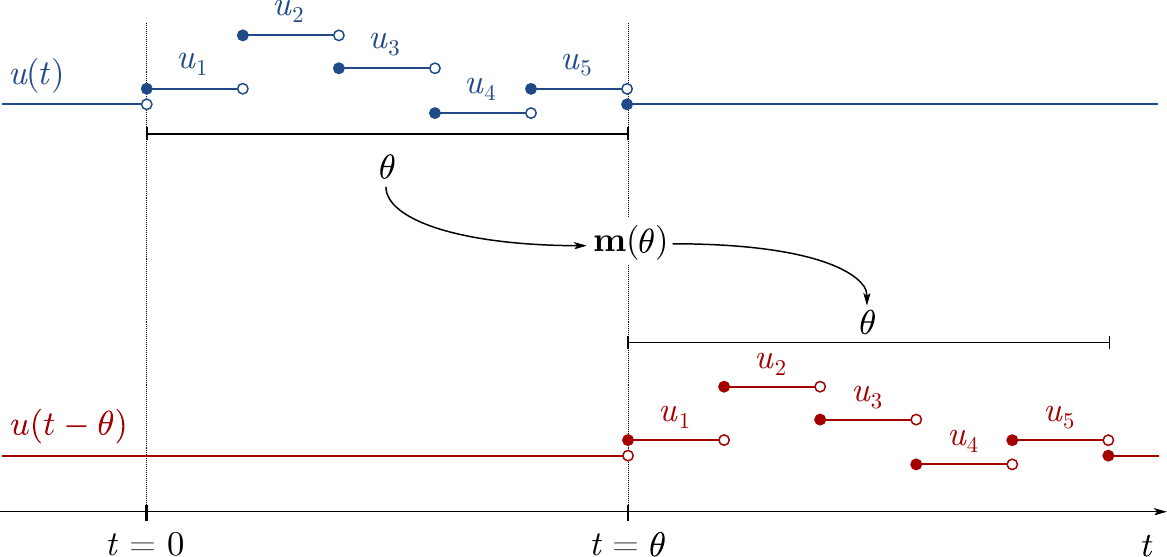}
  	\caption{Diagram illustrating a perfect delay. A stair-step function $u(t)$ representing $N = 5$ values $u_1, \ldots, u_5$ between $t = 0$ and $t = \theta$ is delayed by exactly $\theta$ seconds, resulting in $u(t - \theta)$. When implementing the delay, the LDN system must represent information about the last $\theta$ seconds in its state vector $\vec m(t)$. In this example, all samples \emph{must} be represented in $\vec m(\theta)$.}
  	\label{fig:delay_network_compression}
  \end{figure}
  Feeding a signal $u(t)$ into the Legendre Delay Network allows us, as the name suggests, to decode a delayed signal $u(t - \theta)$ from its state vector $\vec m(t)$.
  Consider what happens if we present $N$ samples $u_1, \ldots, u_N$ to the LDN system over the time-window $\theta$, for example using a stair-step function
  \begin{align}
  	u(t) &= \begin{cases}u_{i} & \text{if } 0 \leq t < \theta \,, \\ 0 & \text{otherwise,}\end{cases} & \quad \text{ where } i &=  1 + \left\lfloor \frac{N t}{\theta} \right\rfloor \,.
  	\label{eqn:stairstep}
  \end{align}
  At time $t = \theta$ all samples have been presented to the LDN.
  If the LDN were to implement a perfect delay, we would decode $u(t - \theta)$, which is equal to the first sample $u_1$.
  A bit later, at time $t = \frac{N + 1}N \theta$, the network would output $u_2$, and so on (cf.~\cref{fig:delay_network_compression}).
  This means that at $t = \theta$, the network has \enquote{compressed} all $N$ samples into its $q$-dimensional state vector $\vec m(t)$.
  Of course, the LDN system acts as a basis function transformation that represents $\vec u = (u_1, \ldots, u_N)$ as a vector $\vec m(\theta) = (m_1(\theta), \ldots, m_q(\theta))$ with respect to a discrete function basis.

  Our goal is to find a linear expression mapping $\vec u$ onto $\vec m(\theta)$. Given such an expression, we could extract the corresponding basis transformation matrix $\mat H$.
  
  \begin{figure}
  	\centering
  	\includegraphics[trim=0cm 0.25cm 0cm 0.25cm,clip]{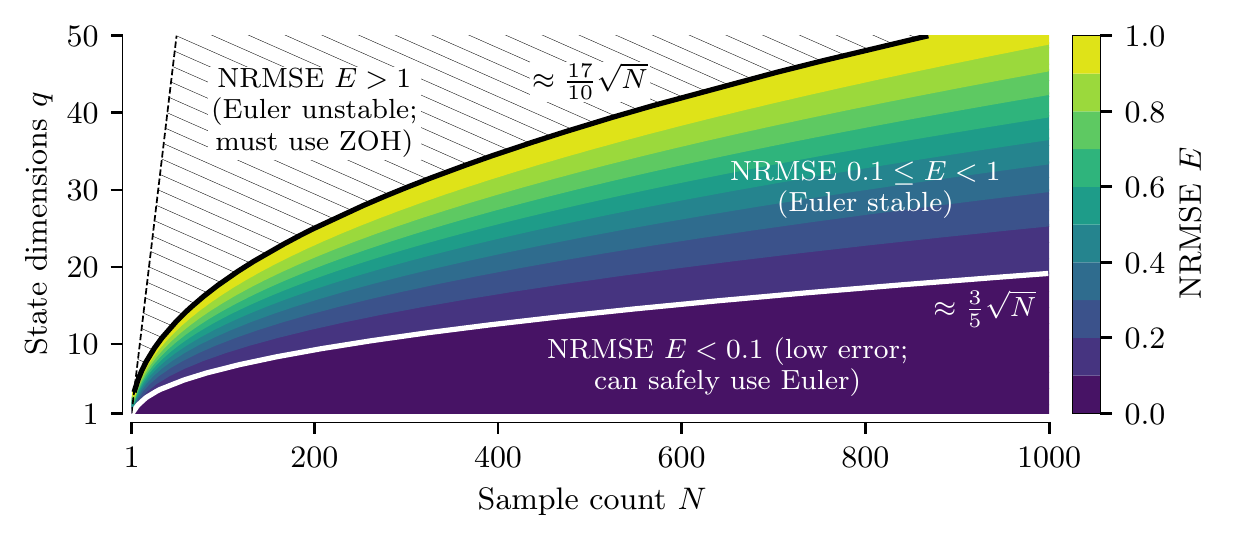}
  	\caption{Normalised RMSE between the matrix $\mat H$ computed using zero-order hold discretisation (eq.~\ref{eqn:ldn_h_matrix}) and a $\mat H'$ computed using Euler's method (eq.~\ref{eqn:ldn_euler}).
  	For Euler's method to yield NRMSEs below 0.1, $N$ must be larger than $\frac{25}{9} q^2$.}
  	\label{fig:ldn_zoh_vs_euler}
  \end{figure}

  \paragraph{Naive Euler recurrence relation}
  In a first step, let us coarsely discretise the so far continuous functions. Let $\theta = N\Delta t$, where $\Delta t$ is the timestep.
  In this case, each sample $u_i$ will be presented for exactly one timestep, and we need to derive an expression for the state vector at timestep $N$, i.e., $\vec m_N$.

  \marginnote{~~\\[0.25em]This equation is implemented in \texttt{mk\_ldn\_basis\_\\euler}.}%
  Let $\vec m_0 = 0$. Using Euler integration, we get the recurrence relation
  \begin{align}
  	\vec m_{i + 1}
  		&= \vec m_{i} + \Delta t \left( \frac{\mat A\vec m_{i} }{\theta} + \frac{\mat B}{\theta} u_i  \right)
  	     = \frac{1}N \big( ( \mat A  + N \mat I ) \vec m_i + \mat B u_i \big) \,.
  	\label{eqn:ldn_euler}
  \end{align}%
  Unfortunately, using an Euler integrator in this manner without finer-grained update steps is generally a bad idea.
  Judging from numerical experiments (\cref{fig:ldn_zoh_vs_euler}), it must approximately hold $N > 2.78q^2$ for Euler's method to not introduce large errors.
  If $N < 0.35q^2$, this method will diverge within the first $N$ timesteps.

  \newcommand{\mA}{\mat{\tilde A}}%
  \newcommand{\mB}{\mat{\tilde B}}%
  \paragraph{Closed-form solution with zero-order hold assumption}
  There is a relatively simple solution to this problem that is in the spirit of the above idea.
  Since our system is purely linear, we can advance the system $\Delta t$ seconds into the future given an initial state $\vec m(t)$, under the condition that $u(t)$ stays constant for the next $\Delta t$ seconds.
  This is exactly how we defined $u(t)$ in \cref{eqn:stairstep} if $\theta = N \Delta t$. In general, assuming that $u(t)$ is a stair-step function is called a \enquote{zero-order hold assumption}.
  We obtain a new recurrence relation that uses a matrix exponential
  \begin{align}
    \begin{aligned}
  	{\vec m}_{i}    &= \mA {\vec m}_{i - 1} + \mB u_{i} \,, \\
  	\text{where } \mA &= \exp\left(\Delta t \frac{\mat A}{\theta}\right) = \exp\left(\frac{\mat A}{N}\right) \,, \\
  	              \mB &= {\mat A}^{-1} (\mA - \mat I) \mat B \,.
  	\end{aligned}
  	\label{eqn:ldn_recurrence_update}
  \end{align}
  \marginnote{~~\\[-7.75em]This equation is implemented in the function \texttt{discretize\_lti}.}%
  This is a standard technique for the discretisation of LTI systems \citep[cf.][Section 5.1.1, p.~100]{voelker2019}.
  Expanding the recurrence relation for ${\vec m}_{N}$ we get
  \begin{align}
    \vec{m}_{N}
    	&= \mA \vec{m}_{N - 1} + \mB u_{N} \notag\\
    	&= \mA \big( \mA \vec{m}_{N - 2} + \mB u_{N - 1} \big) + \mB u_{N} \notag\\
    	&= \mA^2 \vec{m}_{N - 2} + \mA \mB u_{N - 1} + \mB u_{N} \notag \\
    	&= \mA^3 \vec{m}_{N - 3} + \mA^2 \mB u_{N - 2} + \mA \mB u_{N - 1} + \mB u_{N} \notag \\
    	&= \ldots \notag \\
    	&= \sum_{i = 1}^N \mA^{N - i} \mB u_i =  \sum_{i = 0}^{N - 1} \mA^i \mB u_{N - i} \,.
    \label{eqn:ldn_recurrence}
  \end{align}%
  \begin{figure}
  	\centering
  	\includegraphics{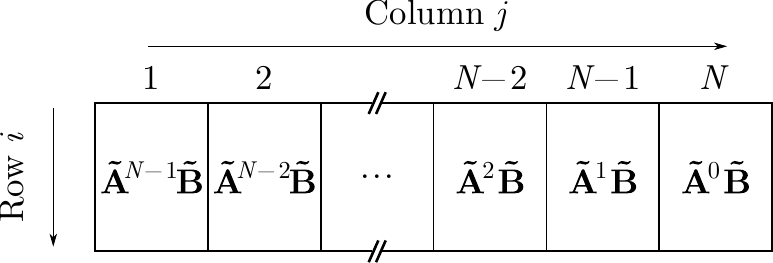}
  	\caption{Illustration of the unnormalised LDN basis transformation matrix $\mat H'$ (eq.~\ref{eqn:ldn_h_matrix}). The same principle can be applied to other LTI systems.}
  	\label{fig:matrix_H_construction}
  \end{figure}%
  As desired, this is a linear equation. We can write this sum in terms of a matrix-vector product $\mat H' \vec u$, where $\mat H' \in  \mathbb{R}^{q \times N}$ is an unnormalised basis transformation matrix. The $k$th column of $\mat H'$, denoted $\big(\mat H'^T\big)_k$ is simply given as (cf.~\cref{fig:matrix_H_construction})
  \begin{align}
  	\big(\mat H'^T\big)_k &= \mA^{N - k} \mB \,. \label{eqn:ldn_h_matrix}
  \end{align}
  The time-complexity of evaluating this matrix is in $\mathcal{O}(q^2 N)$.
  Importantly, and in contrast to all other discrete function bases discussed so far, the corresponding discrete function basis $H'^q_n(k; N) = \mat H'_{n, k}$ depends on the state-dimensionality $q$.
  In other words, when $q$ is changed, all $q$ discrete basis functions change; in general, $H^q_n(k; N) \neq H^{q'}_n(k; N)$.

  \paragraph{Qualitative comparison to the Legendre and DLOP bases}
  \begin{figure}
  	\centering
  	\includegraphics{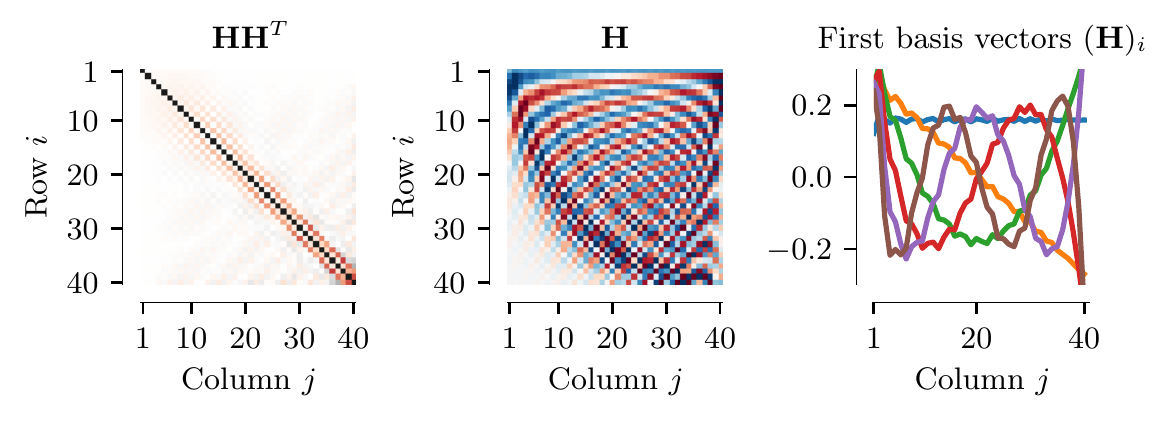} 
  	\caption{Visualisation of the LDN basis as defined in \cref{eqn:ldn_h_matrix} for $q = N = 40$. See \Cref{fig:dft_dct_basis} for the complete legend and a description of the colour scheme.}
  	\label{fig:ldn_basis}
  \end{figure}
  The normalised function basis transformation matrix $\mat H$ is depicted in \Cref{fig:ldn_basis}. $\mat H$ is \emph{almost} orthogonal and is similar to both the DLOP and the Legendre basis in some respects.
  The left half of $\mat H$ somewhat resembles DLOPs, whereas the right half is very similar to the discrete Legendre basis from \cref{eqn:leg}.

  \paragraph{Equivalence to the mean-sampled impulse response}
  Applying mean sampling as defined in \cref{eqn:mean_sampling} to the LDN system impulse response for $\theta = 1$
  \begin{align*}
  	H^q_n(k; N) &= \frac{\sqrt{N \theta}}{\sqrt{2 n + 1}} \int_a^b \big( e^{\mat A t} \mat B \big)_n \,\mathrm{d}t \,, \quad \text{where } a = \frac{k}{N}, \quad b = \frac{k + 1}{N} \,.
  \end{align*}
  The integral of a matrix exponential is \citep[cf.][pp.~171-172]{derusso1998state}
  \begin{align*}
  	\int_{0}^b e^{\mat A t} \,\mathrm{d}t = \big( e^{\mat A b} - \mat I \big) {\mat A}^{-1} 
  	\quad\Rightarrow\quad
  	\int_{a}^b e^{\mat A t} \,\mathrm{d}t = \big( e^{\mat A b} - e^{\mat A a} \big) {\mat A}^{-1} \,.
  \end{align*}
  Abbreviating the normalisation term as $\gamma$ and factoring out $\mat B$ we get
  \begin{align*}
  	H^q_n(k; N)
  		&= \gamma \Big( \Big( \int_a^b e^{\mat A t} \,\mathrm{d}t \Big) \mat B \Big)_n
  		 = \gamma \Big( \Big(  e^{\mat A b} - e^{\mat A a} \Big) \mat A^{-1}  \mat B \Big)_n \\
  		&= \gamma \Big( \Big(  \mA^{k + 1} - \mA^{k} \big) \mat A^{-1} \mat B \Big)_n
  		 = \gamma \Big( \Big( \mA^k \big( \mA - \mat I \big) \mat A^{-1}  \mat B \Big)_n \,.
  \end{align*}
  The exponential of matrix $e^{\alpha \mat A}$ and its inverse $\mat A^{-1}$ are commutative \citep[cf.][p.~170 for the definition of the matrix exponential]{derusso1998state}
  \begin{align*}
  	\mat A^{-1} e^{\alpha\mat A} &= \mat A^{-1} \sum_{i=0}^\infty \frac{\mat A^i \alpha^i}{i!}
  		= \sum_{i=0}^\infty \frac{\mat A^{i - 1} \alpha^i}{i!}
  		= \sum_{i=0}^\infty \frac{\mat A^i \alpha^i}{i!} \mat A^{-1}
  		= e^{\alpha \mat A} \mat A^{-1}\,.
  \end{align*}
  Hence, we can move the $\mat A^{-1}$ to the left-hand side of the term $\mA - \mat I$. We get
  \begin{align}
  	H^q_n(k; N)
  		&= \gamma \Big( \Big( \mA^k \mat A^{-1} \big( \mA - \mat I \big) \mat B \Big)_n
  		 = \gamma \Big( \mA^k \mB \Big)_n \,.
 	\label{eqn:ldn_basis}
  \end{align}
  Scaling and ordering ($k$ and not $N - k$, cf.~eq.~\ref{eqn:fir}) aside, this is exactly \cref{eqn:ldn_h_matrix}.

  \subsection{When to Use the LDN Basis}
  \label{sec:ldn_when}
  Up to this point, it may seem as if the LDN was merely a convoluted way to construct a discrete function basis that resembles the Legendre polynomials.
  So, why---compelling connections to biology aside \citep{voelker2018improving}---should we care about the LDN system at all, and not just use exactly orthogonal discrete function basis such as DLOPs?

  The answer to this is not clear-cut.
  The gist is that the LDN is the optimal online, zero-delay (or \enquote{sliding}) transformation that weighs each point in time equally and only requires $\mathcal{O}(q)$ \emph{state} memory \citep[see][for the \enquote{equal weight} aspect]{gu2020hippo}.
  Still, there are some trade-offs that are worth discussing.

  \paragraph{Sliding transformations}
  As mentioned at the beginning of this section, one way to think about the LDN system is as a means to convolve an input signal $u(t)$ with the Legendre polynomials at every point in time $t$, or, in other words, to compute the generalised Fourier coefficients $\xi_n(t)$ \emph{online}; i.e.,
  \begin{align*}  	
  	\xi_n(t) &= \big\langle u_{[t - \theta, t]}, \tilde m^q_n \big\rangle \,,
  \end{align*}
  where $u_{[t - \theta, t]} : [0, \theta] \longrightarrow \mathbb{R}$ corresponds to a function representing the past $\theta$ seconds of the input $u$, and $\tilde m^q_n$ is as defined in \cref{eqn:ldn_impulse_response}.
  That is, by simply advancing a $q$-dimensional LTI system for an input $u(t)$, the generalised Fourier coefficients are stored in the momentary LTI system state $\vec m(t)$.
  
  A transformation that is evaluated at every point in time over a window of the input history is also called a \emph{sliding transformation}.
  Most of the discrete transformations we discussed so far have sliding versions; examples being the sliding discrete Fourier (SDFT; \cite{jacobsen2003sliding}), cosine (SCT; \cite{kober2004fast}) and Haar transformations \citep{kaiser1998fast}.
  These \enquote{classic} sliding transformations mandate that a window $u_{[t - \theta, t]}$ is kept in memory, i.e., $\mathcal{O}(N)$ memory in the discrete case.
  Each update step requires only $\mathcal{O}(q)$ operations.

  \paragraph{Efficiency of the LDN compared to FIR filters}
  In contrast, the discrete LDN LTI system only requires $\mathcal{O}(q)$ \emph{state} memory (in addition to storing the $q \times q$ matrix $\mA$) and can be advanced using \cref{eqn:ldn_recurrence_update}.
  Each update requires $\mathcal{O}(q^2)$ operations and yields the updated discrete generalised Fourier coefficients $\vec m_t$.

  For bases where we do not have an efficient sliding transformation (as, for example, for DLOPs), we must treat the basis transformation matrix $\mat E$ as a set of $q$ FIR filters (cf.~eq.~\ref{eqn:fir}).
  Filtering a signal $\vec u$ with $q$ FIR filters requires holding the past $N$ input samples in memory in addition to the $q \times N$ matrix $\mat E$.
  When done naively, convolution with the filters requires $\mathcal{O}(q N)$ operations in every timestep.
  Hence, updating the discrete LDN system is more efficient than repeated convolution if $q < N$.

  \paragraph{Efficient zero-delay FIR filtering}
  The above characterization is a bit misleading.
  The time complexity of $\mathcal{O}(q N)$ for online convolution of a signal with a set of $q$ FIR filters only applies to the naive algorithm.
  In general, this operation can be performed using $\mathcal{O}(q \log(N)
  )$ operations and $\mathcal{O}(N \log(N))$ memory \Citep[amortized; cf.][]{gardner1995efficient}.
  Note that the corresponding algorithm has a relatively large constant scaling factor of approximately~$34$ in the number of operations (cf.~Section 5.2, p.~132 of \cite{gardner1995efficient}).

  This means that---much lower memory requirements aside---using \cref{eqn:ldn_recurrence_update} to compute the discrete generalised Fourier coefficients $\vec m_t$ of the LDN basis is still attractive when compressing a large number of samples $N$ into relatively few dimensions $q$.
  To be precise, using an LTI system to implement a sliding transformation is more efficient as long as $q < 34 \log_2(N)$. This relationship is depicted in \Cref{fig:ldn_vs_fir_complexity}.
  As a rule of thumb, and assuming that memory is not a constraint, FIR filters should be used if $N \geq q > 300$.

  \begin{figure}[t]
  	\centering
  	\includegraphics[trim=0.25cm 0.25cm 0.25cm 0.25cm,clip]{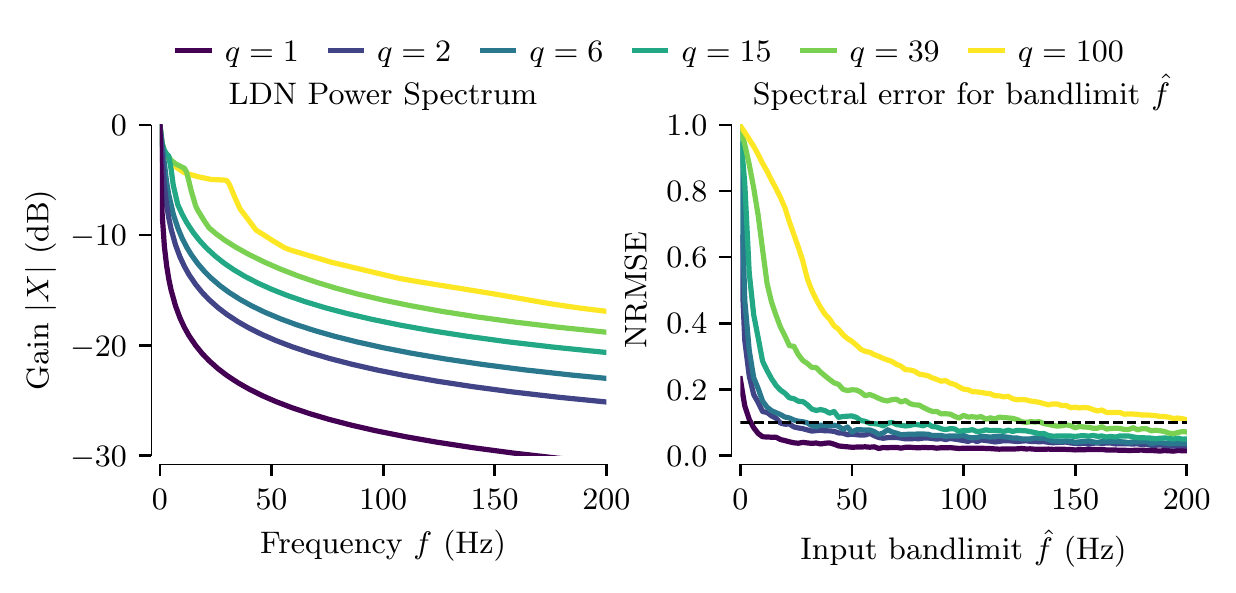}
  	\caption{LDN impulse response power spectrum and low-pass characteristics.
  	\emph{Left:} Power spectrum for different $q$ assuming $\theta = 1\,\mathrm{s}$; the LDN system acts as a low-pass.
  	\emph{Right:} Error $\mat H \vec u - \mat H \vec{\hat u}$ for a white-noise signal $\vec u$ and a band-limited version $\vec{\hat u}$ thereof with band-limit $\hat f$. It must approximately hold $\hat f > 2 q$ for the NRMSE to not surpass $0.1$ (dashed line).}
  	\label{fig:ldn_spectrum}
  \end{figure}

  \paragraph{Fast Euler update}
  Requiring $\mathcal{O}(q^2)$ operations per update step and $\mathcal{O}(q^2)$ of memory in total for the zero-order hold discrete LDN system may be a little disappointing.
  After all, the standard sliding transformations mentioned above only require $\mathcal{O}(q)$ operations per update steps in exchange for $\mathcal{O}(N)$ memory.

  In fact, \citet[Figure~6.6]{voelker2019} shows that the LDN can be advanced using only $\mathcal{O}(q)$ operations per step and $\mathcal{O}(q)$ \emph{total} memory when using the Euler update from (cf.~eq.~\ref{eqn:ldn_euler}).
  This low memory requirement can be seen as the \emph{defining property of the LDN}.
  After all, the LDN was derived from first principles to represent a window of data using $q$ state variables over time.

  \paragraph{Limitations of the fast Euler update}
  While the low memory requirements of the Euler method can be important, the time-complexity $\mathcal{O}(q)$ comes with a caveat.
  Remember from above that the Euler update is only feasible approximately if $N > 2.78 q^2$.
  Consequently, and as we will explain in the following, both the $\mathcal{O}(q^2)$ zero-order hold algorithm and the $\mathcal{O}(q)$ Euler update have the same asymptotic time-complexity in practice under the condition that the input signal $u(t)$ is appropriately band-limited and sampled near the Nyqist limit.\footnote{Down-sampling to the band-limit introduces latency, which is not desirable in real-time control applications; the $\mathcal{O}(q)$ Euler update is optimal in this case.}

  To see this, consider a $q$-dimensional LDN.
  As depicted in the left half of \Cref{fig:ldn_spectrum}, the LDN system acts as a low-pass filter.
  That is, higher frequencies in the input signal barely have an influence on the output of the system.
  The high frequencies can even be filtered out completely without changing the generalised LDN Fourier coefficients (the \enquote{LDN spectrum}) much.

  This is depicted in the right half of \Cref{fig:ldn_spectrum}.
  Band-limiting a signal $u(t)$ to a maximum frequency $\hat f > 2q$ results in an NRMSE of at most 0.1 in the LDN spectrum, even if $u(t)$ is a white noise-signal.
  A discrete representation of this band-limited signal would require $N > 2 \hat f = 4q$ samples per second according to Nyquist-Shannon (we discuss this in more detail in \Cref{sec:nyquist-shannon}).
  The $q^2$ zero-order hold update algorithm then requires $4q^3$ operations per second, plus a low-pass filter, which can be cheaply implemented as a short FIR filter on $u(t)$.

  The Euler update on the other hand requires at least $N = 2.78q^2$ samples to reach an NRMSE of $0.1$, resulting in a total of $2.78q^3$ operations per second.
  While this is a little smaller than $4q^3$, remember that the estimate $N > 2 \hat f = 4q$ was derived under very conservative circumstances (i.e., a white noise signal as an input); lower band-limits can likely be used in practice.

  \begin{figure}[p]
    \centering
  	\includegraphics{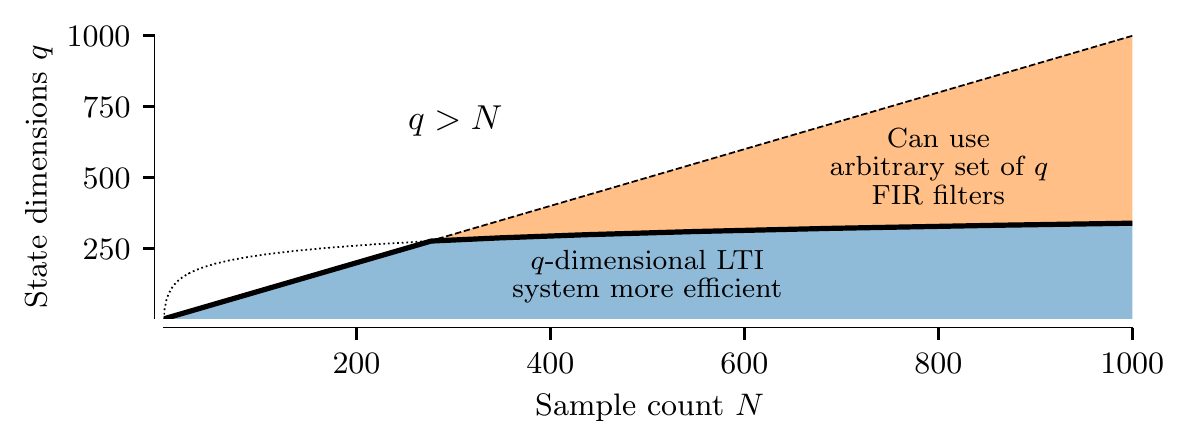}
  	\caption{Combinations of state dimensions $q$, and sample counts $N$ for which a suitable LTI system (such as the LDN) can compute discrete generalised Fourier coefficients ${\vec m}_t$ more efficiently compared to a FIR filter. Advancing an LTI system requires on the order of $q^2$ operations, whereas online evaluation of $q$ FIR filters requires about $34 \, q \log_2(N)$ operations per timestep \citep{gardner1995efficient}.}
  	\label{fig:ldn_vs_fir_complexity}
  \end{figure}

  \begin{table}[p]
  \caption{Run-time and memory costs for performing discrete function basis transformations of order $q$ and filter length $N$ (generally we assume $q \leq N$; for the batch FFT, FCT and FHT $q = N$).
  \enquote{Batch} describes transforming a $N$ samples into a different basis at once.
  \enquote{Sliding/Online} corresponds to computing the basis transformation for each incoming sample, where $N$ is the length of the filter. All memory costs include constant matrices.}
  \label{tbl:sliding_trafos}
  \centering\small
  \begin{tabular}{r l l l l l}
  	\toprule
  		& &
  		\multicolumn{2}{c}{\parbox{2cm}{\centering\textbf{Batch}\\\footnotesize($N$ samples)}} &   		\multicolumn{2}{c}{\parbox{2cm}{\centering\textbf{Sliding/Online}\\\footnotesize(per sample)}} \\
	  	\cmidrule(r){3-4}\cmidrule(l){5-6}
  	\emph{Basis} &
  	\emph{Algorithm} &
  	\emph{Run-time} &
  	\emph{Memory} &
  	\emph{Run-time} &
  	\emph{Memory} \\
  		\midrule
  	\textbf{FIR} &
  		Naive &
  		$\mathcal{O}(qN^2)$ &
  		$\mathcal{O}(qN)$ &
  		$\mathcal{O}(qN)$ &
  		$\mathcal{O}(qN)$ \\
		\cmidrule(l){2-2}\cmidrule(l){3-4}\cmidrule(l){5-6}
  		&
  		FFT conv. &
  		$\mathcal{O}(q N \log N)$ &
  		$\mathcal{O}(qN)$ &
  		/ &
  		/ \\
		\cmidrule(l){2-2}\cmidrule(l){3-4}\cmidrule(l){5-6}
  		&
  	    Gardner\textsuperscript{[1]} &
  		/ &
  		/ &
  		$\mathcal{O}(q \log N)$ &
  		$\mathcal{O}(q N \log N)$ \\
  	\midrule
	\textbf{Haar} &
		FHT\textsuperscript{[2]} &
  		$\mathcal{O}(N)$ &
  		$\mathcal{O}(N)$ &
  		$\mathcal{O}(q)$ &
  		$\mathcal{O}(N)$ \\
  	\midrule
	\textbf{Fourier} &
		FFT\textsuperscript{[3]} &
  		$\mathcal{O}(N \log N)$ &
  		$\mathcal{O}(N)$ &
  		/ &
  		/ \\
		\cmidrule(l){2-2}\cmidrule(l){3-4}\cmidrule(l){5-6}
		&
  		SDFT\textsuperscript{[4]} &
  		$\mathcal{O}(qN)$ &
  		$\mathcal{O}(N)$ &
  		$\mathcal{O}(q)$ &
  		$\mathcal{O}(N)$ \\
  	\midrule
	\textbf{Cosine} &
		FCT\textsuperscript{[5]}&
  		$\mathcal{O}(N \log N)$ &
  		$\mathcal{O}(N)$ &
  		/ &
  		/ \\
		\cmidrule(l){2-2}\cmidrule(l){3-4}\cmidrule(l){5-6}
		&
  		SCT\textsuperscript{[6]}&
  		$\mathcal{O}(qN)$ &
  		$\mathcal{O}(N)$ &
  		$\mathcal{O}(q)$ &
  		$\mathcal{O}(N)$ \\
  	\midrule
	\textbf{LDN} &
		ZOH LTI &
		$\mathcal{O}(q^2N)$ &
		$\mathcal{O}(N + q^2)$ &
  		$\mathcal{O}(q^2)$ &
  		$\mathcal{O}(q^2)$ \\
		\cmidrule(l){2-2}\cmidrule(l){3-4}\cmidrule(l){5-6}
		&
		Euler LTI\textsuperscript{[7]} &
  		$\mathcal{O}(qN)$ &
  		$\mathcal{O}(N)$ &
  		$\mathcal{O}(q)$ &
  		$\mathcal{O}(q)$ \\
  	\bottomrule
  \end{tabular}\\[0.25cm]
  \raggedright
  {\footnotesize \enquote{FIR} corresponds to an arbitrary set of FIR filters. \enquote{ZOH LTI} refers to the zero-order-hold discrete LTI system (eq.~\ref{eqn:ldn_h_matrix}). [1] \cite{gardner1995efficient}; [2] \cite{kaiser1998fast}; [3] \cite{cooley1965algorithm}; [4] \cite{jacobsen2003sliding}; [5] \cite{makhoul1980fast}; [6] \cite{kober2004fast}; [7] \cite[Figure~6.6]{voelker2019}.}
  \end{table}

  \paragraph{Sliding transformations and neural networks}
  In the context of neural networks, we can use basis transformations in their FIR filter representation for training.
  This avoids recurrences and can increase throughput, particularly when training the network on GPUs.

  During inference, a more efficient update rule can be used for the sliding transformation.
  A summary of the memory and run-time costs for the various transformations discussed in this report is given in \Cref{tbl:sliding_trafos}.

  As a side note, \Citet{gu2020hippo} derive sliding transformations similar to the LDN LTI system for different polynomial bases and weightings. 
  Our discussion of the LDN in terms of run-time and memory applies to these systems as well, although the low-pass filter characteristics may be different.

  \vfill

  \section{LTI Systems From Discrete Function Bases}
  \label{sec:reconstruct_lti}

  In the previous section, we derived a basis transformation matrix $\mat E$ from the LDN system $\mat A$, $\mat B$.
  We also saw that the LDN system can be used to directly compute discrete generalised Fourier coefficients $\vec m_t$;
  all we need to do is to simply advance the LTI system for each input sample $u_t$.
  For small $q$ this can be more efficient than repeated convolution with the past $N$ input samples.

  Of course, this raises the question whether we can reverse what we did above; in other words, derive an LTI system $\mat A$, $\mat B$ from arbitrary discrete function bases $\mat E$.
  This problem is known in the literature as a \enquote{system identification problem}.
  See \citet{verhaegen2007filtering} for a thorough treatise; the solution presented in this report is as crude as it is simple.

  \vfill

  \subsection{Solving For $\mA$, $\mB$ Using Least Squares}
  \marginnote{This method is implemented in \texttt{reconstruct\_lti}.}%
  In the last section, we constructed the unnormalised transformation matrix $\mat E$ as
  \begin{align*}
  	\big(\mat E^T\big)_k &= \mA^{N - k} \mB \,,
  	&\text{and correspondingly}\;\, \big(\mat E^T\big)_N &= \mB \,,
  	&\big(\mat E^T\big)_{i} &= \mA \big(\mat E^T\big)_{i - 1} \,,
  \end{align*}
  \clearpage
  where $k$ is the $k$th column of $\mat E$ and $\mA$, $\mB$ are as defined in \cref{eqn:ldn_recurrence_update}.%
  \begin{figure}
    \centering
  	\includegraphics{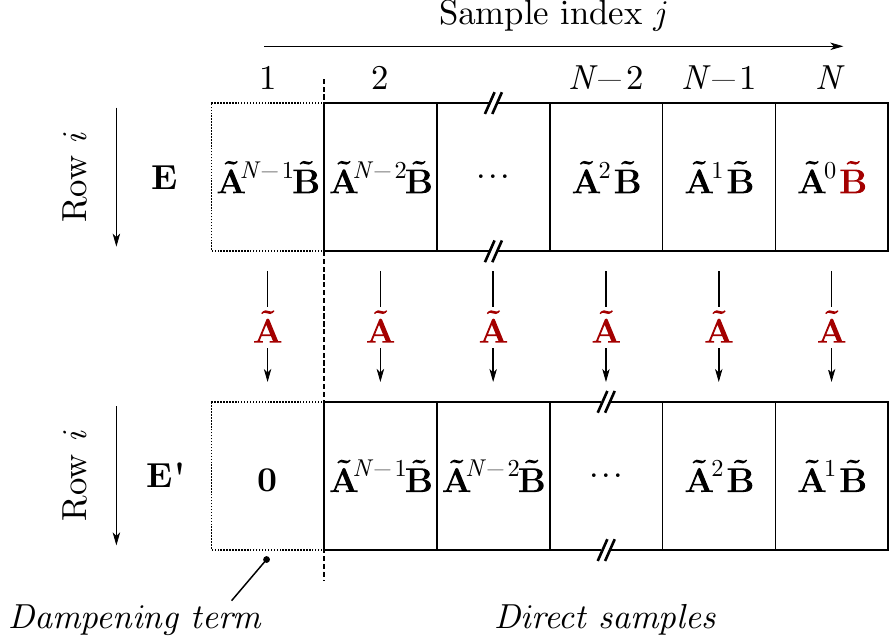}
  	\caption{Constructing a discrete LTI system $\mA$, $\mB$ from any discrete basis transformation matrix $\mat E$. $\mB \in \mathbb{R}^{q \times 1}$ is the last column of $\mat E$. $\mA$ can be obtained by solving $\mat E' \mA = \mat E$. Dampening encourages a finite impulse response.}
  	\label{fig:lti_system_reconstruction}
  \end{figure}
  We can directly read $\mB$ off $\mat E$, and estimate $\mA$ by solving for a matrix that translates between individual columns using least squares.
  This is illustrated in \Cref{fig:lti_system_reconstruction} (ignore dampening for now).
  Reverting discretisation yields an LTI system $\mat A, \mat B$:
  \begin{align}
  	\mat A &= \frac{N}{\theta}\log(\mA) \,, &
  	\mat B &= \frac{1}{\theta}(\mA - \mat I)^{-1} \mat A \mB \,.
  	\label{eqn:lti_inverse_discretisation}
  \end{align}

  One caveat with this approach is that the discrete basis transformation matrix $\mat E$ must be at least of size $q \times (q + 1)$, the total number of degrees of freedom in $\mA \in \mathbb{R}^{q \times q}$ and $\mB \in \mathbb{R}^{q \times 1}$.
  Methods for solving for $\mA$, $\mB$ with fewer degrees of freedoms exist (see \cite{verhaegen2007filtering}, Chapter 7 onward).

	\begin{figure}[p]
	 	\centering
	 	\includegraphics[trim={0.125cm 0.05125cm 0.125cm 0cm},clip]{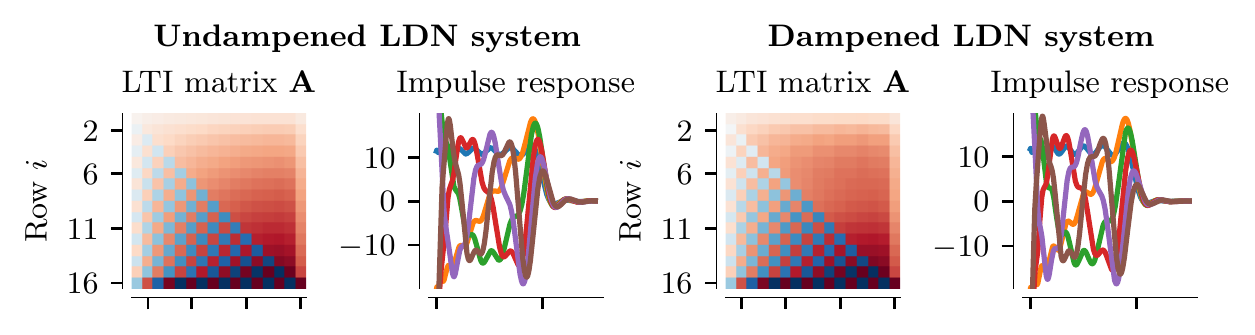}
	 	\includegraphics[trim={0.125cm 0.05125cm 0.125cm 0cm},clip]{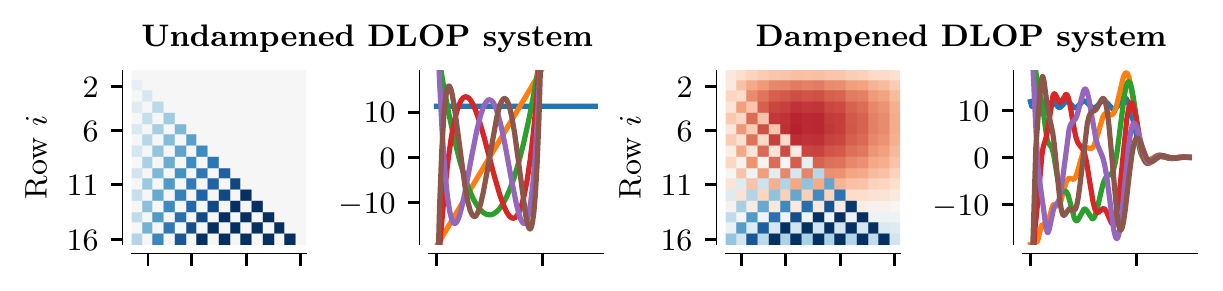}
	 	\includegraphics[trim={0.125cm 0.05125cm 0.125cm 0cm},clip]{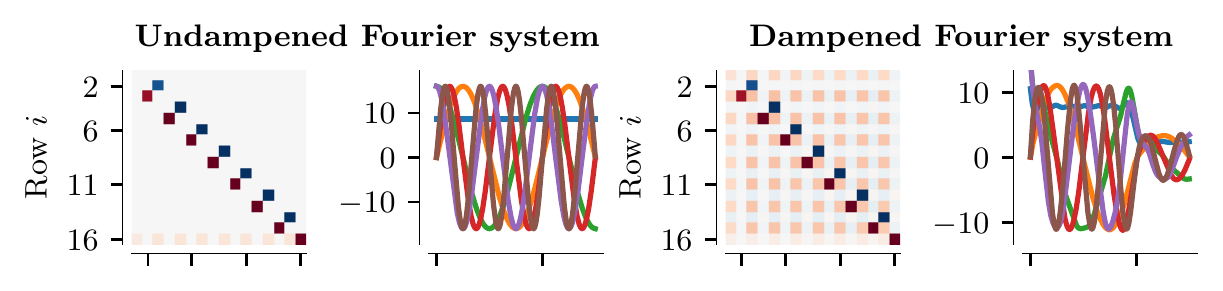}
	 	\includegraphics[trim={0.125cm 0.05125cm 0.125cm 0cm},clip]{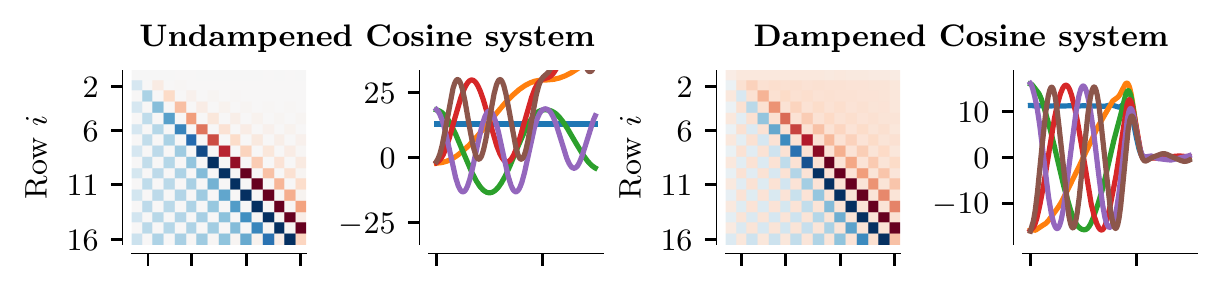}
	 	\includegraphics[trim={0.125cm 0.05125cm 0.125cm 0cm},clip]{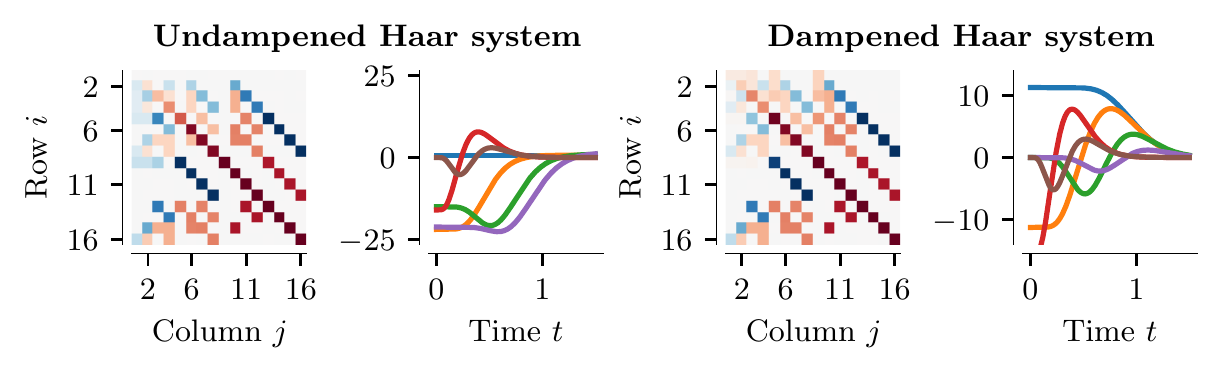}
	 	\caption{Constructing LTI systems from discrete function bases using least squares. Parameters are $q = 16$, $N = 128$, and $\theta = 1$.
	\emph{Left half:} LTI systems reconstructed without dampening. \emph{Right half:} LTI systems reconstructed with dampening.
	\emph{Left:} Feedback matrix $\mat A$. Positive values are blue, negative values red. Colour map is rescaled such that 95\% of the values are depicted without saturation.
	\emph{Right:} Impulse response of the first six state dimensions.
	 	}
	 	\label{fig:lti_reconstruction}
	\end{figure}

  \paragraph{Description of the resulting LTI systems}
  The left half of \Cref{fig:lti_reconstruction} depicts feedback matrices $\mat A$ and impulse responses for LTI systems constructed from the transformation matrices discussed so far.
  We now briefly discuss each system.\\

  \noindent\emph{LDN system.}
  Notably, the LDN system is---apart from some scaling factors due to regularisation---perfectly reconstructed.
  This should not surprise; our reconstruction method reverts the steps taken to construct $\mat H$ from $\mat A$, $\mat B$.\\

  \noindent\emph{DLOP system.}
  The DLOP LTI system perfectly generates the Legendre polynomials over the interval $[0, 1]$ as its impulse response.
  Unfortunately, being able to do so is rather pointless, since the impulse response quickly diverges for $t > 1$.
  Notice how the DLOP system $\mat A$ matrix is strictly positive.
  The positive coefficients in the LDN system matrix are very similar to the positive DLOP system coefficients; hence, the LDN system can be thought of implementing the Legendre basis with some dampening through the added negative coefficients.\\

  \noindent\emph{Fourier system.}
  As one would expect, the Fourier system consists of $\frac{N}2$ independent oscillators with increasing frequency. Just like the DLOP system, the Fourier system has an infinite impulse response, but unlike the DLOP system, the system is stable.\\

  \noindent\emph{Cosine system.}
  Apart from some scaling and shifting issues, the cosine system reconstructs the cosine basis over the interval $[0, 1]$.
  For $t > 1$ the impulse response no longer resembles the cosine basis.\\

  \noindent\emph{Haar system.}
  We cannot expect a finite-dimensional LTI system to reproduce the discontinuities in the Haar basis.
  In contrast to the other systems, the Haar system has a finite impulse response for $q = 16$, but this is not guaranteed.

  \subsection{Dampening the Reconstructed System}
  Most of the reconstructed systems have an infinite impulse response.
  This is not desirable unless the system is never advanced beyond $\theta$, or, in other words, only a fixed number of samples is processed.

  In this report, we are more concerned with online processing, which implies a potentially unbounded number of samples and the inability to perform batch updates.
  To be suitable for online processing,
  the LTI system must not only optimally approximate a function basis, but also \enquote{optimally forget} information originating from more than $\theta$ seconds in the past.

  In the following, we discuss two extensions to the above method, that encourage, but do not guarantee, a finite impulse response.
  The first method adds a \enquote{dampening} term to the least-squares equations.
  The second method explicitly erases information older than $\theta$ seconds from the state $\vec m$.

  \paragraph{Dampening term in the least-squares system}
  \marginnote{This method is implemented in \texttt{reconstruct\_lti} when passing \texttt{dampen = "lstsq"} as a parameter.}%
  One way to encourage a finite impulse response is to add the equation $\mA (\mat E^T)_1 = \mat 0$ to the least-squares problem.
  As illustrated in \Cref{fig:lti_system_reconstruction}, multiplying the first column of the discrete function basis (which corresponds to $t = \theta$) with $\mA$ should extinguish the impulse response. This dampening term should be weighted by $\frac{N - 1}{q - 1}$,
  maintaining a weight ratio of $1 : q - 1$ between the dampening term and the remaining samples.

  The least-squares system now contains exactly $N$ samples, so, in theory, it could be used for matrices $\mat E$ of shape $q \times N$ with $N = q$.
  However, in practice, this often results in a singular $(\mA - \mat I)$ or non positive definite $\mA$, for which the matrix logarithm cannot be computed.

  This least-squares approach to system identification may not lead to optimal results.
  It may be beneficial to iteratively enforce the dampening condition using the actual impulse response of the reconstructed system.

  \paragraph{Dampening by information erasure}
  \marginnote{This method is implemented in \texttt{reconstruct\_lti} when passing \texttt{dampen = "erasure"} as a parameter.}%
  For our system to have a finite impulse response, we must ensure that information older than $\theta$ seconds (or $N$ samples) is \enquote{forgotten}.
  We can enforce this by \emph{decoding} the input $u_{t - N + 1}$ from the current state $\vec m_t$ and computing the state vector $\vec m'_t$ \emph{encoding} this input.
  Subtracting $\vec m'_t$ from $\vec m_t$ erases $u_{t - N + 1}$ from the state.
  Hence, this \enquote{old} sample can no longer influence the system state and the impulse response must be finite.

  In practice, assuming that the discrete LTI system $\mA$, $\mB$ reconstructs the discrete function basis $\mat E$, the sample $u_{t - N + 1}$ can be decoded as follows
  \begin{align*}
  	u_{t - N + 1} &= \big(\mat E^+\big)_1 \vec m_t \,, & \text{where } \mat E^+ = \big(\mat E^T \mat E\big)^{-1} \mat E^T\,,
  \end{align*}
  and $\big(\mat E^+\big)_1$ denotes the first row of the pseudo-inverse. Re-encoding $u_{t - N + 1}$ under a zero-order hold assumption yields
  \begin{align*}
  	\vec m'_t &= \big( \mat E^T \big)_1 u_{t - N + 1} = \big( \mat E^T \big)_1 \otimes \big(\mat E^+\big)_1 \vec m_t \,,
  \end{align*}
  where $\otimes$ is the outer product and $\big( \mat E^T \big)_1$ is the first column of $\mat E$.
  Sequentially interleaving an \enquote{update} and an \enquote{erasure} step we obtain
  \begin{align*}
  	\vec m_t &\gets \mA \vec m_{t - 1} + \mB u_t \,, & \text{(Update)}\\
  	\vec m_t &\gets \vec m_t - \vec m'_t = \big( \mat I - \big( \mat E^T \big)_1 \otimes \big(\mat E^+\big)_1 \big) \vec m_t \,. & \text{(Erasure)} \\
  	&= \underbrace{(\mat I - \big( \mat E^T \big)_1 \otimes \big(\mat E^+\big)_1 ) \mA}_{\mA'} \vec m_t +
  	   \underbrace{(\mat I - \big( \mat E^T \big)_1 \otimes \big(\mat E^+\big)_1 ) \mB}_{\mB'} u_t \,.
  \end{align*}
  Applying the inverse discretisation from \cref{eqn:lti_inverse_discretisation} to $\mA'$, $\mB'$ results in a dampened, continuous LTI system.
  Again, this is not an optimal solution, since the equations assume that $\mA$ and $\mB$ perfectly reconstruct the discrete function basis over $[0, \theta]$.

  A continuous version of this method can be used to directly derive the LDN system from the Legendre polynomials \citep{stockel2021constructing}.

  \paragraph{Description of the dampened LTI systems}
  Surprisingly, both dampening methods yield similar results---at least with respect to the impulse response of the resulting systems.
  In the following, we discuss the results obtained when using the information erasure method.
  The right half of \Cref{fig:lti_reconstruction} depicts the impulse responses of the dampened LTI systems and the corresponding feedback matrices $\mat A$.
  As before, we will quickly comment on these for each discrete function basis.\\

  \noindent\emph{Dampened LDN system.}
  The LTI system and its impulse response remain almost unchanged.
  This includes the higher order dimensions (not depicted).\\

  \noindent\emph{Dampened DLOP system.}
  The dampened DLOP system has---apart from a different scaling of the individual systems---\emph{almost} exactly the same impulse response as the LDN system.
  For all practical purposes, the LDN and dampened DLOP system seem equivalent.\\

  \noindent\emph{Dampened Fourier system.}
  Dampening the Fourier system leads to some reduction in magnitude for $t > \theta$.
  While the impulse response eventually does decay to zero, it does so very slowly.\\

  \noindent\emph{Dampened Cosine system.}
  Especially the depicted lower order terms of the dampened cosine system decay to zero quite quickly; however, this is not true for the higher order terms (not depicted), which behave more like the dampened Fourier system.
  Notice that the impulse response somewhat resembles the Legendre basis.\\
  
  \noindent\emph{Dampened Haar system.}  
  The dampened Haar system behaves well; though it is not clear whether the impulse response resembles a sensible function basis.

  \subsection{Summary}
  \label{sec:lti_discussion}
  The technique we presented in this section could be improved by parametrising $\mat A$, $\mat B$ in a way that enforces stability and results in well-conditioned systems.
  Nevertheless, our naive least-squares solution produces LTI systems that very well approximate a discrete function basis $E_n(k; N)$ over an interval~$[0, \theta]$.

  However, the true challenge lies in generating an LTI system that has a \enquote{finite impulse response}, or, in other words, quickly converges to zero for $t > \theta$.
  While introducing a dampening term accomplishes this to some degree, most of the systems that end up with a finite impulse response (LDN, DLOP, partially the Cosine system) resemble the Legendre polynomials.

  It is unclear whether LTI systems approximating bases other than the Legendre polynomials are really necessary.
  If so desired, similar approximations of other bases can be constructed by equipping the LDN system with a corresponding readout matrix $\mat T \in \mathbb{R}^{q \times q}$.
  Due to linearity, this generally yields results superior to what was shown here in terms of the finiteness of impulse response.
 
  \section{Low-pass Filtering Discrete Function Bases}
  \label{sec:lp}

  If the number of discrete basis functions $q$ is smaller than the number of samples~$N$, then a discrete basis transformation $\mat E \in \mathbb{R}^{q \times N}$ is a \emph{projection} mapping from a higher- onto a lower-dimensional space.
  Computing $\vec m = \mat E \vec u$ can be thought of as \enquote{packing} $N$ samples $\vec u$ into a smaller $q$-dimensional vector $\vec m$.
  In general, projections are \enquote{lossy}, that is, they decrease the amount of information present in $\vec m$ compared to $\vec u$.

  In this section, we talk about the conditions under which information loss occurs, leading us to the Nyquist-Shannon sampling theorem discussed in \Cref{sec:nyquist-shannon}, as well as a simple related lemma that can be applied to discrete function bases.
  Violating the conditions in the Nyquist-Shannon sampling theorem or the lemma results in a phenomenon called \enquote{aliasing}.
  We discuss this in \Cref{sec:aliasing}.
  We close with a technique for constructing discrete function bases that apply an anti-aliasing filter to $\vec u$ in \Cref{sec:filtering}.
  It is not immediately clear whether this is necessary or beneficial in the context of discrete function bases.

  \subsection{The Nyquist-Shannon Sampling Theorem}
  \label{sec:nyquist-shannon}
  In most cases, projections such as $\mat E$ with $q < N$ inadvertently destroy information.
  That is, a signal $\mat u$ cannot be reconstructed from $\mat E \vec u$. An exception are signals $\vec u$ that are a linear combination of the $q$ rows of the projection matrix $\mat E$, i.e., $\vec u = \mat E^T \vec m$, where $\vec m \in \mathbb{R}^q$.
  We express this formally in the lemma below.

  \begin{lemma}
  \label{lem:reconstruction}
  Let $\mat E \in \mathbb{R}^{q \times N}$ with $q \leq N$, $\mat E \mat E^T = \mat I$, and $\vec u \in \mathbb{R}^N$. It holds $\mat E^T \mat E \vec u = \vec u$ exactly if a unique $\vec m \in \mathbb{R}^q$ exists such that $\vec u = \mat E^T \vec m$.
  \begin{proof}
  Consider the \enquote{$\Leftarrow$} direction.
  Let $\vec m \in \mathbb{R}^q$ and $\vec u = \mat E^T \vec m$. We need to show $\mat E^T \mat E \vec u = \vec u$. Expanding the left-hand side of the latter expression we get
  \begin{align*}
	\mat E^T \mat E \vec u = \mat E^T \mat E \mat E^T \mat m = \mat E^T \mat m = \vec u \,.\; \checkmark
  \end{align*}
  Now, consider  the \enquote{$\Rightarrow$} direction. We have $\mat E^T \mat E \vec u = \vec u$ and must show that a unique $\vec m$ exists with $\vec u = \mat E^T \vec m$. Equating the two terms we get $\mat E^T \mat E \vec u = \mat E^T \vec m$, hence $\mat E \vec u = \vec m$. This is the only solution. Assume another solution $\vec m' \neq \vec m$ with $\vec u = \mat E^T \vec m'$ exists. Then $\mat E^T (\vec m - \vec m') = \vec 0$. Hence, $\vec 0 \neq \vec m' - \vec m \in \ker(\mat E^T)$. Contradiction: $\ker(\mat E^T)$ is zero-dimensional and only contains $\vec 0$. This is because $\mat E^T$ is of rank $q$ (as $\mat E \mat E^T = \mat I$) and $\dim(\ker(\mat E^T)) + \mathrm{rank}(\mat E^T) = q$. ~\lightning
  \end{proof}
  \end{lemma}

  \begin{figure}[t]
  	\centering
  	\includegraphics{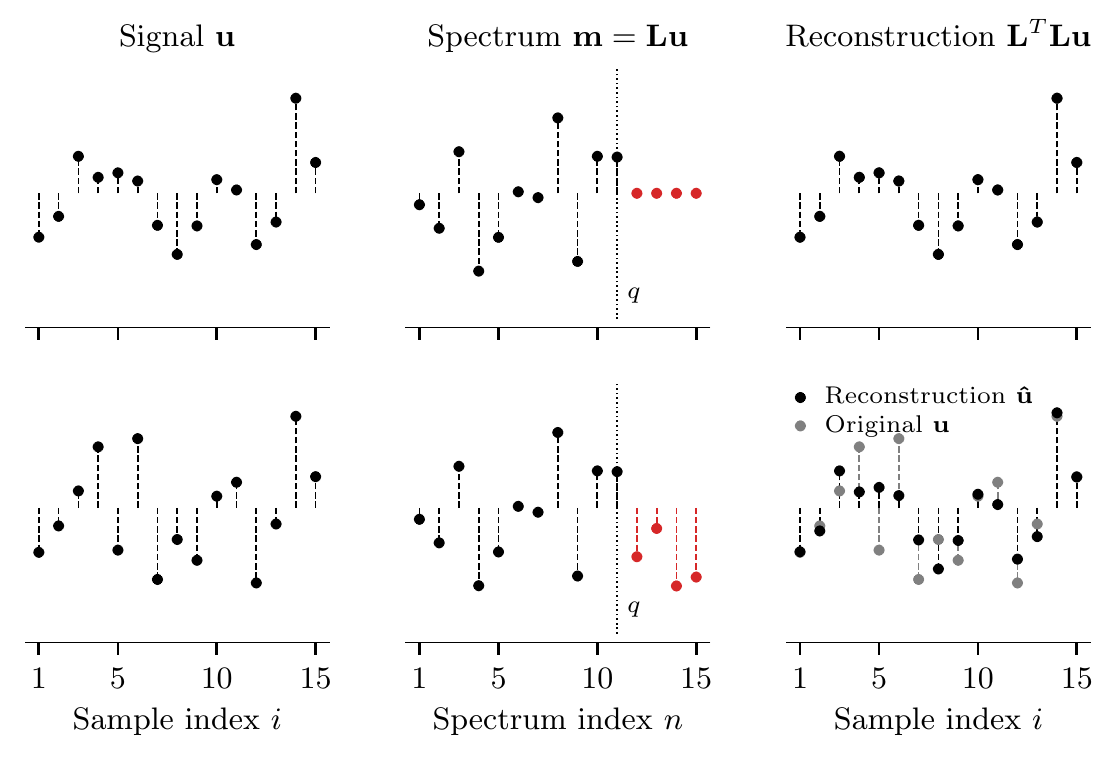}
  	\caption{Illustration of \Cref{lem:reconstruction}.
  	The discrete signal $\mat u$ has $N = 15$ samples.
  	The signal can be losslessly transformed using $\mat L \in \mathbb{R}^{q \times N}$ with $q = 11$ if $\vec u$ can be constructed from the first $q$ basis functions.
  	\emph{Top left:} Appropriately band-limited signal. \emph{Top centre:} The Fourier coefficients $m_n$ for $n \geq q$ (red) are all zero. \emph{Top right:} The signal can be perfectly reconstructed.
  	\emph{Bottom:} The depicted signal is not properly bandlimited and cannot be losslessly reconstructed. Due to aliasing, it is reconstructed as if it was the signal in the top of the figure.
  	}
  	\label{fig:signal_reconstruction}
  \end{figure}
  Essentially, this lemma states a condition under which a unique reconstruction of a signal $\vec u$ with respect to any basis transformation matrix $\mat E$ is possible.
  Put differently, the signal $\vec u$ must have been \enquote{bandlimited} to be constructed from at most $q$ functions of the function basis it is being transformed into and $N \geq q$ samples are required. This is illustrated in \Cref{fig:signal_reconstruction}.
  For non-orthogonal matrices $\mat E$ of rank $q$, the Moor-Penrose pseudo inverse $\mat E^+ = (\mat E^T \mat E)^{-1} \mat E^T$ can be used instead of $\mat E^T$.
  
  The Nyquist-Shannon sampling theorem \citep{shannon1949communication} is similar to the above lemma in spirit. It bridges the continuous and discrete time domains and states a condition under which continuous signals can be perfectly represented by a set of samples $\vec u = (u_1, \ldots, u_N)$.
  Specifically, if a function over $[0, 1]$ only contains frequencies up to $\phi$ Hertz, then $N \geq 2\phi + 1$ uniform samples are required to perfectly represent $f(t)$. 

  \begin{thm}[Nyquist-Shannon Sampling Theorem]
  \label{thm:nyquist}
  Let $f(t)$ be a bandlimited function over $\mathcal{C}[0, 1]$; that is $\langle f, f_n \rangle = 0$ for all $n > \hat q$, where $(f_n)_{n \in \mathbb{N}}$ is the Fourier series as defined in \cref{eqn:fourier_series}.
  $f(t)$ is completely determined by $N$ uniform sample points over $[0, 1]$ if $N > \hat q$ 
  (cf.~\cite{shannon1949communication} for the proof).
  \end{thm}

  \subsection{Aliasing}
  \label{sec:aliasing}

  The above lemma and theorem state under which conditions signals can be represented and transformed without information loss.
  But what happens if we violate these conditions? If we discretise a signal $f(t)$ with fewer samples than necessary, or transform a signal $\vec u$ into a function space where it cannot be represented using $q$ coefficients?
  Besides losing information, the answer to this question is \emph{aliasing}, though to different degrees of severity.
  Aliasing means that infinite \enquote{invalid} signals $f(t)$ or $\vec u$ are mapped onto the representation of a single \enquote{valid} signal.

  \paragraph{Discrete case (violation of \Cref{lem:reconstruction})}
  In the discrete case, for a fixed $\vec m \in \mathbb{R}^q$, there are an infinite number of $\vec u \in \mathbb{R}^{N}$ with $\vec m = \mat F\vec u$ if $N > q$.
  Fortunately, assuming that the basis vectors are sorted by frequency, computing $\mat F \vec u$ merely discards higher frequency terms.
  We saw this in \Cref{fig:signal_reconstruction}, where the \enquote{invalid} signal is aliased onto the signal with the same first $q$ spectral coefficients.

  \begin{figure}
  	\centering
  	\includegraphics[trim={0.125cm 0.25cm 0.125cm 0.25cm},clip]{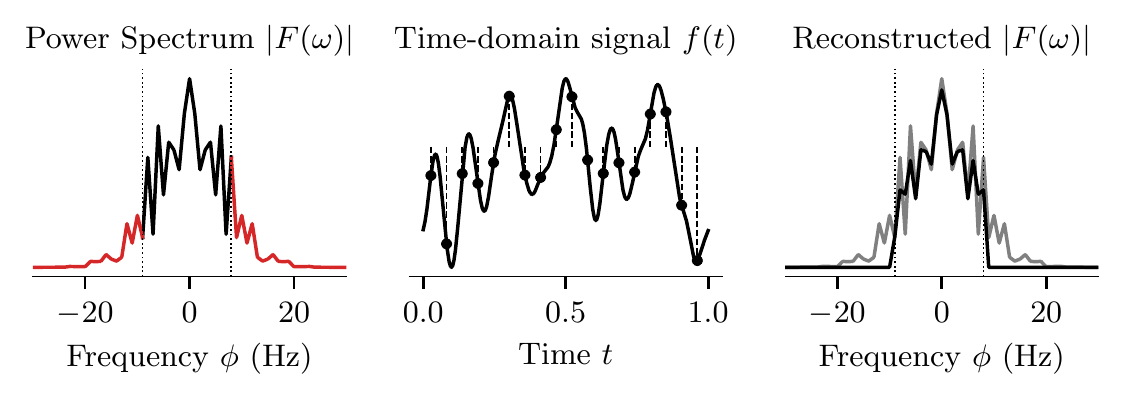}
  	\caption{Violation of the Nyquist-Shannon theorem causes distortion. \emph{Left:} Original power spectrum. Red segments should be zero for the $N$ used. \emph{Centre:} Original signal and samples. \emph{Right:} Sampled power spectrum. Note the difference to the original spectrum (grey) within the valid bounds (dotted lines).}
  	\label{fig:nyquist_violation}
  \end{figure}
  \paragraph{Sampling continuous functions (violation of \Cref{thm:nyquist})}
  Violating the Nyquist-Shannon theorem causes distortions inside the frequency range representable by $N$ samples.
  This is depicted in \Cref{fig:nyquist_violation}.
  Not only are frequency coefficients outside the $N / 2 \,\mathrm{Hz}$ range discarded when sampling a one-second signal $f(t)$ too sparsely, there are also changes (\enquote{distortions}) in the magnitude of the frequency coefficients within that limit.
  That is, our incorrectly sampled function is aliased onto a function with (potentially) completely different frequency content.
  See \citet[Section 12.1.1, pp.~605-606]{press2007numerical} for more information.
  This is significantly worse than what happens when we transform a discrete signal $\mat u$ into a discrete function space that does not span $\mat u$.

  \subsection{Anti-Aliasing Filters}
  \label{sec:filtering}

  To combat aliasing (i.e., \enquote{anti-aliasing}), we simply map each violating signal onto a valid signal that does not violate any of the above constraints.
  For example, continuous functions $f(t)$ can be low-pass filtered in the continuous domain to discard frequencies outside the $N / 2\,\mathrm{Hz}$ range.
  For discrete $\vec u$, we similarly discard components of $\vec u$ that cannot be represented in the target basis.

  Put differently, \enquote{anti-aliasing} can be thought of as \enquote{controlled aliasing}.
  The invalid signals are deterministically mapped (\enquote{aliased}) onto a valid signal.
  As we saw, this is important when sampling continuous functions, as invalid signals cause distortions in the representable frequencies.

  \paragraph{Band-limiting discrete signals}
  Given a basis transformation matrix $\mat E$, a valid signal $\vec u'$ (in the sense of \Cref{lem:reconstruction}) is simply given as $\vec u' = \mat E^+ \mat E \vec u$, where $\mat E^+ = (\mat E^T \mat E)^{-1} \mat E^T$ is the Moore-Penrose pseudo inverse of $\mat E$.
  This filtering does not affect the generalised Fourier coefficients $\vec m$. We have
  \begin{align*}
  	\vec m = \mat E \vec u' = \mat E \mat E^+ \mat E \vec u = \mat E \vec u \,.
  \end{align*}

  \paragraph{Filtering basis transformation matrices}
  To spice things up a little, we can use two different discrete function bases; one for the transformation, and one for filtering.
  For example, let $\mat E \in \mathbb{R}^{q \times N}$ be an arbitrary discrete basis transformation matrix, and let $\mat F \in \mathbb{R}^{q' \times N}$ be the Fourier matrix.
  Projecting $\vec u$ onto the first $q'$ coefficients of the Fourier basis, i.e., computing $\vec u' = \mat F^T \mat F \vec u$, and then applying the transformation $\mat E$ yields
  \marginnote{~~\\[-0.7em]This equation is implemented in \texttt{lowpass\_filter\_\\basis}.}
  \begin{align}
  	\vec m = \mat E \vec u' = \mat E \mat F^T \mat F \vec u = (\mat F^T \mat F \mat E^T)^T \vec u = \mat E_\ell \vec u \,.
  	\label{eqn:filter}
  \end{align}
  Crucially, the term $\mat E_\ell^T = \mat F^T \mat F \mat E^T$ can be thought of as filtering the individual discrete basis functions in $\mat E$.
  That is, instead of filtering the incoming signal $\vec u$, we band-limit the basis functions with respect to the Fourier basis.

  \begin{figure}[t]
  	\includegraphics{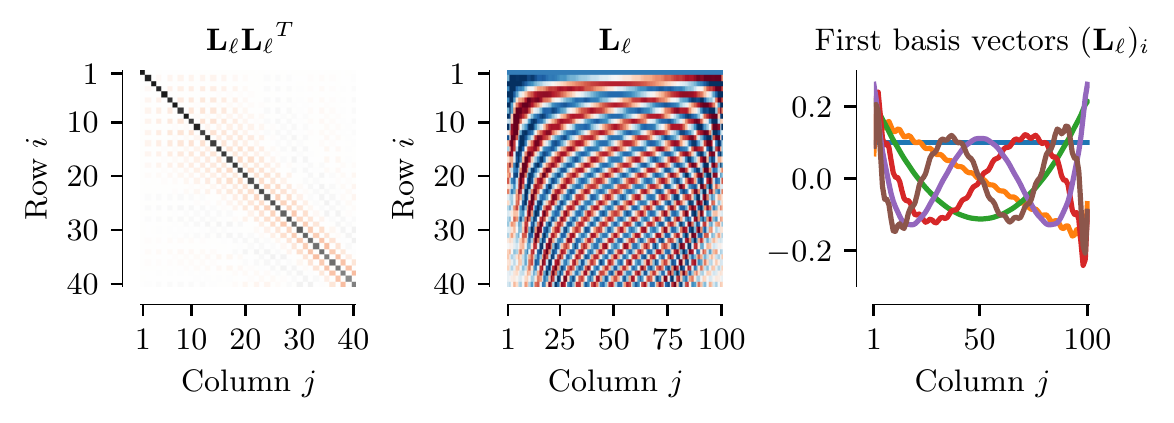}
  	\caption{Examples of a low-pass filtered DLOP basis $\mat L_\ell$ for $q = q' = 40$ and $N = 100$.
  	The filtered DLOP matrix $\mat L_\ell$ is no longer orthogonal, and instead exhibits ringing patterns similar to those found in the LDN basis.}
  	\label{fig:lp_examples}
  \end{figure}
  \Cref{fig:lp_examples} depicts a low-pass filtered versions of the DLOP function basis.
  The DLOP basis transformation matrix is no longer orthogonal after filtering and exhibits ringing artefacts similar to those seen in the LDN basis (cf.~\cref{fig:ldn_basis}).
  This may suggest that the LDN basis intrinsically contains an \enquote{optimal low-pass filter}, but this needs further investigation.
  Judging from the LTI experiments in the last section, the ringing artefacts may instead be required to realize the finite impulse response (e.g., compare the impulse response of the undampened and dampened DLOP system in \cref{fig:lti_reconstruction} to that of the LDN system).

  \paragraph{Connection to the Nyquist-Shannon theorem}
  Of course, any other discrete basis transformation matrix may be used for filtering instead of $\mat F$ (using the pseudo-inverse $\mat E^+$ if $\mat E$ is not orthogonal).
  However, using $\mat F$ establishes a connection to the Nyquist-Shannon theorem.
  Any signal $\mat u'$ constructed from a Fourier-filtered basis transformation matrix $\mat E_\ell$ (i.e., $\vec u' = \mat E_\ell^T \vec m$) uniquely corresponds to a continuous function $f(t)$ as defined by \citet{shannon1949communication}. 
  This $f(t)$ is the same as the function we would obtain when increasing $N \to \infty$.

  \paragraph{Filtering with the LDN system}
  Using the LDN basis as a filter transformation can be beneficial when constructing LTI systems as discussed in \Cref{sec:reconstruct_lti}.
  This ensures that the desired impulse response is a linear combination of the LDN system impulse response.
  The resulting LTI system is essentially a \enquote{rotated} version of the LDN system, guaranteeing a finite impulse response while still approximating the desired basis function reasonably well.

  However, this approach is equivalent to applying a readout matrix $\mat T$ to the LTI system state, as discussed in \Cref{sec:lti_discussion}.
  This is generally the better solution compared to using a different $\mat A$, $\mat B$, since the LDN system is already well conditioned.

  \section{Evaluating Discrete Function Bases}

  So far, we have discussed several discrete function bases, but we have not given any practical recommendation as for when to use which discrete function basis.
  This is because the notion of the \enquote{best} basis highly depends on the application.

  In this section, we will first try to provide some very general recommendations as for when to use which function basis.
  We then perform a series of benchmarks.
  First, we evaluate how well delayed signals can be decoded from the representations formed by each function basis.
  Second, we test the discrete function bases as fixed temporal convolutions in a neural network context.

  \subsection{General Considerations}

  The \emph{continuous} function bases discussed above can be divided into three categories: periodic, aperiodic, wavelets.\\

  \noindent\emph{Periodic bases.}
  	      A function basis $(f_n)_{n \in \mathbb{N}}$ over $[0, 1]$ is \emph{periodic} if for every $f_n$ the following function $\tilde f_n$ is infinitely continuously differentiable over $\mathbb{R}$:
  	      \begin{align*}
  	      	\tilde f_n(x) = f_n\left( x - \left\lfloor x \right\rfloor \right) \,.
  	      \end{align*}
  	      Intuitively, a function $f(x)$ fulfilling this requirement has no observable boundary between $x = 1$ and $x = 0$.
  	      Of all the bases we saw, only the Fourier basis (eq.~\ref{eqn:fourier_series}) fulfils this requirement.
  	      Periodic bases are particularly suitable for representing periodic functions.
  	      While these bases can, of course, represent any aperiodic function $g \in L^2(0, 1)$, a \emph{finite} generalised Fourier series of $g$, that is,
  	      \begin{align*}
  	      	\hat g = \sum_{n = 0}^{q - 1} \langle g, f_n \rangle f_n \,,
  	      \end{align*}
  	      will always be periodic (e.g., observe the \enquote{spikes} at the boundaries in \cref{fig:sign_sine}).\\

  	\noindent\emph{Aperiodic bases.}
  	      If a basis is not periodic in the sense defined above, it is \emph{aperiodic}.
  	      This includes the cosine (eq.~\ref{eqn:cosine_basis}), Legendre (eq.~\ref{eqn:legendre_basis}) and Haar basis (eq.~\ref{eqn:haar_basis}).
  	      Analogously to the above, such bases are well suited to representing aperiodic functions.\\

  	\noindent\emph{Wavelet bases.}
  	      Without defining this concept thoroughly, wavelet bases, such as the Haar basis (eq.~\ref{eqn:haar_basis}), are characterized by being sparse.
  	      This is useful when approximating localized functions $g$, i.e., functions that are exactly zero for most~$x$.
  	      Wavelet bases can also be thought of as intrinsically segmenting signals $\vec u$ into smaller \enquote{chunks}. Hence, a small change at one point in time does not affect all generalised Fourier coefficients at once.
  	      One issue with wavelet bases is that increasing $q$ by one (i.e., adding one basis function) does not decrease the approximation error $\hat g(x) - g(x)$ uniformly for all $x$.

  \paragraph{Properties of various discrete function bases}
  This list briefly characterises the \emph{discrete} function bases we discussed in this report in terms of their computational properties and common applications where applicable.\\

  \noindent\emph{Discrete Fourier basis.}
  This periodic basis provides natural interpretations of signals in terms of \enquote{frequency} and \enquote{phase}.
  Periodicity can be important when solving certain differential equations.
  Computing $\mat F \vec u$ for $\vec u \in \mathbb{R}^N$ and $\mat F \in \mathbb{R}^{N \times N}$ requires $\mathcal{O}(N \log(N))$ operations \citep{cooley1965algorithm}.\\

  \noindent\emph{Discrete Cosine basis.}
  This aperiodic basis tends to approximate the principal components of natural signals (such as sounds or images) and is thus extensively used in audio, video, and image compression.
  Furthermore, the derivative of these basis functions is zero at the boundaries, which is another common condition encountered when solving differential equations \citep[Section 12.4.2,~p.~624]{press2007numerical}.
  Computing $\mat C \vec u$ for $\vec u \in \mathbb{R}^N$ and $\mat C \in \mathbb{R}^{N \times N}$ requires $\mathcal{O}(N \log(N))$ operations \citep{makhoul1980fast}.\\

  \noindent\emph{DLOP basis.}
  As discussed by \citet{neuman1974discrete}, this aperiodic basis is particularly useful when performing smooth polynomial interpolation.
  In contrast to the LDN basis, the DLOP basis is orthogonal.
  To our knowledge, there is no fast computation scheme; computing $\mat L \vec u$ for $\vec u \in \mathbb{R}^N$ and $\mat L \in \mathbb{R}^{q \times N}$ requires $\mathcal{O}(q N)$ operations.\\

  \noindent\emph{LDN basis.}
  This aperiodic basis is similar to the DLOP basis; computing $\mat H \vec u$ for $\vec u \in \mathbb{R}^N$ and $\mat H \in \mathbb{R}^{q \times N}$ requires $\mathcal{O}(qN)$ operations.
  Because of the underlying LTI system, the LDN basis has a potentially faster online, zero-delay update scheme, requiring $\mathcal{O}(q^2)$ operations per timestep.
  Refer to \Cref{sec:ldn_when} for a thorough discussion.\\

  \noindent\emph{Haar basis.}
  As noted above, convolving a signal with this wavelet basis has a low time complexity.
  It is therefore extensively used in classic computer vision algorithms, including tasks such as face recognition \citep{viola2001rapid}.
  Computing $\mat W \vec u$ for $\vec u \in \mathbb{R}^N$ and $\mat W \in \mathbb{R}^{q \times N}$ requires $\mathcal{O}(N)$ operations.
  Online, zero-delay updates require $\mathcal{O}(q)$ operations  per timestep \citep{kaiser1998fast}.
  A downside is that this basis is not orthogonal for non-power-of-two $N$, and that $q$ \emph{should} be a power of two as well.

  \subsection{Delay Decoding Benchmark}

  An objective measure for the \enquote{quality} of a discrete function basis is the amount of information lost when encoding a signal $\vec u$. That is, we could simply compute
  \begin{align*}
  	E = \| \vec u - \mat E^+ \mat E \vec u \| = \| \vec u - (\mat E^T \mat E)^{-1} \mat E^T \mat E \vec u \|
  \end{align*}
  to evaluate a discrete function basis with an associated basis transformation matrix $\mat E \in \mathbb{R}^{q \times N}$.
  We can vary the number of discrete basis functions $q$ and see in how far this affects the decoding error $E$.
  In fact, this is essentially what we will do, though a few details require further explanation.

  \paragraph{Input signal selection}
  One challenge with this approach is that the error measurement depends on the specific $\vec u$.
  Remember that each basis has a set of signals $\vec u$ that can be represented without error; as stated in \Cref{lem:reconstruction}, these are exactly the linear combinations of the basis vectors.
  If we are not careful, we could generate test signals $\vec u$ that fall into this category and thus skew the results favourably towards one particular basis.

  We try to mitigate this in two ways.
  First, we average over a large number of randomly generated input signals.
  Second, the input signal itself is a low-pass filtered white noise signal.
  As such it contains frequencies up to the Nyquist frequency $N / 2$, though with a small magnitude.
  We explicitly do not use hard band-limiting (eq.~\ref{eqn:filter}), as this would drastically bias the results toward the Fourier basis.
  Still, one shortcoming of our methodology is that we do not test a large corpus of different (natural) input signals.
  The neural network experiments in the next subsection aim to address this.

  \paragraph{Sample decoder}
  Instead of using the pseudo-inverse $\mat E^+$ as suggested above, we reconstruct individual input samples using a \enquote{learned} decoding vector $\vec d_\theta$.
  This emulates a simple machine learning context.

  Let $\vec u_t = (u_{t - (N - 1)}, \ldots, u_t) \in \mathbb{R}^N$ be an input signal.
  Given the discrete generalised Fourier coefficients $\mat m_t = \mat E \vec u_t$, we would like to determine how well a specific $u_{t - \theta}$ with $\theta \in \{0, \ldots, N - 1\}$ can be reconstructed from $\mat E \vec u_t$.
  That is, given a fixed $\vec d_\theta$, we measure the error $E = \langle \vec d_\theta, \mat E \vec u_t \rangle - u_{t - \theta}$
  over many $\vec u$ and $t$.

  The decoder $\vec d_\theta$ is fixed for each pair of $\mat E$ and $\theta$ and is determined using regularised least squares over a large set of separate test signals.\footnote{For regularisation we pass $\mathtt{rcond} = 10^{-4}$ to the Numpy $\mathtt{lstsq}$ method.}
  In contrast to the corresponding row of $\mat E^+$, the decoder is fine-tuned to the class of signals $\vec u_t$ encountered in this experiment.
  Regularisation enforces a robust $\vec d_\theta$; the decoder produces a correct result even if noise is added to $\vec m$.
  As a side effect, the individual coefficients of $\vec d_\theta$ have a relatively small magnitude.
  The delay decoders $\vec d_\theta$ could be learned via stochastic gradient descent in a neural network.

  \begin{figure}
  	\centering
  	\includegraphics[trim={0cm 0.125cm 0cm 0.3cm},clip]{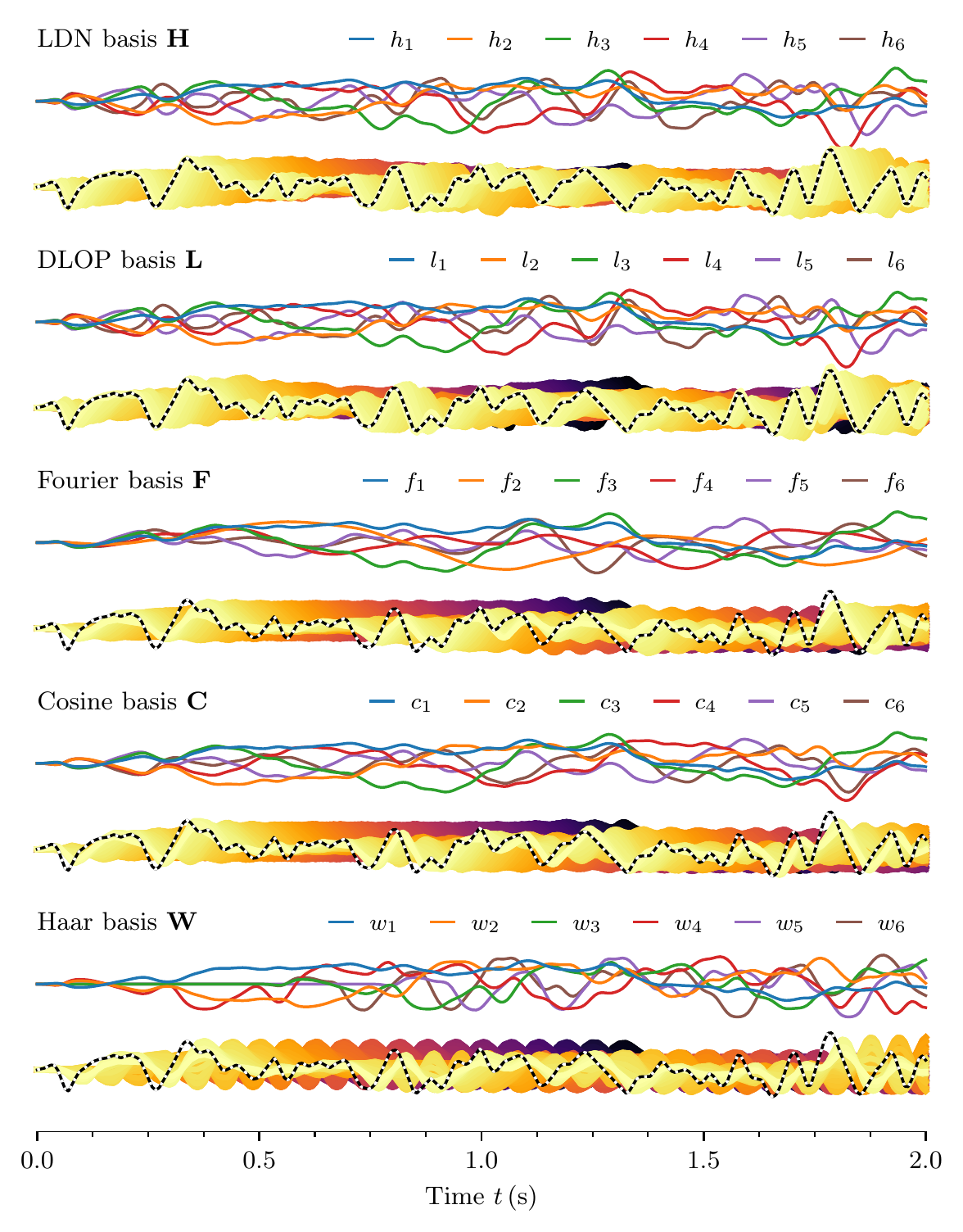}
  	\includegraphics[trim={0cm 0.125cm 0cm 0.2cm},clip]{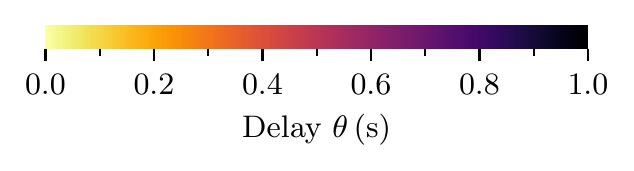}
  	\caption{Approximating delays using discrete function basis representations ($q = 16$ and $N = 128$).
  	Dotted black line is a sampled low-pass filterd white noise signal $\vec u$ with 256 samples.
  	Light yellow to purple lines correspond to a delayed version of $\vec u$ given as $\langle \vec d_\theta, \mat E \vec u_t \rangle$, where $\vec d_\theta$ is a decoding vector, and $\vec u_t$ are the last $N$ samples leading up to $t$.
  	The first dimensions of $\vec m = \mat E \vec u_t$ are depicted in the top half of each diagram.
  	$\theta = 1\,\mathrm{s}$ corresponds to $N = 128$ samples.
  	}
  	\label{fig:delay_reconstruction_example}
  \end{figure}
  \paragraph{Interpretation as \enquote{delay decoding}}
  The technique described above can be seen as decoding delayed versions of the discrete input signal $\vec u_t$.
  If the generalised Fourier coefficients $\vec m_t$ are computed over time for each new input sample $u_t$, then $\mat E \vec u_t$ describes a set of FIR filters (cf.~eq.~\ref{eqn:fir}), and $\langle d_\theta, \mat E \vec u_t\rangle$ reconstructs the sample from $\theta$ timesteps ago.

  \Cref{fig:delay_reconstruction_example} visualises this idea of \enquote{delay decoding} for an exemplary input signal $\vec u$ and the discrete Function bases discussed in this report (except for the naive discrete Legendre basis, which is very similar to the DLOP basis).
  Furthermore, the figure depicts the first six generalised Fourier coefficients $\vec m_t$. At each point in time, $\vec m_t$ encodes the last $N$ samples of the input signal.

  \begin{sidewaysfigure}
    \centering
    \includegraphics{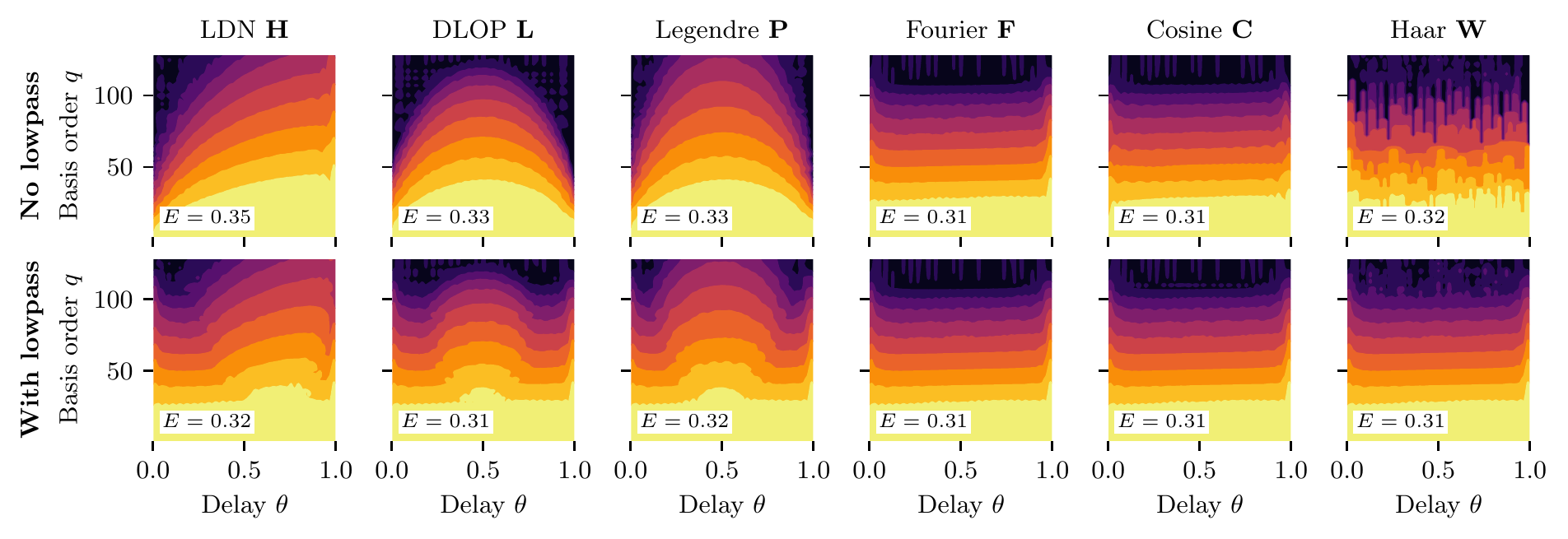}
    \hspace*{1.12cm}\includegraphics{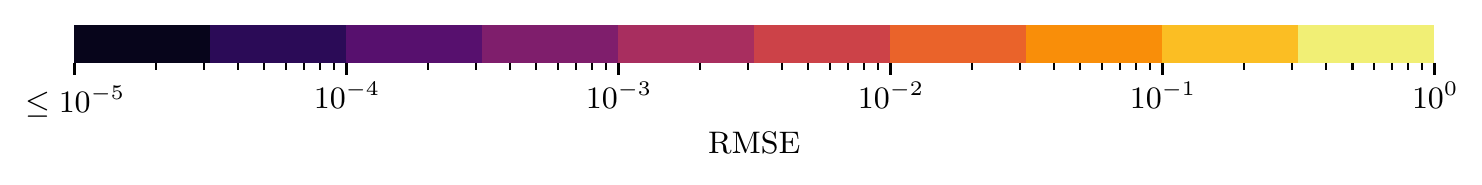}    
    \caption{Comparison between individual discrete function bases when decoding a delayed signal. For all bases, $N = 128$ corresponding to $1\,\mathrm{s}$. Each point is the RMSE over $1000$ test signals (for $51$ different $q$s and $51$ different $\theta$s). Each test signal is generated from low-pass filtered white noise (low-pass frequency is $15\,\mathrm{Hz}$).
  	The error $E$ in each diagram corresponds to the RMSE over all cells and trials.
  	The LDN is optimal for decoding a delay of $\theta$ close to zero. The DLOP and Legendre basis are optimal for both $\theta$ close to zero and one. The Fourier and cosine basis are similarly good for all decoded delays.}
  	\label{fig:evaluate_bases_contours}
  \end{sidewaysfigure}

  \paragraph{Results}
  \Cref{fig:evaluate_bases_contours} depicts the results of the experiment described above for $N = 128$ and a signal length of $256$.
  The figure depicts the overall average error $E$ for the discrete bases functions discussed in this report, including band-pass filtered versions of each basis $\mat E_\ell$ (cf.~eq.~\ref{eqn:filter}).
  Furthermore, contour plots break down the error for different delays $\theta$ and basis function counts $q$.
  Note that the displayed error scale is logarithmic.
  Even though some plots look drastically different, the absolute differences in error are not.\\

  \noindent\emph{Discrete LDN basis.}
  The LDN basis has the highest overall error with $E = 0.35$.
  Even though it was derived from an LTI system that optimally implements a delay of length $\theta = 1$, information is gradually lost as it propagates through the system.
  This is also visible in \Cref{fig:delay_reconstruction_example}, as well as the basis transformation matrix $\mat H$ in \Cref{fig:ldn_basis}, where the left half of the matrix is \enquote{fading out} for larger $q$.
  The lowest error is for a delay of length zero, which is of limited practical use.

  \noindent\emph{Discrete DLOP basis.}
  The DLOP basis has a slightly smaller overall error with $E = 0.33$.
  In contrast to the LDN basis, it can both reconstruct a delay of length $\theta = 0$ and $\theta = 1$ very well.
  The highest error is reached for $\theta = 0.5$.
  This can be explained by looking at the Legendre polynomials in \Cref{fig:legendre_series}.
  Half of the polynomials are zero at $x = 0.5$, and the other half has a relatively small magnitude for larger $q$, reducing the amount of information that can be decoded.\\

  \noindent\emph{Legendre basis.}
  The discretised Legendre basis reaches the same overall error as the DLOP basis with $E = 0.33$, and the overall shape of the contour plots is the same.
  The error is a little higher for larger $q$ compared to the DLOP basis, which is likely a result of the DLOP basis being orthogonal.\\

  \noindent\emph{Fourier and cosine basis.}
  These two bases have the smallest overall error with $E = 0.31$.
  In both cases, the decoding error is uniform over all delays $\theta$.\\

  \noindent\emph{Haar basis.}
  Surprisingly, the Haar basis has an error of only $E = 0.32$, which is slightly smaller than the overall error for the DLOP basis.
  However, as visible in \Cref{fig:delay_reconstruction_example}, only a fixed number of delays can be decoded well from the Haar basis; for $q = 16$, only $16$ different delay values can be decoded.
  Furthermore, in contrast to all other bases, increasing $q$ does not uniformly decrease the decoding error.
  This is because an individual basis function only provides local support.\\

  \noindent\emph{Filtering.}
  Filtering reduces the error for all bases to $E = 0.32$.
  This is not surprising, since, after filtering, each basis function is expressed in terms of the Fourier basis (cf.~\Cref{sec:filtering}).
  The decoding vector $\vec d_\theta$ can partially \enquote{undo} the corresponding linear combination and thus treat the generalized Fourier coefficients as if they were with respect to the Fourier basis.

  The most striking result is the effect filtering has on the Haar basis. After band-limiting the basis, or equivalently---see \Cref{sec:filtering}---the input signal, the error-over-delay plot for the Haar basis looks very similar to those of the Fourier and cosine basis.
  This suggests that the Haar basis is an excellent choice if the input signal $\vec u$ is band-limited in the Fourier domain.

  \subsection{Neural Network Experiments}
  \label{sec:nn}

  It is unclear in how far the results of the above experiments generalise to a neural network context.
  To gain a better understanding of how basis transformations behave in neural networks, we perform a series of experiments on two toy datasets: permuted sequential MNIST (psMNIST), and the Mackey-Glass system.

  Our experiments are mainly meant to answer to questions.
  First, we would like to know in how far the performance of the network depends on the selected basis, or, \enquote{temporal convolution}.
  Second, we compare fixed temporal convolutions to networks with the same architecture, where the temporal convolution is initialized in the same way, but trained along with the other parameters.
  This latter approach is equivalent to \enquote{Temporal Convolutional Networks}, an older idea that has lately been popularized by \citet{bai2018empirical}.

  There is an argument to be made that fixed convolutions can---under some circumstances---be a better choice compared to learned convolutions.
  Reduced run-time and memory requirements aside, fixed convolutions reduce the number of parameters in the network and may thus lead to faster convergence.
  At the same time, a fixed orthogonal basis spans a convenient space from which arbitrary functions can be linearly decoded.

  This is not to say that convolutional neural networks cannot learn good function bases---they surely can.
  An example of this---albeit in image processing---are Gabor-like filters found in the first convolution layer in deep convolutional networks \citep[cf.][]{krizhevsky2017imagenet}.
  However, learning such clean basis functions from scratch requires large datasets and many epochs of training.
  Furthermore, judging from the examples provided in the Krizhevsky paper, learned convolutions tend to be somewhat redundant.
  That is, they do not achieve the same degree of orthogonality as a hand-picked basis.

  \subsubsection{psMNIST}

  This task has originally been proposed by \citet{le2015simple}.
  The MNIST dataset \citep{lecun1998gradientbased} contains images of hand-written digits between zero and nine.
  Each image consists of $28 \times 28$ greyscale pixels.

  The idea of the psMNIST task is to treat each image as a sequence of $N = 784$ individual samples over time.
  Once the final sample has been processed, the system must correctly classify the digit.
  To eliminate spatial correlations in the sequence, a random but fixed permutation $\pi$ is applied to the samples.
  This results in a signal of the form $\vec u = (u_{\pi(1)}, \ldots, u_{\pi(784)})$.

  \paragraph{Comparability of psMNIST implementations}
  The psMNIST task as described above is a little under-defined.
  When comparing different neural architectures, two additional constraints should be met.
  First, all architectures must use the same amount of \emph{memory} \citep{voelker2019lmu}.

  Second, the resulting architecture should support \emph{serial execution}, or, borrowing terminology from \citet{chandar2019nonsaturating}, have a good \enquote{forgetting ability}.
  In other words, the network must still produce correct classifications over time, even if multiple input signals are concatenated.

  FIR filters intrinsically fulfil the serial execution constraint, but violate the memory constraint.
  Each filter requires access to all $N$ samples in memory.
  This essentially reduces the task to a non-sequential MNIST task---despite a set of $q$ FIR filters with $q < N$ forming an information bottleneck.

  Hence, unless an LTI system with $q$ state variables is used to implement the system, our results are not directly comparable to other psMNIST benchmarks.
  As we saw, reasonably good LTI systems can be constructed for most of the bases discussed here (cf.~\Cref{sec:filtering}).
  However, only the LDN basis has a near-finite impulse response and supports serial execution without external state resets.

  In the light of these constraints---and apart from the LDN basis results---our experiments are merely meant to compare different sets of basis functions, and not to contend with other approaches implementing psMNIST.

\begin{algorithm}[t]
\begin{lstlisting}
N = 28 * 28; M = 346; H = mk_basis(q, N)
model = tf.keras.models.Sequential([
  tf.keras.layers.Reshape((N, 1)),             # (N, 1)
  TemporalBasisTrafo(H, units=1),              # (1, q)
  tf.keras.layers.Dropout(0.5),                # (1, q)
  tf.keras.layers.Dense(M, activation="relu"), # (1, M)
  tf.keras.layers.Dense(10, use_bias=False),   # (1, 10)
  tf.keras.layers.Reshape((10,))               # (10)
])
\end{lstlisting}
\caption{Code used for the psMNIST experiment.
Comments $\mathtt{(Nt, Nd)}$ indicate the output dimensions of each layer.
$\mathtt{Nt}$ denotes the number of temporal samples; $\mathtt{Nd}$ is the number of non-temporal dimensions. $\mathtt{q}$ is between $1$ and $N$.}
  \label{alg:psmnist}
\end{algorithm}

	\paragraph{Methods}
	Our model is specified in \Cref{alg:psmnist}.
	A single temporal basis transformation layer\footnote{The code for the \texttt{TemporalBasisTrafo} class can be found at \url{https://github.com/astoeckel/temporal_basis_transformation_network}.} applies a set of $q$ FIR filters to $N$ input samples; 346 ReLU neurons non-linearly processes the transformed input, followed by a linear read-out layer.
	The psMNIST training set is randomly split into $50\,000$ training and $10\,000$ validation samples.
	An Adam optimizer with default parameters is used over 100 epochs with a batch size of 100.
	The reported test errors are computed for the parameter set with the smallest validation error.
	For $q = 468$ the number of trainable parameters is ${\sim}166$k. 
	Per default, the set of FIR filters are fixed and initialised with one of the discrete function bases.
	Another ${\sim}350$k parameters are added if the FIR filters are trained as well.

	\begin{figure}[t]
	  \centering
	  \includegraphics{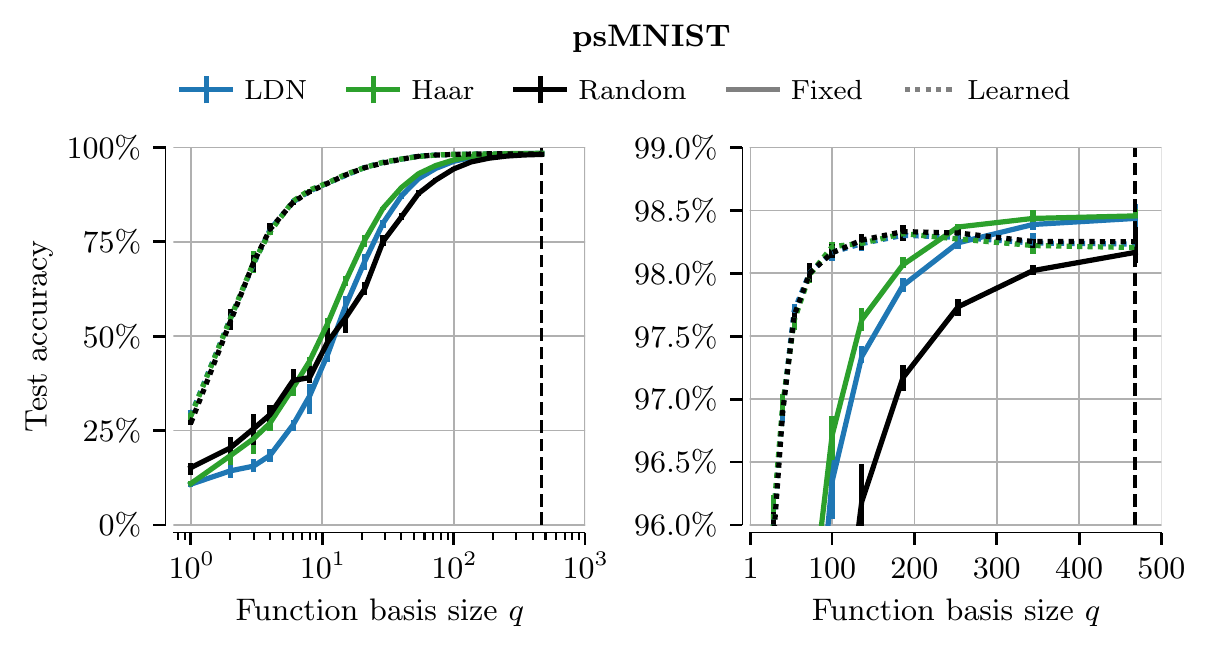}
	  \caption{Mean psMNIST classification accuracy for different discrete temporal function bases.
	  Data for the Haar basis are representative for the other fixed function bases (omitted for clarity).
	  Dotted lines are errors obtained when learning the temporal convolution.
	  Each data point is for $n = 11$ trials; error bars correspond to the first and third quartile.
	  Dashed vertical line is at $q = 468$; detailed data for this $q$ are given in \Cref{tbl:psmnist_single} and \Cref{fig:psmnist_single}.}
	  \label{fig:psmnist}
	\end{figure}

	\begin{figure}
	  \centering
	  \includegraphics{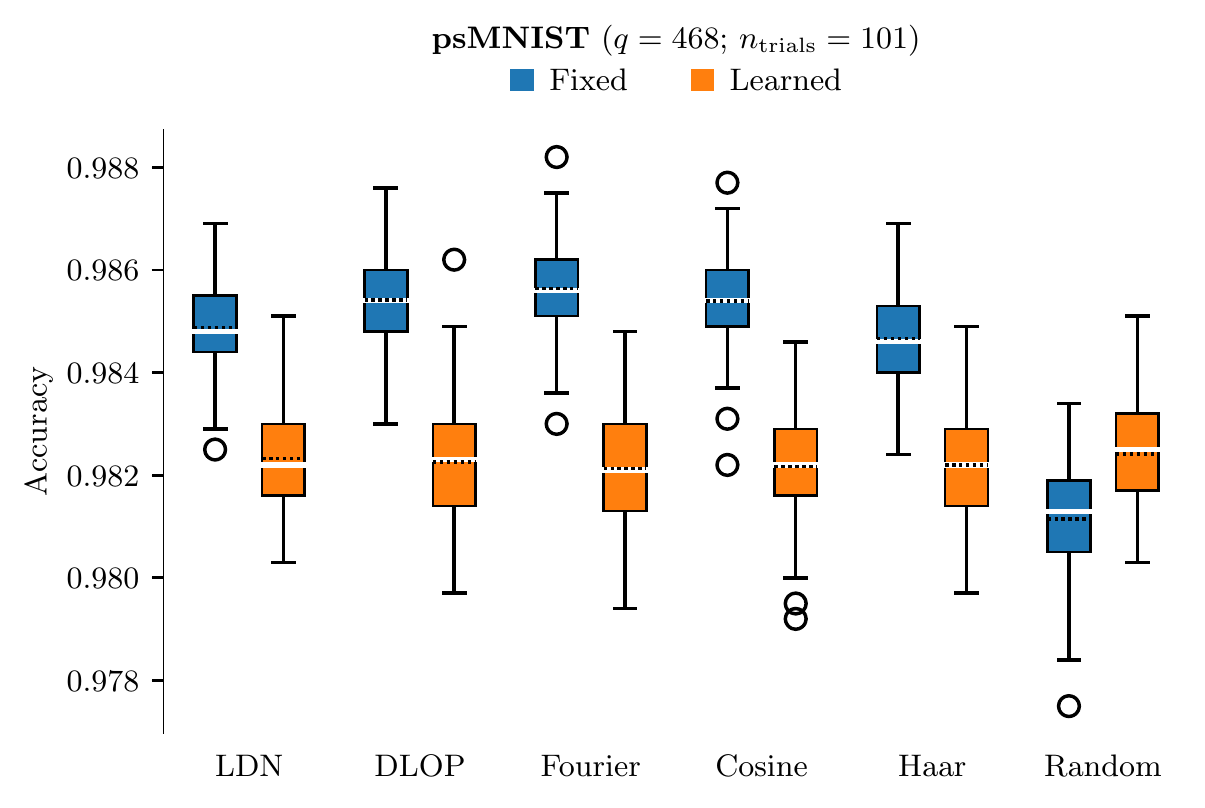}
	  \caption{Classification accuracies for the psMNIST dataset using different discrete function bases as temporal convolutions with $q = 468$.
	  Depicted are standard box-plots over $n = 101$ trials for each basis. Box corresponds to the first and third quartile; whiskers are the minimum/maximum after outlier rejection; outliers are depicted as circles.
	  Thick white line is the median, dashed black line is the mean.
	  Numerical values are given in \Cref{tbl:psmnist_single}.}
	  \label{fig:psmnist_single}
	\end{figure}

	\begin{table}
		\caption{Test accuracies for the psMNIST experiment for $q = 468$. Data over $n = 101$ trials and $100$ epochs. Q1 and Q3 are the 25- and 75-percentile, respectively. The best three results are highlighted in each column (darker colours are better).}
		\label{tbl:psmnist_single}
		\centering\small
		\begin{tabular}{r  r r r r  r r r r}
		\toprule
		& \multicolumn{4}{c}{{\color{tabBlue}$\blacksquare$} \textbf{Fixed convolution}}
		& \multicolumn{4}{c}{{\color{tabOrange}$\blacksquare$} \textbf{Learned convolution}} \\
		\cmidrule(r){2-5}\cmidrule(l){6-9}
		\parbox{1cm}{\raggedleft\vspace{-0.41cm}Initial\\basis} &
		Mean &
		Median &
		Q1 &
		Q3 &
		Mean &
		Median &
		Q1 &
		Q3\\
		\midrule
		LDN &
		98.49 &
		98.48 &
		98.44 &
		98.55 &
		 \cellcolor{CornflowerBlue!50}{98.23} &
		 \cellcolor{CornflowerBlue!25}{98.22} &
		 \cellcolor{CornflowerBlue!50}{98.16} &
		 \cellcolor{CornflowerBlue!50}{98.30} \\
		DLOP &
		 \cellcolor{CornflowerBlue!50}{98.54} &
		 \cellcolor{CornflowerBlue!50}{98.54} &
		 \cellcolor{CornflowerBlue!25}{98.48} &
		 \cellcolor{CornflowerBlue!50}{98.60} &
		 \cellcolor{CornflowerBlue!50}{98.23} &
		 \cellcolor{CornflowerBlue!50}{98.23} &
		98.14 &
		 \cellcolor{CornflowerBlue!50}{98.30} \\
		Fourier &
		 \cellcolor{CornflowerBlue!75}{98.56} &
		 \cellcolor{CornflowerBlue!75}{98.56} &
		 \cellcolor{CornflowerBlue!75}{98.51} &
		 \cellcolor{CornflowerBlue!75}{98.62} &
		98.21 &
		98.21 &
		98.13 &
		 \cellcolor{CornflowerBlue!50}{98.30} \\
		Cosine &
		 \cellcolor{CornflowerBlue!50}{98.54} &
		 \cellcolor{CornflowerBlue!50}{98.54} &
		 \cellcolor{CornflowerBlue!50}{98.49} &
		 \cellcolor{CornflowerBlue!50}{98.60} &
		98.22 &
		 \cellcolor{CornflowerBlue!25}{98.22} &
		 \cellcolor{CornflowerBlue!50}{98.16} &
		98.29 \\
		Haar &
		98.47 &
		98.46 &
		98.40 &
		98.53 &
		98.22 &
		 \cellcolor{CornflowerBlue!25}{98.22} &
		98.14 &
		98.29 \\
		Random &
		98.11 &
		98.13 &
		98.05 &
		98.19 &
		 \cellcolor{CornflowerBlue!75}{98.24} &
		 \cellcolor{CornflowerBlue!75}{98.25} &
		 \cellcolor{CornflowerBlue!75}{98.17} &
		 \cellcolor{CornflowerBlue!75}{98.32} \\
		\bottomrule
		\end{tabular}
	\end{table}

    \paragraph{Results}
    Results are depicted in \Cref{fig:psmnist}, \Cref{fig:psmnist_single} and \Cref{tbl:psmnist_single}.
    For $q > 20$ there is no appreciable difference between the function bases, with the LDN basis having slightly lower accuracies on average.
    When learning the temporal convolution, differences between the individual initializations disappear.
    Learning yields drastically better results for $q < 200$ with a peak at about $q = 190$ with an accuracy of about $98.4\%$.
    The accuracy is monotonically increasing with $q$ when using fixed convolutions, reaching about $98.5\%$ for $q = 468$.
    A random initialization reaches error rates of about $98.1\%$.

	\subsubsection{Mackey-Glass}
	The point of this task is to predict the time-course of the chaotic Mackey-Glass dynamical system for a certain number of samples.
	The Mackey-Glass system is
	\begin{align*}
		\frac{\mathrm{d}}{\mathrm{d}t} \vec x(t) &= \frac{a x(t - \tau)}{1 + x(t - \tau)^{10}} - b x(t) \,,
	\end{align*}
	where $a = 0.2$, $b = 1.2$, and $x(t) = 1.2 + \eta$ with $\eta \sim \mathcal{N}(\mu=0, \sigma=1)$ for $t < 0$.
	This system is chaotic for $\tau > 17$ and has been extensively used as a benchmark in time-series prediction \citep[cf.][Section 4.3.1]{mendel2017uncertain}.

	\paragraph{Dataset}
	For our experiments, we choose $\tau = 30$.
	As a training dataset we generate 400 Mackey-Glass trajectories of length 10\,000.
	The system as been discretised with a timestep of $\Delta t = 1$ using a fourth order Runge-Kutta integrator.
	All trajectories are different due to the stochasticity of the initial $x(t)$ for $t < 0$.
	We randomly extract 100 input sequences of length 33 followed by a target sequence of length 15 from each trajectory.
	This results in 40\,000 training samples.
	The task is to predict the 15 target samples from the preceding 33.

	We similarly generate 10\,000 validation and 10\,000 test samples.
	Training, validation, and test samples are all taken from separately generated trajectories.

\begin{algorithm}
\begin{lstlisting}
N_units0, N_units1, N_units2, N_units3 = 1, 10, 10, 10
N_wnd = N_wnd0 + N_wnd1 + N_wnd2 + N_wnd3 - 3
q0, q1, q2, q3 = 16, 8, 8, 4
H0, H1 = mk_basis(q0, N_wnd0), mk_basis(q1, N_wnd1)
H2, H3 = mk_basis(q2, N_wnd2), mk_basis(q3, N_wnd3)
model = tf.keras.models.Sequential([
  tf.keras.layers.Reshape((N_wnd, 1)),
  # (N_wnd0 + N_wnd1 + N_wnd2 + N_wnd3 - 3, 1)
  TemporalBasisTrafo(H0, n_units=N_units0, pad=False),
  # (N_wnd1 + N_wnd2 + N_wnd3 - 2, q0 * N_units0)
  tf.keras.layers.Dense(N_units1, activation="relu"),
  # (N_wnd1 + N_wnd2 + N_wnd3 - 2, N_units1)
  TemporalBasisTrafo(H1, n_units=N_units1, pad=False),
  # (N_wnd2 + N_wnd3 - 1, q1 * N_units1)
  tf.keras.layers.Dense(N_units2, activation="relu"),
  # (N_wnd2 + N_wnd3 - 1, N_units2)
  TemporalBasisTrafo(H2, n_units=N_units2, pad=False),
  # (N_wnd3, q2 * N_units2)
  tf.keras.layers.Dense(N_units3, activation="relu"),
  # (N_wnd3, N_units3)
  TemporalBasisTrafo(H3, n_units=N_units3, pad=False),
  # (1, q3 * N_units3)
  tf.keras.layers.Dense(N_pred, use_bias=False),
  # (1, N_pred)
  tf.keras.layers.Reshape((N_pred,))
  # (N_pred)
])
\end{lstlisting}
	\caption{Code used for the Mackey-Glass experiment.
	Comments $\mathtt{(Nt, Nd)}$ indicate the output dimensions of the layer in the preceding line.
	$\mathtt{Nt}$ is the number of temporal samples; $\mathtt{Nd}$ is the number of non-temporal dimensions.
	Total number of trainable parameters (with fixed convolutions) is 2390.}
	\label{alg:mackey_glass}
\end{algorithm}

	\paragraph{Methods}
	We use the neural network architecture described in \Cref{alg:mackey_glass}.
	This architecture is inspired by the architecture described by \citet{voelker2019lmu}.
	To summarise, there are four cascading temporal basis transformation layers.
	The number of FIR filters $q$ in each layer is equal to the number of input samples $N$, so no stage loses information.
	The first layer uses $q = N = 16$, the second and third $q = N = 8$, and the final layer $q = N = 4$, resulting in a total input window length of $16 + 8 + 8 + 4 - 3 = 33$.

	The second to fourth layers consist of ten independent units using exactly the same set of convolutions.
	Layers are densely connected using ReLUs.
	Again, we compare fixed convolutions---including random initialization---to a version of the network where the convolutions are initialized in the same way, but then further refined during training.
	The learned convolutions are normalized, such that each learned basis function has norm one.
	The total number of trainable parameters is 2760; another 400 are added if the convolutions are trained.

	We train the network for 100 epochs using a standard Adam optimizer.
	The final test error is computed for the parameters in the epoch with the smallest validation error.
	We measure the normalised RMSE, i.e., the RMSE divided by the RMS of the Mackey-Glass trajectories (${\approx}0.94$).

\begin{figure}
  \centering
  \includegraphics{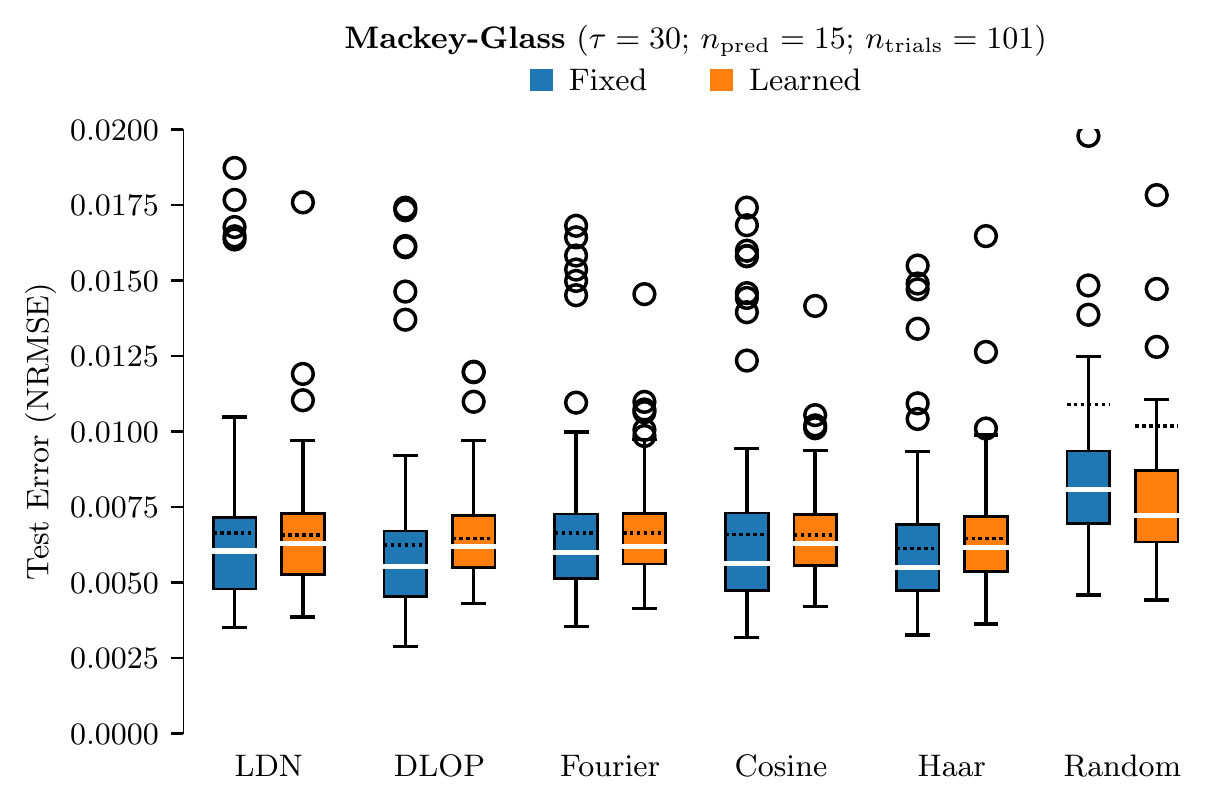}
  \caption{Prediction errors for the Mackey-Glass dataset using different discrete function bases as fixed temporal convolutions. See \Cref{fig:psmnist_single} for more detail on the box-plots.
  Numerical values are given in \Cref{tbl:mackey_glass}.}
  \label{fig:mackey_glass}
\end{figure}

\begin{table}
	\caption{Test errors for the Mackey-Glass experiment. Data over $n = 101$ trials and $100$ epochs. Q1 and Q3 are the 25- and 75-percentile, respectively. The best three results are highlighted in each column (darker colours are better).}
	\label{tbl:mackey_glass}
	\small\hspace*{-0.3cm}
	\begin{tabular}{r  r r r r  r r r r}
	\toprule
	& \multicolumn{4}{c}{{\color{tabBlue}$\blacksquare$} \textbf{Fixed convolution}}
	& \multicolumn{4}{c}{{\color{tabOrange}$\blacksquare$} \textbf{Learned convolution}} \\
	\cmidrule(r){2-5}\cmidrule(l){6-9}
	\parbox{1cm}{\raggedleft\vspace{-0.41cm}Initial\\basis} &
	Mean &
	Median &
	Q1 &
	Q3 &
	Mean &
	Median &
	Q1 &
	Q3\\
	\midrule
	LDN &
	0.0067 &
	0.0061 &
	0.0048 &
	 \cellcolor{CornflowerBlue!25}{0.0072} &
	 \cellcolor{CornflowerBlue!25}{0.0066} &
	0.0063 &
	 \cellcolor{CornflowerBlue!75}{0.0053} &
	 \cellcolor{CornflowerBlue!25}{0.0073} \\
	DLOP &
	 \cellcolor{CornflowerBlue!50}{0.0063} &
	 \cellcolor{CornflowerBlue!75}{0.0055} &
	 \cellcolor{CornflowerBlue!75}{0.0045} &
	 \cellcolor{CornflowerBlue!75}{0.0067} &
	 \cellcolor{CornflowerBlue!75}{0.0065} &
	 \cellcolor{CornflowerBlue!75}{0.0062} &
	 \cellcolor{CornflowerBlue!25}{0.0055} &
	 \cellcolor{CornflowerBlue!75}{0.0072} \\
	Fourier &
	0.0067 &
	0.0060 &
	0.0051 &
	0.0073 &
	0.0067 &
	 \cellcolor{CornflowerBlue!75}{0.0062} &
	0.0056 &
	 \cellcolor{CornflowerBlue!25}{0.0073} \\
	Cosine &
	 \cellcolor{CornflowerBlue!25}{0.0066} &
	 \cellcolor{CornflowerBlue!25}{0.0057} &
	 \cellcolor{CornflowerBlue!50}{0.0047} &
	0.0073 &
	 \cellcolor{CornflowerBlue!25}{0.0066} &
	0.0063 &
	0.0056 &
	 \cellcolor{CornflowerBlue!25}{0.0073} \\
	Haar &
	 \cellcolor{CornflowerBlue!75}{0.0061} &
	 \cellcolor{CornflowerBlue!75}{0.0055} &
	 \cellcolor{CornflowerBlue!50}{0.0047} &
	 \cellcolor{CornflowerBlue!50}{0.0069} &
	 \cellcolor{CornflowerBlue!75}{0.0065} &
	 \cellcolor{CornflowerBlue!75}{0.0062} &
	 \cellcolor{CornflowerBlue!50}{0.0054} &
	 \cellcolor{CornflowerBlue!75}{0.0072} \\
	Random &
	0.0109 &
	0.0081 &
	0.0070 &
	0.0094 &
	0.0102 &
	0.0072 &
	0.0063 &
	0.0087 \\
	\bottomrule
	\end{tabular}
\end{table}

  \paragraph{Results}
  Results are depicted in \Cref{fig:mackey_glass} and \Cref{tbl:mackey_glass}.
  All fixed convolutions exhibit approximately the same performance.
  The random initialization has the highest error, and the Haar and DLOP filters result in the lowest error.
  Learning the convolution results in a slightly higher error overall, but closes the performance gap between the individual basis transformations.

  \section{Discussion}

  We provided a whirlwind overview of discrete function bases with a particular focus on the LDN and DLOP bases.
  When used as FIR filters, these bases can compress temporal data into a stream of generalized Fourier coefficients.

  Since the LDN basis was constructed from an LTI system with an almost finite impulse response, this convolution operation can be approximated \enquote{online} by recurrently advancing the discretised LTI system.
  This requires only $\mathcal{O}(q)$ \emph{state} memory ($\mathcal{O}(q^2)$ in total), but, in general, $\mathcal{O}(q^2)$ operations per sample.
  This can be faster than online, zero-delay convolution using a set of FIR filters, which require at least $\mathcal{O}(q\log(N))$ operations per sample, but $\mathcal{O}(N\log(N))$ memory.

  The LDN is similar in principle to the sliding transformations for the Haar, cosine, and Fourier basis.
  These transformations require $\mathcal{O}(q)$ update operations at the cost of $\mathcal{O}(N)$ memory.
  In contrast, memory and run-time requirements of the LDN can be reduced to $\mathcal{O}(q)$ when using the Euler update, although there are some potential caveats when using this approach (cf.~\Cref{sec:ldn_when}).

  While the DLOP basis performs better than the LDN basis, we are not aware of a fast update scheme.
  This basis should only be used if this is not a concern.

  It is surprising to see that fixed, (almost) orthogonal temporal convolutions can yield better results than learned convolutions in neural networks.
  As seen in the psMNIST experiment, increasing the number of basis functions $q$ can decrease the performance of the network when using learned convolutions.
  Doing the same for fixed convolutions seems to monotonically increase the performance of the system.
  This is somewhat intriguing, since fixed convolutions result in fewer trainable parameters, and, as discussed above, enable the use of efficient update schemes.
  In general, learned convolutions must resort to the $\mathcal{O}(q \log(N))$ \citet{gardner1995efficient} algorithm.
  Future work is required to validate these results and to understand under which circumstances fixed convolutions can make sense.

  \section*{Acknowledgements}

  This technical report was inspired by work done by Narsimha R.~Chilkuri on feed-forward networks exploiting the LDN basis transformation.
  The original approach outlined in \Cref{sec:construct_ldn_basis} is his idea, including \cref{eqn:ldn_recurrence,eqn:ldn_h_matrix}.
  He also suggested to think about the LDN in terms of \enquote{optimally forgetting} signals outside the time-window, which is not necessary when using FIR filters.
  We furthermore thank Aaron R.~Voelker and Chris Eliasmith for their feedback on this report.

  \printbibliography

  \section*{Revisions}
  \begin{enumerate}[(1)]
  	\item January 11, 2021. First internal version of this document.
  	\item January 26, 2021. Added the ZOH vs.~Euler error diagram \Cref{fig:ldn_zoh_vs_euler}. Discussed sliding transforms in \Cref{sec:ldn_when}; expanded the discussion of the asymptotic complexity of the ZOH vs.~Euler LDN update. Added \Cref{fig:ldn_spectrum} and \Cref{tbl:sliding_trafos}. Added discussion of dampening by \enquote{information erasure}. Added \Cref{sec:nn}.  Using a reordered Haar transformation matrix; commented on the effects of filtering this better Haar matrix. Rewrote the abstract and discussion to take these changes into account.
  	\item March 9, 2021. Publication to arXiv. Added reference to the \enquote{Constructing Dampened LTI Systems Generating Polynomial Bases} technical report.
  \end{enumerate}

\end{document}